\documentclass[final,5p,times,twocolumn]{elsarticle}

\usepackage{float,placeins,etoolbox}
\usepackage{hyperref}

\journal{-}

\usepackage{hhline}

\usepackage{adjustbox}

\usepackage{amsmath}
\usepackage{diagbox}
\usepackage{subfigure}
\usepackage{listings} 

\usepackage[utf8]{inputenc}
\usepackage{booktabs}
\usepackage{multirow}
\usepackage{geometry}
\usepackage{tabularx}

\DeclareFixedFont{\ttb}{T1}{txtt}{bx}{n}{8} 
\DeclareFixedFont{\ttm}{T1}{txtt}{m}{n}{8}

\usepackage{color}
\definecolor{tensorred}{rgb}{1,0,0}
\let\RSEOriginalTextColor\textcolor
\renewcommand{\textcolor}[2]{\ifstrequal{#1}{red}{\RSEOriginalTextColor{black}{#2}}{\RSEOriginalTextColor{#1}{#2}}}
\definecolor{deepblue}{rgb}{0,0,0.5}
\definecolor{deepred}{rgb}{0.6,0,0}
\definecolor{deepgreen}{rgb}{0,0.5,0}

\definecolor{dkgreen}{rgb}{0,0.6,0}
\definecolor{mauve}{rgb}{0.58,0,0.82}
\definecolor{desertpurple}{rgb}{0.525490,0.176470,0.525490}

\newcommand\pythonstyle{\lstset{
language=Python,
basicstyle=\ttm,
numberstyle={\tiny},
otherkeywords={self,as},             
keywordstyle=\ttb\color{deepblue},
commentstyle=\color{mauve},
emph={MyClass,__init__},          
emphstyle=\ttb\color{deepred},    
stringstyle=\color{deepgreen},
frame=tb,                         
showstringspaces=false            
}}

\usepackage{booktabs}    
\usepackage{pifont}      
\newcommand{\cmark}{\ding{51}} 
\newcommand{\xmark}{\ding{55}} 

\usepackage[utf8]{inputenc}   
\usepackage[T1]{fontenc}
\usepackage{enumitem}

\lstnewenvironment{python}[1][]
{
\pythonstyle
\lstset{#1}
}
{}

\usepackage{comment}
\newsavebox{\measurebox}

\hypersetup{
    colorlinks,
    linkcolor={red!50!black},
    citecolor={blue!50!black},
    urlcolor={blue!50!black}
}

\biboptions{authoryear}

\usepackage{float,placeins,etoolbox}
\usepackage{pgfplots}
\pgfplotsset{compat=1.18}
\usepackage{microtype}
\usepackage{xurl}
\let\RSEOriginalTextColor\textcolor
\renewcommand{\textcolor}[2]{\ifstrequal{#1}{red}{\RSEOriginalTextColor{black}{#2}}{\RSEOriginalTextColor{#1}{#2}}}
\hypersetup{linkcolor=blue,citecolor=blue,urlcolor=blue,pdfauthor={Foivos I. Diakogiannis et al.}}
\makeatletter
\def\ps@pprintTitle{\let\@oddhead\@empty\let\@evenhead\@empty\def\@oddfoot{\hfil\thepage\hfil}\let\@evenfoot\@oddfoot}
\makeatother
\begin{document}

\begin{frontmatter}
\title{Tackling fluffy clouds: robust \textcolor{red}{agricultural} field boundary delineation \textcolor{red}{from} Sentinel-1 and Sentinel-2 \textcolor{red}{satellite image time series}}
\author[data61]{Foivos I. Diakogiannis\fnref{myfootnote1}}
\author[data61]{Zheng-Shu Zhou}
\author[data61]{Jeff Wang}
\author[AF]{Gonzalo Mata}
\author[AFM]{Dave Henry}
\author[AF]{Roger Lawes}
\author[AS]{Amy Parker}
\author[data61]{Peter Caccetta}
\author[data61]{Suzanne Furby}
\author[uniastra]{Rodrigo Ibata}
\author[imt]{Ondrej Hlinka}
\author[AF]{Jonathan Richetti}
\author[AF]{Kathryn Batchelor}
\author[AF]{Chris Herrmann}
\author[AF]{Andrew Toovey}
\author[ANU]{John Taylor}

\address[data61]{\textcolor{red}{Technology, CSIRO, Kensington WA, Australia}}
\address[AF]{A\& F, CSIRO, Floreat WA, Australia}
\address[AFM]{A\& F, CSIRO, Melbourne, VIC 3030, Australia}
\address[AS]{CSIRO Space and Astronomy, Kensington WA, Australia}
\address[uniastra]{University of Strasbourg, France}
\address[imt]{IM\&T CSIRO,  Australia}
\address[ANU]{Australian National University, School of Computing, ACT, Australia}
\fntext[myfootnote1]{\textcolor{red}{foivos.diakogiannis@csiro.au}}

\begin{abstract}
Accurate delineation of agricultural field boundaries is essential for effective crop monitoring and resource management. However, competing methodologies often face significant challenges, particularly in their reliance on extensive manual efforts for cloud-free data curation and limited adaptability to diverse global conditions.
In this paper, we introduce PTAViT3D, a deep learning architecture specifically designed for processing three-dimensional time series of satellite imagery from either Sentinel-1 (S1) or Sentinel-2 (S2). Additionally, we present PTAViT3D-CA, an extension of the PTAViT3D model incorporating cross-attention mechanisms to fuse S1 and S2 datasets, enhancing robustness in cloud-contaminated scenarios.
The proposed methods leverage spatio-temporal correlations through a memory-efficient 3D Vision Transformer architecture, facilitating accurate boundary delineation directly from \textcolor{red}{preprocessed, cloud-affected} imagery.
We comprehensively validate our models through extensive testing on various datasets, including Australia's ePaddocks™ -- 
CSIRO’s national, continental-scale agricultural field boundary product covering Australia’s cropping regions 
 -- alongside public benchmarks Fields-of-the-World, PASTIS, and AI4SmallFarms.
Our results consistently demonstrate state-of-the-art performance, highlighting excellent global transferability and robustness. Crucially, our approach significantly simplifies data preparation workflows by reliably processing cloud-affected imagery, thereby offering strong adaptability across diverse agricultural environments.
Our code and models are publicly available at \href{https://github.com/feevos/tfcl}{https://github.com/feevos/tfcl}.
\end{abstract}


\end{frontmatter}

\begin{figure}[!htbp]
\begin{center}
\includegraphics[clip, trim=0.25cm 0.0cm 0.0cm 0.1cm,width=\columnwidth,height=0.72\textheight,keepaspectratio]{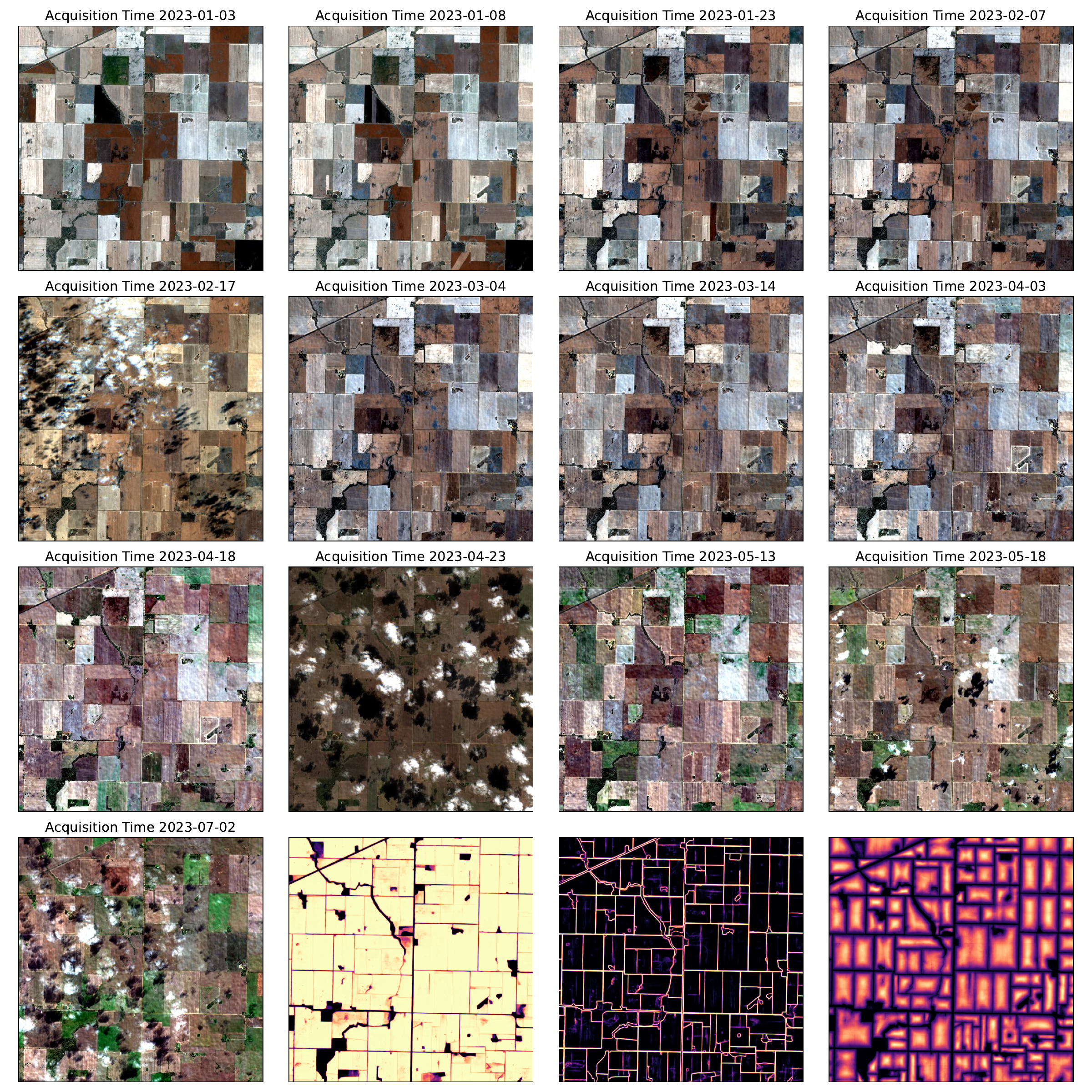}
\end{center}
\caption{\textcolor{red}{The proposed algorithm performs inference on S2 or S1 image time series and can combine both modalities, reducing reliance on labor-intensive cloud-free optical image selection. This example shows inference by the S2-only PTAViT3D model over Australia. The last three panels show the predicted field-extent, field-boundary, and distance-transform rasters \citep{DIAKOGIANNIS202094,WALDNER2020111741,rs13112197}.}}
\label{Main-ptavit3d_cloud_demo}
\end{figure}

\section{Introduction}

The rise of digital agriculture has revolutionized how we monitor, manage, and optimize crop production. At the heart of this transformation lies the precise delineation of field boundaries, a critical component for maximizing efficiency in agricultural resource management. Accurate field boundary mapping significantly impacts applications such as crop yield estimation and food security assessments \citep{deWit2004, blaes2005, rs71013208}. Historically, boundary mapping was a labor-intensive and error-prone process, leaving farmers and digital service providers grappling with inaccuracies. Today, frequent satellite observations from Sentinel-1 (S1) and Sentinel-2 (S2) provide opportunities to continuously refine these boundaries in sync with cropping cycles.

\subsection{Motivation}
Field boundaries are not purely static, geometric entities; in practice they become most separable when adjacent paddocks follow different temporal trajectories (planting dates, crop types, growth dynamics, management interventions, and harvest timing). This makes time series information central to accurate delineation, particularly in rainfed systems where phenology is a strong signal \citep{Zhong2019MultiTemporalCropDL, Blickensdoerfer2022CropSequences}. Consequently, methods that exploit Satellite Image Time Series (SITS) have become increasingly important for large-scale agricultural mapping.

However, the effective temporal density of optical S2 time series is often reduced by clouds, haze, smoke, and associated shadows, especially during the growing season. Cloud masking is therefore a critical preprocessing step, yet it remains imperfect and dependent on thresholds and scene context \citep{TARRIO2020100010, LI202289,  SKAKUN2022112990, WRIGHT2024114122}. A common practical workaround is cloud-free compositing, which can be robust for generating seasonal or annual snapshots. The limitation for boundary delineation is that compositing compresses multiple observations into a single representative image, trading temporal density for cleanliness and making the choice of compositing window strongly application-dependent \citep{Frantz2017PhenologyAdaptiveCompositing, Roberts2017HighDimPixelComposites, Griffiths2019IntraAnnualComposites}. In other words, compositing is valuable, but it can discard exactly the high-frequency temporal information that makes paddocks separable. Prior work has also highlighted that \emph{single-date} predictions can be unstable due to clouds, phenology, and acquisition-specific artefacts, and that robustness improves when delineations are \emph{consolidated across multiple dates} (e.g., via consensus/averaging across several observations), effectively requiring agreement over time \citep{WALDNER2020111741, rs13112197}. 

These considerations motivate models that use the entire time series and learn to integrate partially contaminated observations, rather than relying on strict cloud-free pixels or single-date composites. While S1 SAR is immune to cloud cover, its radiometric and geometric characteristics differ from optical imagery and its signal-to-boundary relationship is not always straightforward. This motivates learning-based approaches that can leverage each modality where it is strongest.

\subsection{Related work}
Traditional field boundary extraction methods, including edge-based, region-based, and hybrid techniques, have had varying degrees of success \citep{MUELLER20041619, TURKER2013106, YAN201442, GRAESSER2017165}. These approaches often struggle with noise, require extensive preprocessing, and generalize poorly across landscapes. Vision-based deep learning models, in contrast, learn features directly from imagery and have demonstrated robust and scalable boundary extraction \citep{PERSELLO2019111253, WALDNER2020111741, 9150690, rs13112197, rs14225738, essd-15-317-2023, Tetteh2023,Lavreniuk2025DelineateAnything}.  

Recent work has increasingly exploited SITS to improve agricultural mapping and parcel delineation. Garnot and Landrieu introduced temporal attention models for panoptic segmentation of satellite image time series \citep{garnot2021panoptic}, and related temporal attention variants have also been explored for crop mapping \citep{Garnot2022MultimodalTemporalAttention}. At national scale, \cite{SolanoCorrea2022CountryScaleFields} demonstrated large-scale field mapping pipelines based on Sentinel-2 SITS. Multimodal fusion approaches integrate complementary data streams: DSTFNet \citep{CAI202334} combines multimodal inputs with spatiotemporal modeling, and FieldSeg-DA2.0 \citep{TIAN2024109050} enhances transferability via domain adaptation. In parallel, transformer architectures for satellite image time series have gained traction, demonstrating that attention mechanisms can effectively model temporal dependencies for remote sensing tasks \citep{Tarasiou2023ViTsForSITS}.

\textcolor{red}{Harmonic vegetation-index representations provide a distinct 2D strategy by encoding seasonal phase, amplitude, and mean as image channels for field delineation \citep{Papic2025HarmonicNDVI,Papic2026HarmonicPhenology}.}

These studies highlight two recurring constraints for field boundary delineation under real-world conditions: (i) an incomplete integration of spatial and temporal dimensions, often treating them as separable, and (ii) continued dependence on cloud-free inputs, aggressive filtering, or external high-resolution data. To address these limitations, we propose a unified 3-D Vision Transformer architecture employing a novel 4D Patch Tanimoto Attention (PTA3D) mechanism that jointly encodes channel, spatial, and temporal dimensions in a single 4-D similarity map. This design enables direct processing of \textcolor{red}{preprocessed} S2 or S1 time series, optionally integrating both via a cross-attention mechanism, and supports cloud-tolerant delineation without reliance on cloud-free compositing.

This work introduces several key contributions:
\begin{enumerate}
\item A 3D Vision Transformer architecture specifically tailored to process SITS data, expanding upon our prior work in semantic segmentation \citep[SSG2;][]{DIAKOGIANNIS202444}.
\item A memory-efficient and effective 3D vision attention mechanism designed explicitly for satellite image time series.
\item Two implementations: PTAViT3D for processing either S2 or S1 time series, and PTAViT3D-CA, integrating S2 and S1 data via cross-attention.
\item \textcolor{red}{Demonstration of model performance on partially cloud-affected Sentinel-2 sequences under operational scene-selection thresholds (up to 40\% scene-level cloud cover) and controlled cloud/shadow perturbations, together with Sentinel-1 time-series processing as an all-weather alternative in persistently cloudy regions.}
\item Demonstration of transferability across regions, by benchmarking against publicly available datasets: Fields of the World \citep[FoTW;][]{kerner2024fields}, the PASTIS dataset \citep{garnot2021panoptic}, and the \linebreak AI4SmallFarms dataset \citep{10278130}, covering diverse agricultural landscapes and challenges.
\item A roadmap application for large-scale agricultural boundary delineation, with the application and validation of PTAViT3D across the Australian cropping regions, with the creation of the 2023 ePaddocks national product of Australia.
\end{enumerate}

\section{Methods}

\subsection{Data}

Here we describe the datasets we used for  evaluating our modeling approach. The first dataset is the ePaddocks dataset, provided by CSIRO, in 2021 version. In addition to that we also describe benchmark datasets that enables us to compare our algorithmic approach with baselines and state of the art in identical settings. We utilize the 2023 ePaddocks product as a benchmark against variable cloud conditions as well as, as a complete walk through / recipe for obtaining agricultural field boundaries at continental scale with detailed pipelines.

\subsubsection{The ePaddocks field boundaries dataset of Australia}
\label{Main-s2_section_data}

\textbf{Sentinel 2}: The development of our training dataset, based on Sentinel-2 imagery for field boundary detection, follows the methodology outlined in \citep{rs13112197}. For completeness, we recap the process briefly. Training data were sourced from 2019 imagery, level L2A, and in particular bands blue (B2 490nm), green(B3 560nm), red (B4 665nm) and near-infrared (NIR 842nm). Additional imagery with 10\% cloud tolerance was obtained for the same dates using the \cite{OpenDataCubeWeb} Software. This new imagery was used after pretraining on the original data, serving as a fine-tuning procedure.

The creation of the annotated data was a model-assisted manual process. Initial field objects were derived through automated segmentation on six selected tiles (50JPL, 53HNC, 54HXE, 55HDA, 55HEE, 56JLQ) with clear-sky imagery. \textcolor{red}{These are six complete Sentinel-2 MGRS tile footprints, each approximately \(100\,\mathrm{km}\times100\,\mathrm{km}\), rather than six local sample windows.} The output was refined by masking non-agricultural areas using CLUM \citep[Catchment Scale Land Use of Australia;][]{ABARES2016} and delineating fields via a meanshift algorithm \citep{6849524}. Parameter optimization was guided by Bayesian techniques \citep{NIPS2012_05311655} against a sample of fields. This was the starting point for further  manual edits and adjustments that ensured accuracy, significantly reducing the initial polygon count. This process yielded over 60,000 fields for training, validation, and testing.

To evaluate the performance of the 2023 ePaddocks product, we collected field boundaries across 67 test tiles, each approximately \textcolor{red}{\(100\,\mathrm{km}^2\)}. These areas are a subset of the test set used in \citep{rs13112197} for the 2019 season. The 2023 annotation was conducted by our CSIRO team.

\begin{figure}[!htbp]
\begin{center}
\includegraphics[width=\columnwidth,height=0.72\textheight,keepaspectratio]{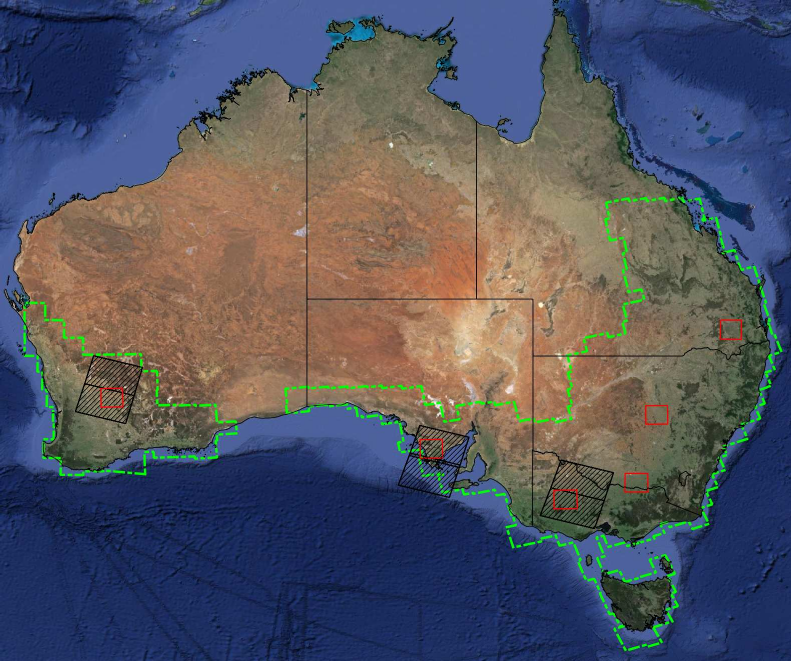}
\end{center}
\caption{Inference and training area selection. Rectangles with red boundaries represent the training tiles (complete S2A scenes) where we have ground truth data. Footprints of Sentinel-1 scenes are represented with hashed black rectangles. The green polygons show the complete ePaddocks inference area. Background imagery  Map data \textcopyright 2023 Google}
\label{Main-TrainingSites}
\end{figure}

\textbf{Sentinel 1}
The Sentinel-1 mission, part of the European Copernicus Program, provides polar-orbiting, all-weather, day-and-night C-band radar imaging for land and ocean services. Sentinel-1A was launched on 3 April 2014, followed by Sentinel-1B on 25 April 2016. The default imaging mode, Interferometric Wide Swath (IW), captures imagery with a 250 km swath at a spatial resolution of 5 m by 20 m. Equipped with dual polarization radars (VV+VH or HH+HV), Sentinel-1 delivers data that is essential for medium- to high-resolution applications, with routine coverage of Australia every 12 days since December 2016, enabling continental-scale land surface mapping and monitoring.

The availability of high-quality dual-pol Sentinel-1 IW data allows the use of dual-polarization Entropy/Alpha decomposition. These decomposition products provide more detailed information about scatterers compared to single-channel backscatter, improving the accuracy of landcover classification and segmentation \citep{cloude1997entropy,8128180}.

For this study, we selected Sentinel-1 dual-pol data in both SLC (Single Look Complex) and Ground Range Detected  (GRD) formats, acquired over the testing sites between May and November 2019. Fig. \ref{Main-TrainingSites} shows the footprints of Sentinel-1 scenes covering the three testing sites. The site in South Australia required one Sentinel-1 scene, while the others required two consecutive scenes to cover the areas of interest. Details of Sentinel-1 observations are provided in Table \ref{Main-TABLE1_ZS}.

The Sentinel-1 SLC data were pre-processed using ESA's Sentinel-1 Toolbox. The following steps were applied to each dual-pol VV+VH SLC product: applying the Precise Orbit File, removing thermal noise, radiometric calibration, S1 TOPS deburst and merge, polarimetric speckle filtering, polarimetric decomposition, multi-looking, and terrain geometric correction (geocoding to a pixel spacing of 0.0002 degrees). After processing, adjacent scenes were mosaiced, and three dates were co-registered, resulting in a data stack ready for further analysis, including VV intensity, VH intensity, and three polarimetric decomposition parameters—alpha ($\alpha$), anisotropy (A), and entropy (H)—using the Eigen-based dual-pol decomposition method \citep{2007ESASP.644E...2C} (see \textcolor{red}{Supplementary Material, Section~\ref{SM-s1_processing_appendix}, for details}). Detailed processing steps are described in \citep{8128180,2022_Mascolo,zhou2023polarimetric}.

\subsubsection{Benchmark datasets}

To showcase transferability across regions, we compare our proposed methodology and benchmark it against established methods and datasets. We trained and evaluated models on the following datasets:

The \textbf{Fields of the World (FoTW)} dataset \citep{kerner2024fields} is a global-scale benchmark specifically designed for agricultural field boundary segmentation. It covers 24 countries across four continents, providing over 70,000 samples paired with semantic and instance segmentation masks aligned to multispectral Sentinel-2 imagery. Each FoTW sample includes two Sentinel-2 L2A cloud-filtered composites captured at contrasting dates within the same year, emphasizing the geographic and morphological diversity of fields, enabling robust cross-region assessments. We utilize FoTW solely for benchmarking purposes, evaluating our model on standardized test splits for selected countries without additional fine-tuning.

The \textbf{PASTIS} dataset \citep{garnot2021panoptic} was initially developed for panoptic segmentation tasks but also serves effectively for crop type semantic segmentation—a task closely related to field boundary delineation. The dataset provides rich Sentinel-2 time series data, ranging from 38 to 61 acquisitions, with partial cloud coverage. This makes PASTIS particularly suitable for evaluating our approach's robustness to varying temporal dynamics and cloud conditions. \textcolor{red}{Both field-boundary extraction and PASTIS crop mapping use the same semantic-segmentation architecture. The target encoding and final output change from \(N_{\mathrm{classes}}=2\) to 19 semantic outputs (18 crop or land-cover classes plus background), while the spatio-temporal backbone is unchanged and is trained for the PASTIS task.}

The \textbf{AI4SmallFarms} dataset \citep{10278130} includes field outlines aligned to Sentinel-2 imagery for three distinct regions: Mekong Delta (Vietnam), Tonlé Sap (Cambodia), and West-Friesland (Netherlands). We specifically focus on the Vietnam  and Cambodia subsets, as they present significant challenges due to their high cloud frequency—forcing reliance on sparse clear-sky observations—and highly fragmented landscapes with sub-hectare paddies separated by narrow earth dykes.

Since the AI4SmallFarms dataset is already included in FoTW using standard Sentinel-2 images at 10m resolution, we opted here to explore an alternative approach. Specifically, we assessed the algorithm's performance on small farm fields using Sentinel-1 data combined with super-resolved Sentinel-2 imagery. The original S2 cloud-free composite provided by AI4SmallFarms was initially upscaled fourfold using the open-source Super Resolution Sentinel-2 (LDSR-S2) model \citep{10887321}. We recreated the ground truth for field extents, boundaries, and distance transforms \citep{DIAKOGIANNIS202094,WALDNER2020111741,rs13112197} from the original reference polygons at this higher resolution. We emphasize that the super-resolution method applied is generative and may introduce minor inaccuracies specifically in the imagery rather than the ground truth.

Additionally, we enriched the dataset by complementing the Sentinel-2 imagery with Sentinel-1 time series, following procedures similar to those used for Australian tiles. Sentinel-1 IW images, in both GRD and SLC formats, acquired from 1 January to 28 February 2021 (descending orbit), consistent with the original optical image acquisition period \citep{10278130}, were downloaded from the ASF data portal\footnote{ \href{https://search.asf.alaska.edu/}{https://search.asf.alaska.edu/}.}. Sentinel-1 data were processed into analysis-ready products using standard SNAP workflows on CSIRO’s HPC systems, resulting in a pixel spacing of 1e-4 degrees (EPSG 4326) and including normalized radar backscatter (Gamma0 VV and VH) and polarimetric decomposition parameters (alpha, Anisotropy, and Entropy), with radiometric terrain flattening and geometric terrain corrections applied using the 30m Copernicus global DEM.

\begin{table*}[!tp]
\centering
\fontsize{8}{10}\selectfont
\caption{Acquisition dates and product numbers of Sentinel-1 observations for 3 sites}
\label{Main-TABLE1_ZS}
\begin{tabular}{|c|c|c|c|c|c|}
\hline
\textbf{State} & \textbf{\# Scenes} & \textbf{Acquisition Period (2019)} & \textbf{\# Acquisition Dates} & \multicolumn{2}{c|}{\textbf{Sentinel-1 Products}} \\ \cline{5-6} 
 &  &  &  & \textbf{GRD} & \textbf{SLC} \\ \hline
WA (50JPL)  & 2 & 04/30-11/20 & 17 & 34 & 34 \\ \hline
SA (53HNC)  & 1 & 05/09-11/29 & 18 & 18 & 18 \\ \hline
VIC (54HXE)  & 2 & 05/01-11/21 & 18 & 36 & 36 \\ \hline
\end{tabular}
\end{table*}

\subsection{Model architecture}
This section describes the adaptation of the base model PTAViT developed for 2D inputs \citep{DIAKOGIANNIS202444}, for the case of time series of imagery. This can be a single type of input (S2 or S1) or a data fusion methodology (S1\&S2) with cross (relative) attention. Both of these approaches share the same building block, the PTAViT3D Stage and utilize 3D hybrid convolutions.

\subsubsection{Patch Tanimoto attention 3D}

Building upon our previously defined 2D attention method \citep{DIAKOGIANNIS202444}, we extend it here to handle the temporal dimension explicitly. Given that we have a limited number of time observations, we retain all individual time instances rather than grouping them into blocks. This allows a clear and detailed comparison at each individual time step. For completeness, we briefly discuss the time partitioning attention below, but readers interested in further details are encouraged to consult \citep{DIAKOGIANNIS202444}\footnote{\href{https://github.com/feevos/ssg2}{https://github.com/feevos/ssg2}. }.

\begin{figure}[!htbp]
  \centering
  \includegraphics[width=0.8\linewidth,height=0.72\textheight,keepaspectratio]{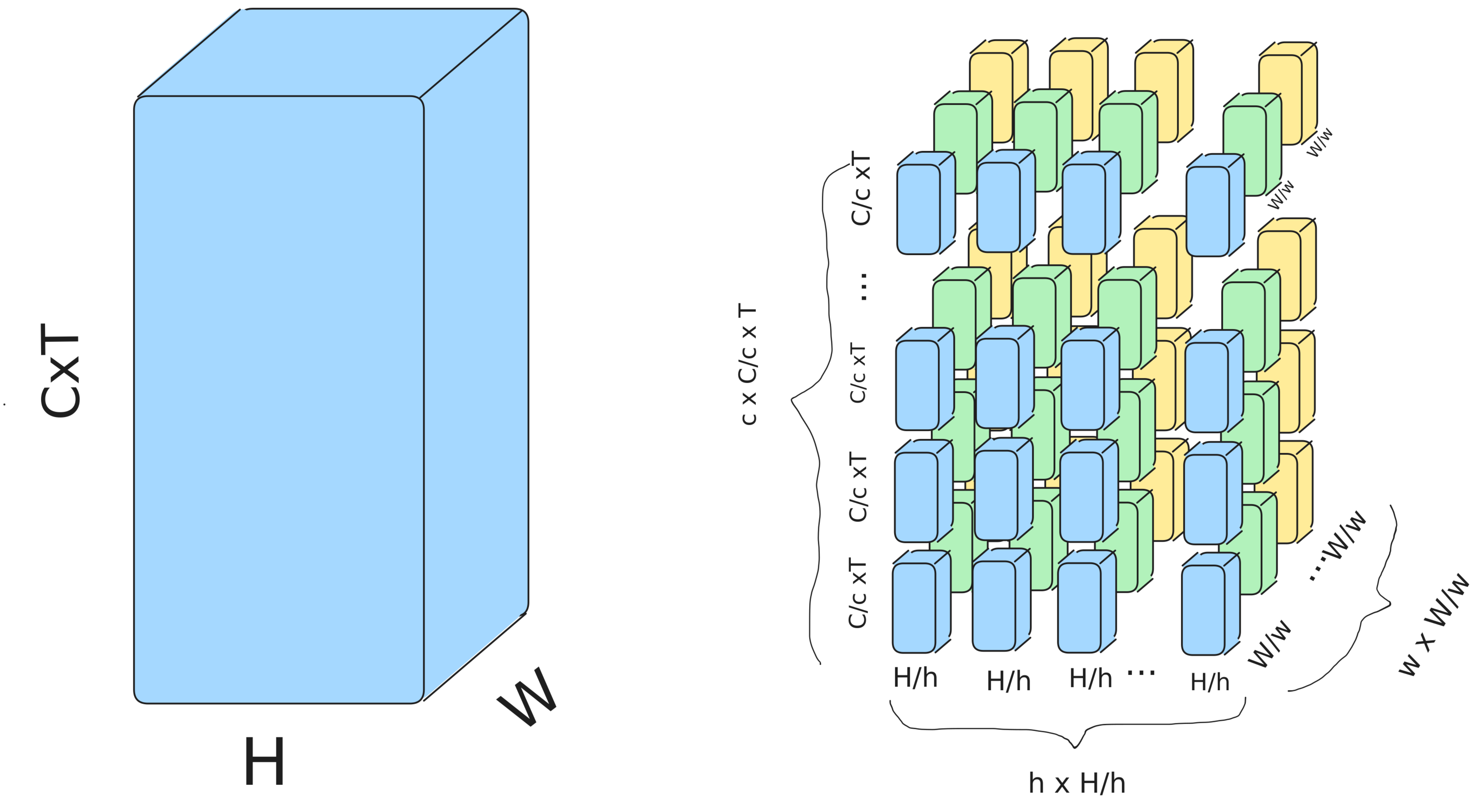}
  \caption{Schematic partitioning of a $C\times T\times H\times W$ tensor in $c \times T \times h \times w$ patches. Each patch has dimensions $(C/c) \times (H/h) \times (W/w)$, where $C$ are the channels of the original 3D tensor, $T$ the time, $H, W$ the height and width respectively. $c,h,w$ are number of partitions in each dimension.
  }
  \label{Main-3d_attention_split}
\end{figure}
Figure \ref{Main-3d_attention_split} illustrates our method of partitioning a tensor with dimensions $C \times T \times H \times W$ (channels, time, height, width) into a set of  $c \times T \times h \times w$ partitions. Each patch in this partition thus has dimensions $(C/c) \times (H/h) \times (W/w)$, where $c, T, h, w$ represent the number of partitions in each respective dimension, and $c\cdot T\cdot h\cdot w$ is their total number.

The Tanimoto similarity is defined as:
\begin{equation}
\label{Main-Tanimoto_similarity}
\mathcal{T}(\mathbf{q}, \mathbf{k}) =
\begin{cases}
	 \frac{ \langle \mathbf{q} | \mathbf{ k}  \rangle}{\langle \mathbf  { q}  | \mathbf { q }  \rangle + \langle \mathbf {k} | \mathbf { k } \rangle - \langle \mathbf { q }  | \mathbf { k}  \rangle   } & \mathbf{q} \neq \mathbf{0} \;\text{or}\; \mathbf{k} \neq \mathbf{0}   
\\
\quad \quad 0 & \mathbf{q} = \mathbf{k} = \mathbf{0}  
\end{cases}
\end{equation}
Here, $\langle \mathbf{q} | \mathbf{k} \rangle$ denotes a form of inner product (tensor contraction) performed along selected dimensions of tensors $\mathbf{q}$ and $\mathbf{k}$. Initially, both query ($\mathbf{q}$), key ($\mathbf{k}$), and value ($\mathbf{v}$) tensors have the shape $C \times T \times H \times W$. 
These are reshaped into smaller patches:
\begin{align*}
q_{ C\times T\times H\times W} &\to q_{ c\times T\times h \times w \; \times \; (C/c) \times (H/h) \times (W/w)}\\
k_{C\times T\times H\times W} &\to k_{ c\times T\times h \times w \; \times \;  (C/c) \times (H/h) \times (W/w)} \\
v_{C\times T\times H\times W} &\to v_{ c\times T\times h \times w \; \times \;  (C/c) \times (H/h) \times (W/w)}. 
\end{align*} 
 These smaller patches (see Fig. \ref{Main-3d_attention_split}) serve as the building blocks for calculating Tanimoto similarity. Specifically, the similarity is calculated by contracting the tensor along the channel, height, and width dimensions (indices $r,s,t$) of these patches:
\begin{align}
	\langle \mathbf{q} | \mathbf{ k}  \rangle &= \sum_{rst} q_{\textcolor{cyan}{cFhw} rst}   k_{\textcolor{magenta}{kTlm} rst}  \equiv \langle \mathbf{q} | \mathbf{ k} \rangle {} _{\textcolor{cyan}{cFhw}\textcolor{magenta}{kTlm}   }  \\
	\langle \mathbf{q} | \mathbf{ q}  \rangle &= \sum_{rst} q_{\textcolor{cyan}{cFhw} rst}   q_{\textcolor{cyan}{cFhw} rst}  \equiv \langle \mathbf{q} | \mathbf{ q} \rangle {} _{\textcolor{cyan}{cFhw}   } \\	
	\langle \mathbf{k} | \mathbf{ k}  \rangle &= \sum_{rst} k_{\textcolor{magenta}{kTlm} rst}   k_{\textcolor{magenta}{kTlm} rst}  \equiv \langle \mathbf{k} | \mathbf{ k} \rangle {} _{\textcolor{magenta}{kTlm}.  } 
\end{align}
The resulting Tanimoto similarity of $q$ and $k$, has dimensions $\mathcal{T}(q, k)_{ \textcolor{cyan}{cFhw}\textcolor{magenta}{kTlm}}$. Here $\textcolor{cyan}{cFhw}$ or $\textcolor{magenta}{kTlm}$ correspond to dimensions of channel partitions, time, height partitions and width partitions. The information of the 8-D coordinate similarity map, is contracted to 4D in two possible ways. We can either use a \texttt{Linear} matrix multiplication or, given that we want to increase the number of time observations during inference, to just use the mean value (reduces memory footprint) along the query dimensions: 
\begin{align}
\label{Main-similarity_coordinate_map_weighted}
\tilde{\mathcal{T}}(q , k)_{\textcolor{magenta}{kTlm}} &= \sum_{\textcolor{cyan}{cFhw}} \mathcal{T}(q , k)_{ \textcolor{cyan}{cFhw}\textcolor{magenta}{kTlm}} W_{\textcolor{cyan}{cFhw}}\\
\tilde{\mathcal{T}}(q , k)_{\textcolor{magenta}{kTlm}} &= \frac{1}{N}\sum_{\textcolor{cyan}{cFhw}} \mathcal{T}(q , k)_{ \textcolor{cyan}{cFhw}\textcolor{magenta}{kTlm}},
\end{align}
 where $N=\dim (\textcolor{cyan}{cFhw})$ is the product of the first four dimensions of the query tensor, $\mathbf{q}$, and $W_{\textcolor{cyan}{cFhw}}$ a weight matrix for the case of Linear matrix multiplication. 

It should be noted that this similarity can be given causal structure, by multiplying element wise with a mask lower triangular matrix, $m_{\textcolor{cyan}{F} \textcolor{magenta}{T}}$ along the time dimensions, i.e. 
\begin{equation}
\mathcal{T}_{\text{csl}}(q, k)_{ \textcolor{cyan}{cFhw}\textcolor{magenta}{kTlm}} = \mathcal{T}(q, k)_{ \textcolor{cyan}{cFhw}\textcolor{magenta}{kTlm}} m_{\textcolor{cyan}{F} \textcolor{magenta}{T}}
\end{equation} 
where
\begin{equation}
\label{Main-causal_mask}
m_{\textcolor{cyan}{F} \textcolor{magenta}{T}} = 
\begin{cases}
1, \text{if} \; F \leq T \\
0
\end{cases}
\end{equation}
and this is useful for forecasting applications of our methodology. 

Finally, the attention layer is produced by element-wise multiplication of this similarity matrix, taking into account the time slot, with the values tensor, $v$, and re-arranging the patches to the original tensor shape, subject to the activation \texttt{d2s}:
\begin{equation}
\label{Main-Attention_map}
\mathcal{A}(q,k,v)_{\textcolor{magenta}{kTlm}rst} = \texttt{d2s}\left( \tilde{\mathcal{T}}(q , k)_{\textcolor{magenta}{kTlm}} \odot v_{\textcolor{magenta}{kTlm}rst} \right)
\end{equation}
The Attention map, $\mathcal{A}(q,k,v)$, upon reshaping, has dimensionality $C\times T\times H\times W$, and encapsulates all time, spatial and channel correlations that exist between query and key tensors to a patch resolution level of $c\times h\times w$, i.e.  $\mathcal{A}(q,k,v) \in \mathfrak{R}^{C\times T\times H\times W}$.  Hereafter, we will refer to this attention map as Patch Tanimoto Attention 3D (PTA3D).

\subsubsection{Limitations of separable channel–spatial–temporal similarity}
\label{Main-sssec:3d-decomp-limits}

\begin{figure}[!htbp]
  \centering
  \includegraphics[width=0.8\linewidth,height=0.72\textheight,keepaspectratio]{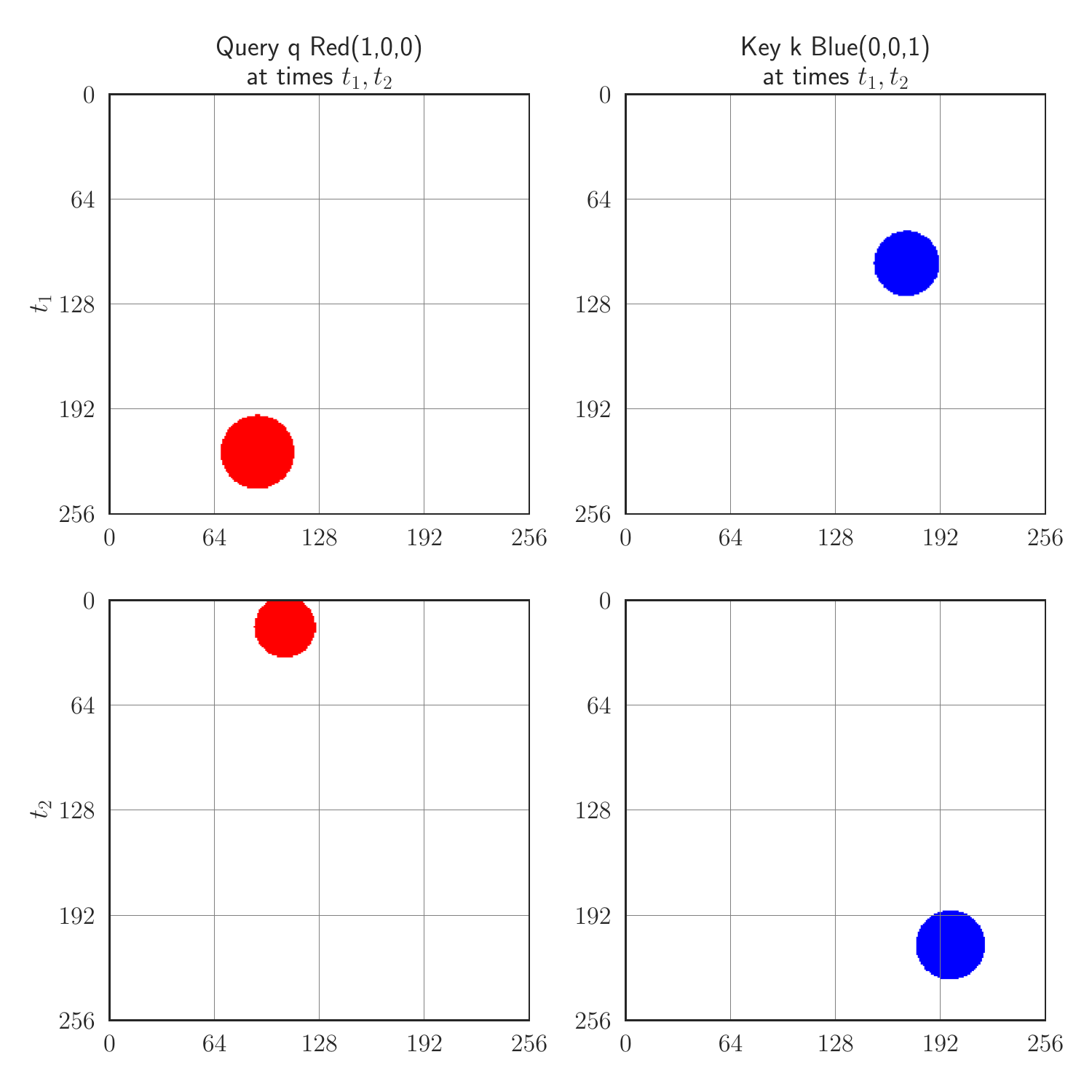}
  \caption{Query, \textcolor{tensorred}{$\mathbf{q}$}, and key, \textcolor{blue}{$\mathbf{k}$}, similarity between a red $(1,0,0)$ and a blue $(0,0,1)$ disjoint
disks at different times $t_1$, $t_2$. The element-
wise multiplication that is used for time-channel-space separable attention results in 0 for all channels ($=(1\cdot0,0\cdot0,0\cdot 1)$).  
  }
  \label{Main-fig:3d-decomp}
\end{figure}

In many Satellite Image Time-Series vision  models the full $4$-D similarity between query and key,
$\mathbf q,\mathbf k\in \mathfrak{R}^{C\times T\times H\times W}$, is approximated by
three independent inner-products:
\begin{align}
\langle \mathbf q | \mathbf k \rangle_{\mathrm{channel}}
&=\sum_{j,k,t}q_{\textcolor{cyan}{i} t j k}\;k_{\textcolor{cyan}{i} t j k}\;\in\;\mathfrak{R}^C,\\
\langle\mathbf q | \mathbf k\rangle_{\mathrm{space}}
&=\sum_{i,t}q_{it\textcolor{cyan}{j k}}\;k_{it \textcolor{cyan}{j k }}\;\in\; \mathfrak{R}^{H\times W},\\
\langle\mathbf q | \mathbf k\rangle_{\mathrm{time}}
&=\sum_{i,j,k}q_{i \textcolor{cyan}{t} j k}\;k_{i \textcolor{cyan}{t} j k}\;\in\;\mathfrak{R}^T,
\end{align}
and then combined via broadcasting. However, as illustrated in Fig.~\ref{Main-fig:3d-decomp},
when two patterns occupy disjoint channels  or time steps all of these separable terms vanish, even if a genuine spatiotemporal correspondence exists. In the example in Fig.~\ref{Main-fig:3d-decomp} we plot two red $(1,0,0)$ discs and two blue $(0,0,1)$ discs that are spatio-temporally separated.

\begin{figure}[!htbp]
  \centering
  \includegraphics[width=0.8\linewidth,height=0.72\textheight,keepaspectratio]{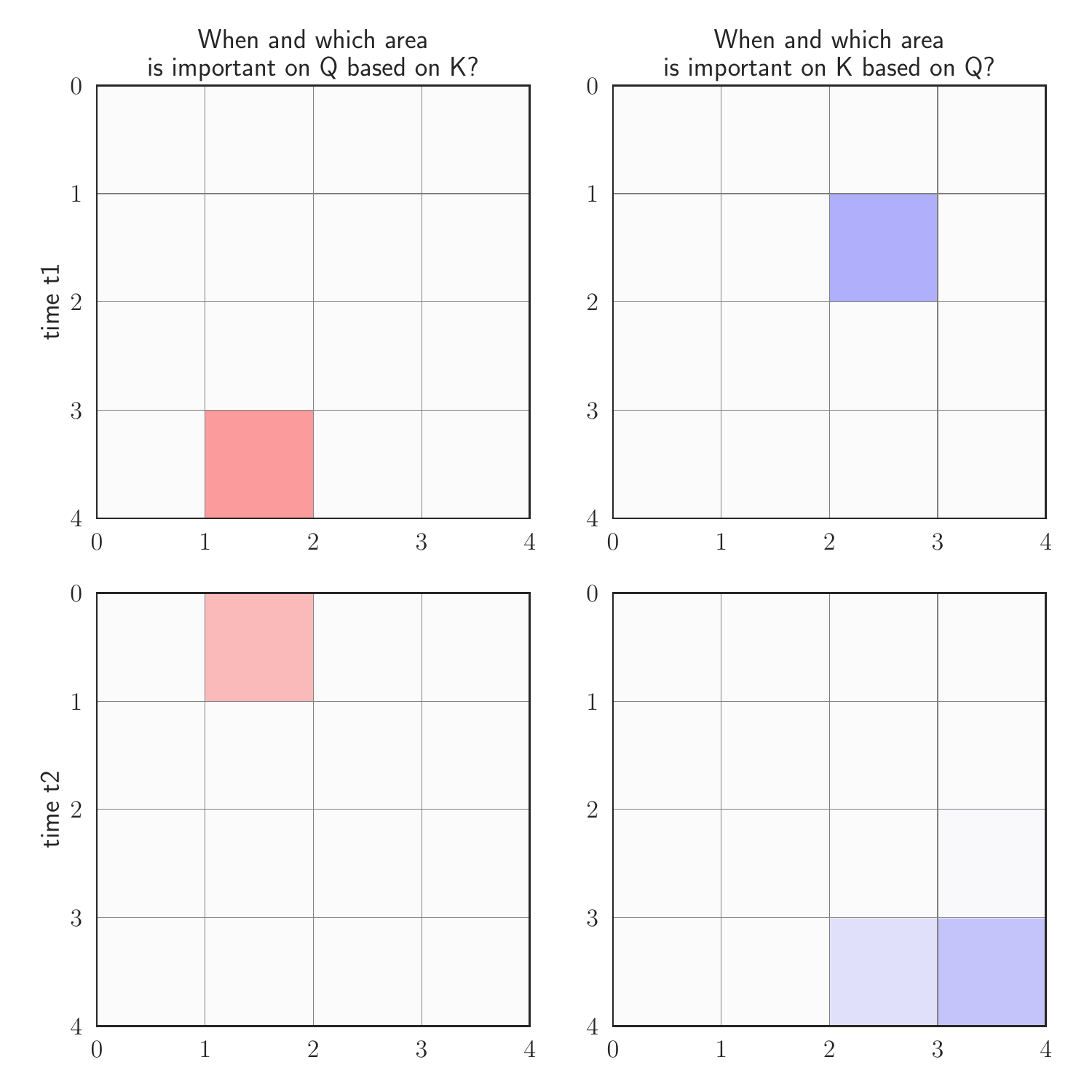}
  \caption{PTA3D emphasis maps for the example in Fig.~\ref{Main-fig:3d-decomp}, correctly identifying the matching patch across time space and channels.}
  \label{Main-fig:pta3d-emphasis}
\end{figure}

By contrast, our \emph{Patch Tanimoto Attention 3D} (PTA3D) preserves cross-dimensional interactions.  We first reshape the tensors (see also Fig. \ref{Main-3d_attention_split})
\[
  \mathbf q\in\mathbb R^{C\times T\times H\times W}
  \;\longrightarrow\;
  \mathbf q\in\mathbb R^{\,c\times T\times h\times w\,\times\,(C/c)\times(H/h)\times(W/w)}\,,
\]
(and similarly for $\mathbf k,\mathbf v$), so that each patch spans all four original axes.  We then compute the Tanimoto similarity of each query–key patch pair via Eq.~\eqref{Main-Tanimoto_similarity}, yielding an 8-D tensor
\[
  \mathcal T(\mathbf q,\mathbf k)\in\mathbb R^{\,c\times T\times h\times w\times c\times T\times h\times w}\,,
\]
and finally contract the paired patch-coordinates (either by a small linear projection or by simple averaging).  This retains true spatiotemporal matches without blowing up memory, as Fig.~\ref{Main-fig:pta3d-emphasis} shows.
The result (Fig.~\ref{Main-fig:pta3d-emphasis}) is a full $c\times T\times h\times w$ attention map that faithfully encodes
all channel, spatial and temporal correlations at patch resolution, yet remains
efficient on modern GPUs.

\begin{figure*}[!tp]
    \centering
    \subfigure[PTA-ViT3D Stage]{
    \centering
        \includegraphics[clip, trim=0.25cm 0.40cm 0.25cm 0.1cm,width=0.25\textwidth,height=0.72\textheight,keepaspectratio]{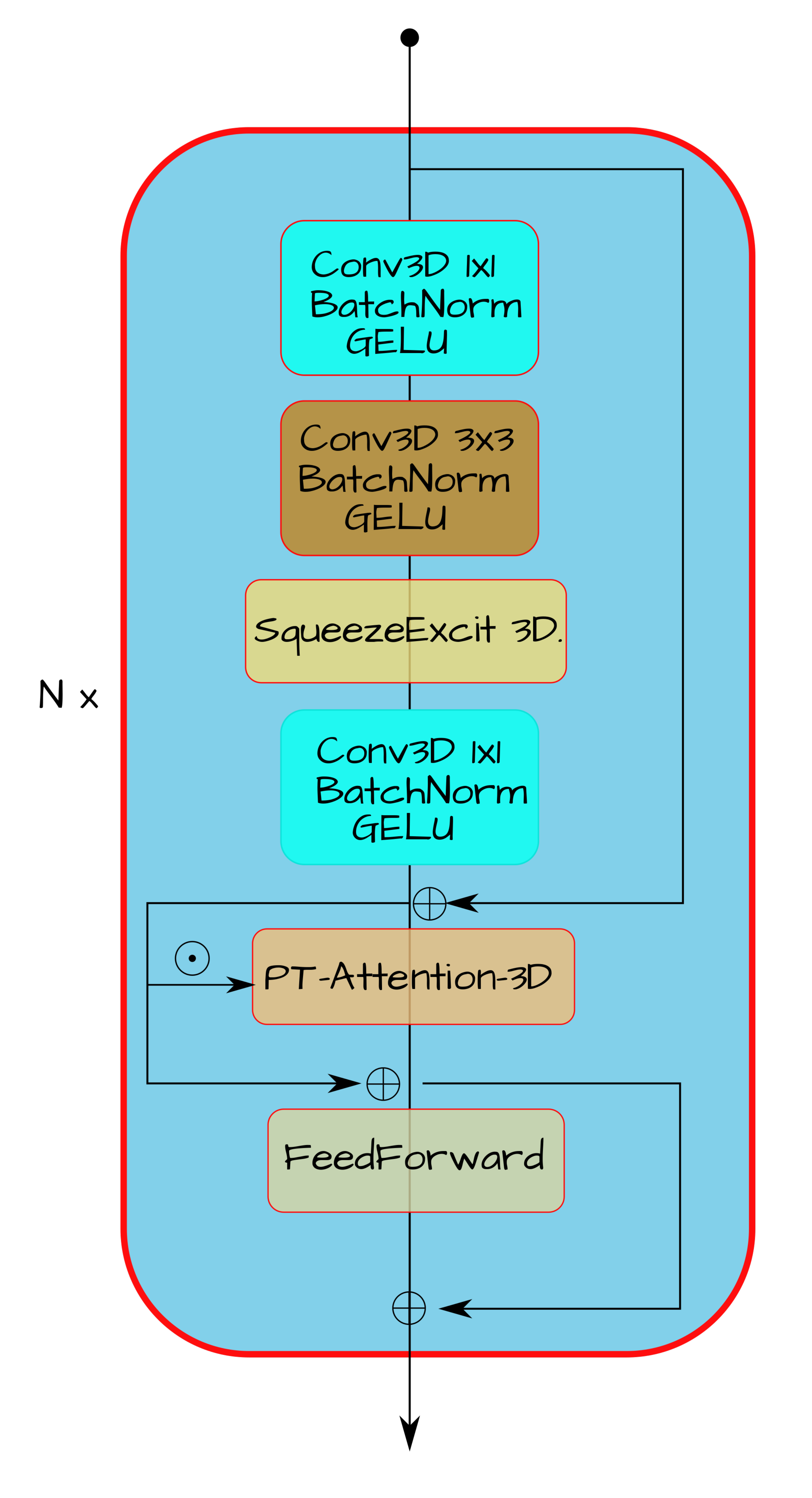}
        \label{Main-ptvitstage:suba}
    }
    \subfigure[U-Net3D macro-topology]{
        \includegraphics[clip, trim=0.0cm 0.20cm 0.25cm 0.1cm,width=0.5\textwidth,height=0.72\textheight,keepaspectratio]{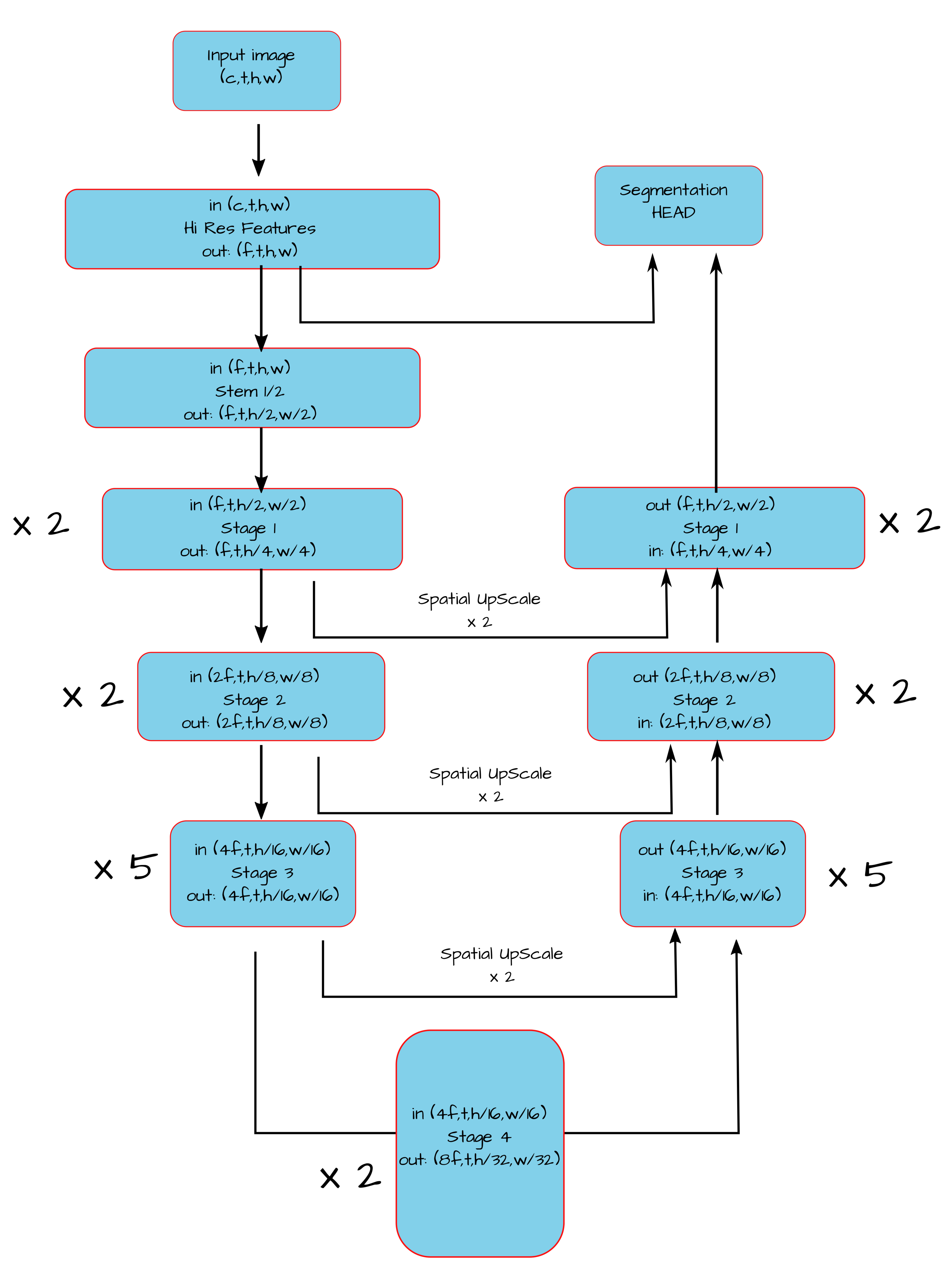}
        \label{Main-ptvitstage:subb}
    }
    \caption{(a) Patch Tanimoto Attention ViT Stage. The \texttt{Stage} comprises a sequence of MBConv blocks, followed by Squeeze Excitation, Patch Tanimoto Attention and a FeedForward network. The architecture is defined by the number of these building blocks that are repeated. (b) The UNet3D macro-topology is symmetric in its encoder and decoder, which is reflected by the same number of stages used in the corresponding points of the Encoder and Decoder layers. In the Figure we show the Tiny configuration of [2,2,5,2,5,2,2] network.} 
    \label{Main-ptvitstage}
\end{figure*}

PTA3D was designed with three main motivations in mind. First, we wanted to introduce broad spatial context from the first stage of the network by allowing different windows to interact directly, rather than relying only on progressively deeper convolutional receptive fields or repeated local-window mixing. Second, we moved away from softmax because, in our earlier experiments, sigmoid-style gating often behaved better for this task, suggesting that emphasizing multiple relevant regions was more useful than enforcing a strict normalized ranking across all interactions. This choice also made it possible to avoid the standard dense attention application: after similarity compaction, PTA3D applies the resulting emphasis map to the values tensor, $\mathbf{v}$,  through element-wise multiplication, which simplifies the operation and reduces memory usage. Third, we adopted a Tanimoto-style similarity because it provides a bounded measure with a natural spatial interpretation for comparing patch contents. While this choice is not necessarily optimal purely in terms of memory efficiency, it reflects the original design goal of combining early non-local context with a simple and spatially meaningful similarity. \textcolor{red}{A more detailed comparison with standard windowed attention is provided within the controlled ablation study (Supplementary Material, Subsection~\ref{SM-supp:pta-vs-window-attention}).}

\subsubsection{Feature extraction units}

Following our previous work on the PTAViT architecture, we replace the 2D convolutions and patch Tanimoto Attention with their 3D equivalents presented here. These units build on  MaXViT \citep{tu2022maxvit} by replacing the 2 blocks of attention with a single Attention block.

\subsubsection{Single Input time series: UNet macro-topology}
For modelling single input (S1 or S2) time series of observations, we adopt a PTAViT 2D model  to account for 3D structure, by using the PTAViT3D Stages.  Details for both the 3D Stage  (Fig. \ref{Main-ptvitstage:suba}) and UNet macrotopology  (Fig. \ref{Main-ptvitstage:subb}) of this implementation can be seen in  Figure \ref{Main-ptvitstage}. We note that in this architecture, the encoder and decoder are symmetric, as this is depicted by the same number of stages used in the encoder and the decoder. Crucially, the Time dimension is kept along all of the blocks of the module, i.e. there is no pooling or dimensionality reduction along the time dimension. All exchange of information through various time snapshots is achieved with the 3D version of the attention, PTA3D. This is in contrast with the alteration of the spatial dimension that is gradually reduced in the encoder and increased again in the decoder. 

At the end of the 3D features created, we attach another PTAViT3D stage before compacting the time dimension to create the final filtered multitasking segmentation predictions. The multitasking segmentation head  is the standard we have used in our previous work developed in \citep{DIAKOGIANNIS202094} and used in various semantic segmentation publications \citep{WALDNER2020111741,rs13183707,rs13112197}. This head predicts in a causal way the extent, boundaries and distance transform of each field (or class for the case of multiple classes segmentation).

\subsubsection{S2\&S1 feature fusion architecture}

\begin{figure*}[!tp]
\centering
\includegraphics[clip, trim=0.25cm 0.40cm 0.1cm 0.1cm,width=\textwidth,height=0.72\textheight,keepaspectratio]{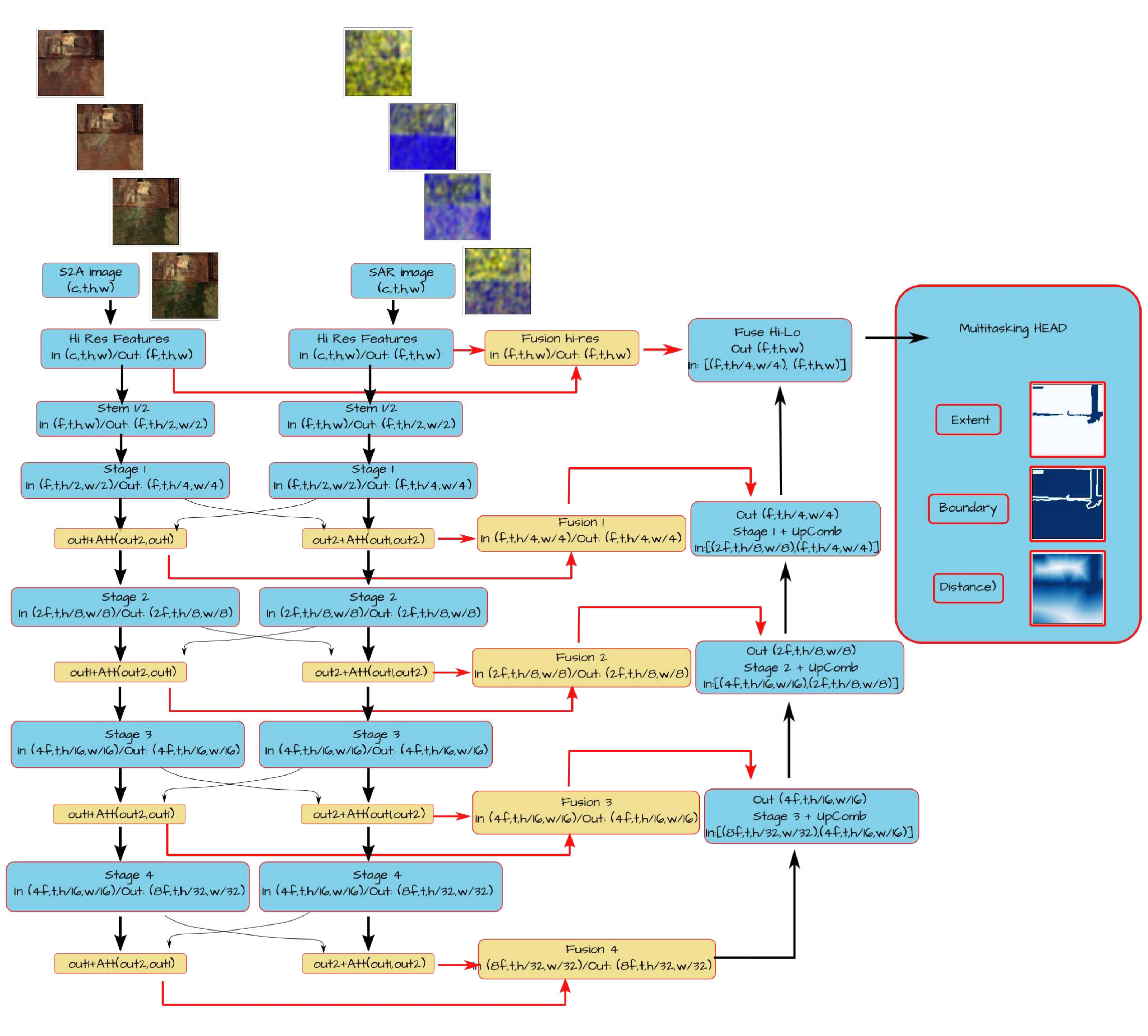}
\caption{S2 (S2A) and S1 (SAR) Time Series Fusion architecture PTAViT3D-CA.} 
\label{Main-mantis_v2_base_arch}
\end{figure*}

The fusion process is achieved with a dual encoder (without shared weights) single decoder architecture utilizing the 3D cross attention we present here. This is similar to the SSG2 base model architecture \citep[][see also \citealt{rs13183707}]{DIAKOGIANNIS202444}, however a crucial distinction is that now all time snapshots are compared with all other time instances. That is, the cross attention ``sees'' all  relative time instances in order to assign emphasis (weighting) to a particular spatio-temporal location. This expands the scope of our previous work, that was developed for a set of observations where order along the sequence dimension was irrelevant.  
The creation of the segmentation masks from the extracted 3D features is similar to the UNet macrotopology as described above. 
Details can be seen in Fig \ref{Main-mantis_v2_base_arch}. The hyper parameter configuration we follow is the same as in SSG2, 4 stages ([2,2,5,2]) for the encoder, and 3 for the decoder ([5,2,2]), starting from 96 initial features, which are doubled in every subsequent stage of the encoder architecture, and then halved for each subsequent stage of the decoder.

To keep the manuscript self-contained while avoiding extensive implementation detail\footnote{The fusion implementation takes place in code file \href{https://github.com/feevos/tfcl/blob/master/models/ptavit3dca/ptavit3dca_dn_features.py}{ptavit3dca\_dn\_features.py} of our \href{https://github.com/feevos/tfcl}{github} repository. }, \textcolor{red}{Supplementary Material, Section~\ref{SM-alg:s1s2_fusion}} summarizes the multimodal fusion strategy used in PTAViT3D-CA. Sentinel-1 and Sentinel-2 sequences are first projected into a shared embedding space, then fused through cross-attention at multiple encoder scales, followed by a scale-wise fusion operation that produces the fused skip hierarchy used by the decoder. This design allows the model to learn non-linear correspondences between SAR and optical modalities directly from data, without imposing an explicit mapping between the two sensor domains.

\subsection{Loss function}
For the loss function we use for all layers the multitasking approach we developed in \citep{DIAKOGIANNIS202094}, i.e. the Tanimoto with complement: 
\begin{equation}
\mathcal{L}_{\mathcal{T}}(\mathbf{p},\mathbf{l})= 1 -\frac{1}{2} \biggl(\mathcal{T}(\mathbf{p},\mathbf{l}) + \mathcal{T}(1-\mathbf{p},1-\mathbf{l})\biggr)
\end{equation}
 where it is applied independently to the predictions of the extent, $\mathbf{e}$, boundaries, $\mathbf{b}$ and distance transform, $\mathbf{d}$ ground truth annotations:
\begin{equation}
\label{Main-TanimotoWthDual}
\mathcal{L} = \frac{1}{3}\left(
\mathcal{L}_{\mathcal{T}}(\hat{\mathbf{e}},\mathbf{e}) + \mathcal{L}_{\mathcal{T}}(\hat{\mathbf{b}},\mathbf{b})+\mathcal{L}_{\mathcal{T}}(\hat{\mathbf{d}},\mathbf{d})
\right)
\end{equation}

\subsection{Data pre-processing and augmentation for deep learning pipeline}

For the S1 data, the initial imperative step involves extracting areas corresponding to the S2 scenes and reprojecting them to match the S2 pixel resolution. This alignment is essential because the ground truth labels are tailored to the spatial resolution of the S2 scenes.

The preprocessing of S1 data is conducted using the function \texttt{transform\_s1} (\textcolor{red}{Supplementary Material, Section~\ref{SM-s1transformation_section}}). This function adjusts the first band, representing the incidence angle \(\alpha\), by converting its values from degrees to radians to standardize its range. Furthermore, it applies a symmetric logarithmic transformation to the VH and VV polarization bands, effectively expanding the dynamic range of these values and making them more suitable for neural network processing. This transformation is particularly beneficial for accommodating the stark contrasts between high-magnitude values (VH and VV) and lower-magnitude values (\(\alpha\), anisotropy, and entropy), thus enhancing stability during the learning phase. Subsequently, the data are standardized to achieve zero mean and unit variance.

The S2 dataset undergoes a similar standardization process to ensure zero mean and unit variance across all four bands. Training patch sizes were selected per dataset based on their spatial characteristics and practical GPU memory constraints. Specifically, patches of size \(256 \times 256\) pixels were used for Fields of the World (FoTW) and AI4SmallFarms, while patches of size \(128 \times 128\) pixels were used for PASTIS and ePaddocks. In all cases, patches were extracted with a stride equal to the patch size. This strategy ensured compatibility with the available GPU memory while providing an appropriate spatial context for each dataset.

\textcolor{red}{Acquisition dates determine the chronological order of each input sequence, but neither the dates nor elapsed inter-acquisition times are supplied as additional features. The architecture accepts non-uniformly spaced acquisitions as an ordered sequence; the acquisition window is selected to span the relevant cropping cycle.}

To further enhance the model's generalization ability, a comprehensive suite of data augmentation techniques is employed. These techniques include equal-probability selections from a set of geometric transformations. The array of transformations comprises horizontal and vertical flips, elastic deformations, grid distortions up to a limit of 0.4, and shift-scale-rotate operations with a shift limit of 0.25, a scale range from 0.75 to 1.25, and full rotational freedom up to 180 degrees.

This augmentation strategy enriches the training dataset, thereby bolstering the robustness of the deep learning model. These transformations are implemented using the Albumentations library \citep{info11020125}.

\subsection{Post-processing: from raster predictions to shapefiles}
\label{Main-post_processing}

Let \( e \) and \( b \) represent the predicted extent and boundaries layers, respectively. These layers capture essential information for instance segmentation, with various algorithms transitioning from semantic to instance segmentation (e.g., watershed segmentation \citep{WALDNER2020111741}, hierarchical segmentation \citep{rs13112197}). Here, we present a practical and easy-to-implement approach that utilizes the predicted extent and boundary layers.

The first step involves refining the extent layer using two distinct thresholds: one for the extent, \( t_e \), and another for the boundaries, \( t_b \). Initially, the predicted boundaries, \( b \), are thresholded with $t_b$ and undergo morphological thinning to reduce their thickness, followed by a dilation to ensure a single-pixel width. This thinned boundary layer provides a fine segmentation that ensures clear separation between neighboring fields.

Next, the extent layer, \( e \), is combined with the inverted thinned boundaries, \( 1 - b \), to produce a refined mask. This combination is achieved by taking the fuzzy set intersection of \( e \) and \( 1 - b \), realized by multiplying the extent with the complement of the boundaries, \( e \cap \overline{b} \), and then thresholding the result based on the extent threshold, \( t_e \). This process  yields a binary mask that highlights the extents of the fields without the influence of boundaries, ensuring accurate delineation.

We then utilize the \texttt{rasterio} library \citep{gillies_2019} to convert this refined raster mask into polygon shapefiles. The refined extent mask is passed to the \texttt{rasterio} \texttt{features.shapes} function, which identifies contiguous regions in the binary mask and converts them into geometric shapes. These shapes are then converted into a \texttt{GeoDataFrame}, maintaining the correct spatial reference system (CRS) from the input metadata. The resulting \texttt{GeoDataFrame} can be saved as a shapefile for further analysis or visualization (\textcolor{red}{Supplementary Material, Section~\ref{SM-alg:raster2polygons}}). This approach leverages the strengths of the predicted extent and boundary layers to accurately delineate field boundaries, providing a straightforward and efficient method for converting raster predictions into vector formats suitable for geospatial applications. The performance of this approach depends on the choice of the extent and boundary thresholds used, that we evaluate using a multi-objective optimization routine. A good value of thresholds, based on the ePaddocks dataset, was found to be $t_b\approx 0.2$ and $t_e \approx 0.4$. We used these values fixed for all datasets in our experiments.

\subsection{Evaluation metrics}
\label{Main-evaluation_methods_all}

In order to quantify the performance of our algorithms, we evaluate metrics for the task of semantic segmentation, which is the primary output of the PTAViT3D algorithm, as well as metrics that quantify the agreement between the boundaries (i.e. the polygon lines) that are created as a post processing step. 
Not all metrics are reported for all datasets, because the experiments serve different objectives.
For the public benchmark datasets (PASTIS, FoTW, AI4SmallFarms), we report the metrics used by the corresponding benchmark protocols to enable direct comparison with prior work. For the  operational national-scale polygon product of Australia (ePaddocks) we additionally report polygon-level geometric metrics after raster-to-polygon conversion to quantify the fidelity of the final vector output. Table~\ref{Main-tab:exp-matrix} summarizes the role of each dataset in the study.

All of the evaluation metrics were based on calculation first of the confusion matrix with the use of the package \texttt{PyCM} \citep{Haghighi2018-2}. For the evaluation of the performance,   we use the metrics described in Sections  \ref{Main-segmentation_metrics_section}, \ref{Main-segmentation_for_fdr_metrics_section} and \ref{Main-boundary_metrics_section}.

\subsubsection{Segmentation Evaluation Metrics}
\label{Main-segmentation_metrics_section}

The metrics used for the quantification of semantic segmentation are the Matthews Correlation Coefficient (MCC),  Intersection over Union, and Oversegmentation (FDR) and Undersegmentation (FOR) rates.

\paragraph{Matthews Correlation Coefficient}
 The Matthews Correlation Coefficient  \citep{MATTHEWS1975442} in its multiclass version \citep{GORODKIN2004367}, defined by a $K\times K$ confusion matrix, $C_{ij}$, where $K$ is the number of classes, is given by:
\begin{equation}
\label{Main-mcc_multiclass}
MCC = \frac{c s - \sum_{i=1}^K p_i t_i }{\sqrt{(s^2-\sum_{i=1}^K p_i^2) (s^2-\sum_{i=1}^K t_i^2)}}
\end{equation}
where 
\begin{align*}
t_i &= \sum_{j=1}^K C_{ji} \; \text{ represents the actual occurrence count of class $i$,}\\
p_i &= \sum_{j=1}^K C_{ij}\; \text{indicates how many times class $i$ was predicted,}\\
c   &= \sum_{i=1}^K C_{ii} \; \text{is the total number of correct predictions,} \\
s   &= \sum_{i=1}^K \sum_{j=1}^K C_{ij} \; \text{is the overall sample count.}
\end{align*}
MCC ranges from [-1,1] in the binary case, where a value of 1 suggests maximum performance. For the multiclass case the lower value $\in$ [-1,0].

\paragraph{Intersection over Union}

For two one-hot encoded binary predictions $P$ and $L$  of shape $N\times H  \times W$, where $N$ is the number of classes, $H$ and $W$ the height and width respectively, we define the mean Intersection over Union, via the (fuzzy) set operations of intersection and union as: 
\begin{equation}
\text{mIoU} = \frac{1}{N}\sum_{i=1}^N \frac{\sum_{j,k}\text{min}(P_{ijk},L_{ijk})}{\sum_{j,k}\text{max}(P_{ijk},L_{ijk})}
\end{equation}

For binary one dimensional masks, we use the following definition:
\begin{equation}
\text{IoU} = \frac{TP}{TP+FP+FN}
\end{equation}

\subsubsection{Oversegmentation and Undersegmentation Rates}
\label{Main-segmentation_for_fdr_metrics_section}

In the evaluation of segmentation algorithms, it is crucial to quantify the extent to which the predicted segments either overestimate or underestimate the actual segments. Two metrics commonly used for this purpose are the False Discovery Rate (FDR) \citep{10.1111/j.2517-6161.1995.tb02031.x} and the False Omission Rate  \citep[FOR,][]{6b87510ce7324df69116f8395644ed77}.

\paragraph{False Discovery Rate (FDR)}

The False Discovery Rate (FDR) measures the proportion of false positives among the predicted positives. In the context of segmentation, it quantifies the extent of oversegmentation, i.e., the degree to which the predicted segmentation includes areas outside the ground truth segmentation. Mathematically, it is defined as:

\begin{equation}
\text{FDR} = \frac{|A \cap \overline{B}|}{|A|}
\end{equation}
where \(A\) is the set of predicted positive pixels, \(\overline{B}\) is the complement of the ground truth positive pixels, and \(|\cdot|\) denotes the cardinality (area) of a set.

\paragraph{False Omission Rate (FOR)}

The False Omission Rate (FOR) \citep{6b87510ce7324df69116f8395644ed77} measures the proportion of false negatives among the predicted negatives. In the context of segmentation, it quantifies the extent of undersegmentation, i.e., the degree to which the predicted segmentation misses areas within the ground truth segmentation. Mathematically, it is defined as:
\begin{equation}
\text{FOR} = \frac{|\overline{A} \cap B|}{|\overline{A}|}
\end{equation}
where \(\overline{A}\) is the complement of the set of predicted positive pixels, \(B\) is the set of ground truth positive pixels, and \(|\cdot|\) denotes the cardinality (area) of a set.

These metrics provide a detailed understanding of the segmentation performance by separately evaluating the extent of oversegmentation and undersegmentation, thereby offering more granular insights compared to traditional metrics like Intersection over Union (IoU).

\subsubsection{Boundary Evaluation Metrics}
\label{Main-boundary_metrics_section}

For the evaluation of polygon agreement, we use the Mean Surface Distance,  the Hausdorff Distance, and, for the case of the AI4SmallFarms dataset, the PoLiS distance.

\paragraph{Mean Surface Distance (MSD)}

The Mean Surface Distance (MSD) is a metric used to evaluate the average distance between the surfaces of two geometries. In the context of polygon boundaries, MSD provides a measure of how closely two boundaries align with each other by averaging the minimum distances between the vertices of one polygon to the vertices of the other polygon. This metric is particularly useful for comparing the overall proximity of two shapes without being overly sensitive to outlier distances.

Let \( X = \{x_1, x_2, \ldots, x_{N_x}\} \) and \( Y = \{y_1, y_2, \ldots, y_{N_y}\} \) be the sets of vertices of two polygons. The pairwise distance matrix \( D \) between these vertices is computed using the Euclidean distance:
\[
D_{ij} = \|x_i - y_j\|_2 \quad \text{for} \quad i = 1, 2, \ldots, N_x \quad \text{and} \quad j = 1, 2, \ldots, N_y
\]
The MSD is then calculated as the average of the mean of the minimum distances from each vertex in \( X \) to the vertices in \( Y \), and the mean of the minimum distances from each vertex in \( Y \) to the vertices in \( X \):
\[
\text{MSD}(X, Y) = \frac{1}{2} \left( \frac{1}{N_x} \sum_{i=1}^{N_x} \min_{j} D_{ij} + \frac{1}{N_y} \sum_{j=1}^{N_y} \min_{i} D_{ij} \right)
\]

\paragraph{Hausdorff Distance}

The Hausdorff Distance is a metric that measures the greatest distance from a point in one set to the closest point in another set. It provides a measure of the discrepancy between two sets of points and is particularly sensitive to outliers. For two polygon boundaries \( X \) and \( Y \), the Hausdorff Distance \( d_H \) is defined as:
\[
d_H(X, Y) = \max \left\{ \sup_{x \in X} \inf_{y \in Y} \|x - y\|_2, \sup_{y \in Y} \inf_{x \in X} \|y - x\|_2 \right\}
\]
where \( \sup \) and \( \inf \) denote the supremum and infimum, respectively. In practical terms, the Hausdorff Distance is computed as the maximum of the directed Hausdorff distances from \( X \) to \( Y \) and from \( Y \) to \( X \). The directed Hausdorff distance \( d_H^d \) from \( X \) to \( Y \) is given by:
\[
d_H^d(X, Y) = \max_{x \in X} \min_{y \in Y} \|x - y\|_2
\]
Thus, the Hausdorff Distance is calculated as:
\[
d_H(X, Y) = \max \left\{ d_H^d(X, Y), d_H^d(Y, X) \right\}
\]

Both the MSD and Hausdorff Distance provide valuable insights into the similarity and alignment of polygon boundaries, with the MSD offering an average-based measure and the Hausdorff Distance highlighting the most significant deviations.

\paragraph{Polygon Line--to--Segment distance (PoLiS)}
The \emph{PoLiS} metric \citep{6849454} evaluates the similarity of two closed polygon boundaries by averaging \emph{vertex--to--edge} distances, instead of the usual vertex--to--vertex formulation.  
For each vertex of one polygon the shortest orthogonal distance to the \emph{edges} (line segments) of the opposite polygon is measured, making PoLiS robust to different vertex samplings of the same curve and resilient to small topological artefacts such as self–intersections or micro–slivers.

Let \(P=\{p_1,\dots,p_{N_P}\}\) and \(Q=\{q_1,\dots,q_{N_Q}\}\) denote the ordered vertex sets of two polygons, and let \(\partial Q\) be the piece-wise linear boundary of \(Q\) (i.\,e.\ its set of edges).  
The \emph{directed} PoLiS from \(P\) to \(Q\) is the mean vertex–to–edge distance
\begin{align*}
d_{\mathrm{PoLiS}}(P\!\rightarrow\!Q)&=
\frac{1}{N_P}\sum_{i=1}^{N_P}\!
\mathrm{dist}  \bigl(p_i,\partial Q\bigr),
\\
\mathrm{dist} \bigl(p_i,\partial Q\bigr) &=
\min_{e\in\partial Q}\mathrm{dist}(p_i,e),
\end{align*}
where \(\operatorname{dist}(p_i,e)\) is the Euclidean distance from vertex \(p_i\) to edge \(e\).
The \emph{symmetric} PoLiS used in this work is the average of the two directed values,
\[
\mathrm{PoLiS}(P,Q)=
\frac{1}{2}\Bigl(
d_{\mathrm{PoLiS}}(P\!\rightarrow\!Q)
+
d_{\mathrm{PoLiS}}(Q\!\rightarrow\!P)
\Bigr).
\]

Because every vertex contributes equally and deviations are measured relative to the exact opposite boundary, PoLiS strikes a useful balance: it is less sensitive to single outliers than the Hausdorff distance, yet more discerning than purely vertex-based averages such as the MSD when one polygon is over- or under-sampled. In this work we use PoLiS in the object level evaluation of the AI4SmallFarms benchmark, to compare with provided baselines by \citet{10278130}.

\paragraph{Polygon Matching Algorithm} For the case of the AI4SmallFarms and ePaddocks datasets, we adopted a unified polygon matching algorithm  to ensure consistency in boundary evaluation metrics. The predicted extent masks were initially vectorized and topologically repaired. Each valid geometry was then matched to a single ground-truth polygon (the reference polygon with the highest Intersection over Union (IoU)), provided the IoU exceeded a threshold of $10^{-3}$. This one-to-one assignment ensures every boundary metric evaluation refers to genuine true-positive polygon pairs, preventing double-counting.

For each matched polygon pair, relevant boundary evaluation metrics were computed. Specifically, we calculated the symmetric Polygon Line-to-Segment (PoLiS) distance by averaging the vertex-to-edge distances in both directions for the AI4SmallFarms dataset, precisely following the definition by \citep{6849454}, so we can compare our results with the \citep{10278130} baselines. For the ePaddocks dataset, we employed the Hausdorff Distance and Mean Surface Distance (MSD), which measure the greatest boundary discrepancies and average boundary deviations, respectively. Polygons unmatched in this process (false positives or omissions) were accounted for within the pixel-level confusion matrix but excluded from boundary-specific metrics. This procedure ensures comparability across datasets irrespective of raster boundary thickness or spatial resolution.

\begin{table*}[!tp]
\fontsize{9}{11}\selectfont
  \centering
  \caption{\textcolor{red}{Experiment map linking the core research questions with the controlled companion experiments reported in the Supplementary Material (\cmark\;= dataset used,\; \xmark\;= not used). The horizontal rule separates the primary operational and public-benchmark evidence from supplementary ablations that isolate specific methodological factors.}}
  \label{Main-tab:exp-matrix}
  \begingroup
  \setlength{\tabcolsep}{3pt}
  \begin{tabularx}{\textwidth}{p{0.19\textwidth}ccccX}
    \toprule
     \textcolor{red}{Experiment / control question} & \textcolor{red}{ePaddocks} & \textcolor{red}{FoTW} & \textcolor{red}{PASTIS} & \textcolor{red}{AI4SF} & \textcolor{red}{Role in the study} \\
    \midrule
    \multicolumn{6}{l}{\textcolor{red}{Core study: operational and public-benchmark evidence}} \tabularnewline
    \textcolor{red}{Sparse-cloud S2 sourcing} & \textcolor{red}{\cmark} & \textcolor{red}{\xmark} & \textcolor{red}{\xmark} & \textcolor{red}{\xmark} & \textcolor{red}{Operational ePaddocks cloud-tolerance experiment and visual examples.} \\
    \textcolor{red}{S1 dense-cloud alternative} & \textcolor{red}{\cmark} & \textcolor{red}{\xmark} & \textcolor{red}{\xmark} & \textcolor{red}{\xmark} & \textcolor{red}{S1-only PTAViT3D evaluated where optical coverage is unreliable.} \\
    \textcolor{red}{S2/S1 modality and fusion} & \textcolor{red}{\cmark} & \textcolor{red}{\xmark} & \textcolor{red}{\xmark} & \textcolor{red}{\cmark} & \textcolor{red}{Sensor-modality evidence, including S2-only, S1-only, and S2--S1 fusion settings.} \\
    \textcolor{red}{3D vs. 2D temporal modelling} & \textcolor{red}{\cmark} & \textcolor{red}{\xmark} & \textcolor{red}{\xmark} & \textcolor{red}{\xmark} & \textcolor{red}{Comparison between time-series models and 2D alternatives under ePaddocks validation.} \\
    \textcolor{red}{Transferability} & \textcolor{red}{\xmark} & \textcolor{red}{\cmark} & \textcolor{red}{\cmark} & \textcolor{red}{\cmark} & \textcolor{red}{Public benchmark evidence on FoTW, PASTIS, and AI4SmallFarms.} \\
    \textcolor{red}{Continental product} & \textcolor{red}{\cmark} & \textcolor{red}{\xmark} & \textcolor{red}{\xmark} & \textcolor{red}{\xmark} & \textcolor{red}{Production workflow and validation for the 2023 national ePaddocks product.} \\
    \midrule
    \multicolumn{6}{l}{\textcolor{red}{Supplementary Material: controlled companion ablations}} \\
    \textcolor{red}{Controlled cloud and shadow stress} & \textcolor{red}{\xmark} & \textcolor{red}{\xmark} & \textcolor{red}{\cmark} & \textcolor{red}{\xmark} & \textcolor{red}{Fixed 16-date PASTIS S2 registry; 28 registered conditions comprising a clean reference, a mixed-family no-op, and 26 non-zero cloud/shadow perturbations.} \\
    \textcolor{red}{Regional sensitivity to simulated contamination frequency} & \textcolor{red}{\xmark} & \textcolor{red}{\xmark} & \textcolor{red}{\cmark} & \textcolor{red}{\xmark} & \textcolor{red}{The same temporal--spatial cloud and shadow schedule is grouped by native PASTIS source tile to quantify region-dependent accuracy loss.} \\
    \textcolor{red}{2D temporal alternatives} & \textcolor{red}{\xmark} & \textcolor{red}{\xmark} & \textcolor{red}{\cmark} & \textcolor{red}{\xmark} & \textcolor{red}{PTAViT2D stacked, four-date, and per-date mean controls against PTAViT3D.} \\
    \textcolor{red}{Per-date DA aggregation} & \textcolor{red}{\xmark} & \textcolor{red}{\xmark} & \textcolor{red}{\cmark} & \textcolor{red}{\xmark} & \textcolor{red}{Modern field-instance comparator applied per date and combined by mean occupancy, vote, union, and intersection.} \\
    \textcolor{red}{Attention mechanism} & \textcolor{red}{\xmark} & \textcolor{red}{\cmark} & \textcolor{red}{\xmark} & \textcolor{red}{\xmark} & \textcolor{red}{FoTW full-model replacement and matched-support resource benchmark, with a separately labelled auxiliary native-F256 training comparison.} \\
    \textcolor{red}{Loss controls} & \textcolor{red}{\xmark} & \textcolor{red}{\xmark} & \textcolor{red}{\cmark} & \textcolor{red}{\xmark} & \textcolor{red}{FTNMT, BCE, and BCE+Dice controls under clean and severe cloud-stress summaries.} \\
    \bottomrule
  \end{tabularx}
  \endgroup
\end{table*}

\subsection{Experimental design}

\textcolor{red}{Table~\ref{Main-tab:exp-matrix} links the four core experiments described below to the controlled companion ablations in Supplementary Table~\ref{SM-tab:app-ablation-design-matrix}.}

\paragraph{Cloud contamination effects} We first assess how cloud cover impacts the algorithmic performance, focusing on two distinct scenarios: (i) sparse cloud cover conditions with Sentinel-2 (S2) Satellite Image Time Series (SITS), detailed in Section \ref{Main-tfcl_s2_solution}, and (ii) dense cloud coverage scenarios where spatial resolution from Sentinel-1 (S1) time series predictions is critical, as discussed in Section \ref{Main-tfcl_s1_solution}. Both analyses leverage the ePaddocks dataset.

\paragraph{Sensor modality comparison} We then compare different sensor modalities and modeling approaches. Specifically, we evaluate:
\begin{itemize}
    \item Sentinel-2 SITS against Sentinel-1 SITS, using the ePaddocks dataset (Section \ref{Main-tfcl_sensor_mod_s2vss1}), and
    \item 3D versus 2D versions of our model, along with the joint modeling of Sentinel-2 and Sentinel-1 SITS. These evaluations are conducted using both the ePaddocks and AI4SmallFarms datasets (Section \ref{Main-tfcl_sensor_modality_s2s1joint}).
\end{itemize}

\paragraph{Generalization and benchmarking} To assess the generalizability of our modeling approach, we conduct experiments across multiple geographic regions and benchmark against established baseline methods. Our evaluations use three publicly available benchmark datasets: (i) Fields of the World (FoTW) \citep{kerner2024fields}, (ii) PASTIS \citep{garnot2021panoptic}, and (iii) AI4SmallFarms \citep{10278130}. This comparative analysis, detailed in Sections \ref{Main-FoTWAppendix}, \ref{Main-PastisEval}, and \ref{Main-AI4SmallFarms}, allows direct assessment of our PTAViT3D model against existing methodologies under matched conditions.

\paragraph{Continental-scale application} Finally, we showcase the practical application and scalability of the PTAViT3D model at a continental scale. Specifically, we apply our model across Australia for the creation of the ePaddocks 2023 national product (Section \ref{Main-ptavit3d_epaddocks}). This section highlights detailed product validation and important algorithmic considerations necessary for operational use.

Table \ref{Main-tab:exp-matrix} summarizes the datasets employed in each experimental scenario. Due to computational constraints, experiments with the ePaddocks dataset for validation purposes utilized only two Sentinel scenes each for training and validation. The final test results and visualizations are derived from independent scenes outside the training and validation sets. For comprehensive national-scale results, the full ePaddocks product employed training on six Sentinel-2 tiles.

Given that ePaddocks is not publicly available, quantitative comparison on this dataset is restricted to the available validation scenes and to baselines that can be retrained under the same data-access conditions. We therefore validated our methodology against an established 2D baseline by retraining the FracTAL ResNet proposed by \cite{rs13183707}, which has previously been applied successfully to field-boundary extraction \citep{rs13112197} and validated independently across multiple datasets \citep{rs14225738,Tetteh2023}. We note, however, that FracTAL is a 2D method and does not perform end-to-end temporal integration over the full Sentinel time series. A fully fair comparison with single-date or 2D methods would require an additional temporal consolidation strategy, such as date selection, averaging, or consensus over per-date predictions. PTAViT3D avoids this extra step by consuming the time series directly and learning the temporal weighting internally.

\paragraph{\textcolor{red}{Modern 2D temporal-aggregation baseline}} \textcolor{red}{We additionally use Delineate Anything (DA), a recent state-of-the-art field-instance delineation model developed on the large FBIS-22M dataset \citep{Lavreniuk2025DelineateAnything}, as a deliberately strong modern 2D comparator. DA is applied independently to each available RGB date and its predictions are combined by mean occupancy, majority vote, union, and intersection. This tests how a specialised single-date 2D model behaves under several post-hoc temporal aggregation rules, whereas PTAViT3D models the image time series jointly. The DA architecture, reported benchmark context, adaptation protocol, and aggregation rules are detailed in Supplementary Material, Section~\ref{SM-app:da-baselines}.}

\section{Results}

\begin{figure}[!htbp]
\begin{center}
\includegraphics[clip, trim=0.25cm 0.0cm 0.0cm 0.1cm,width=\columnwidth,height=0.72\textheight,keepaspectratio]{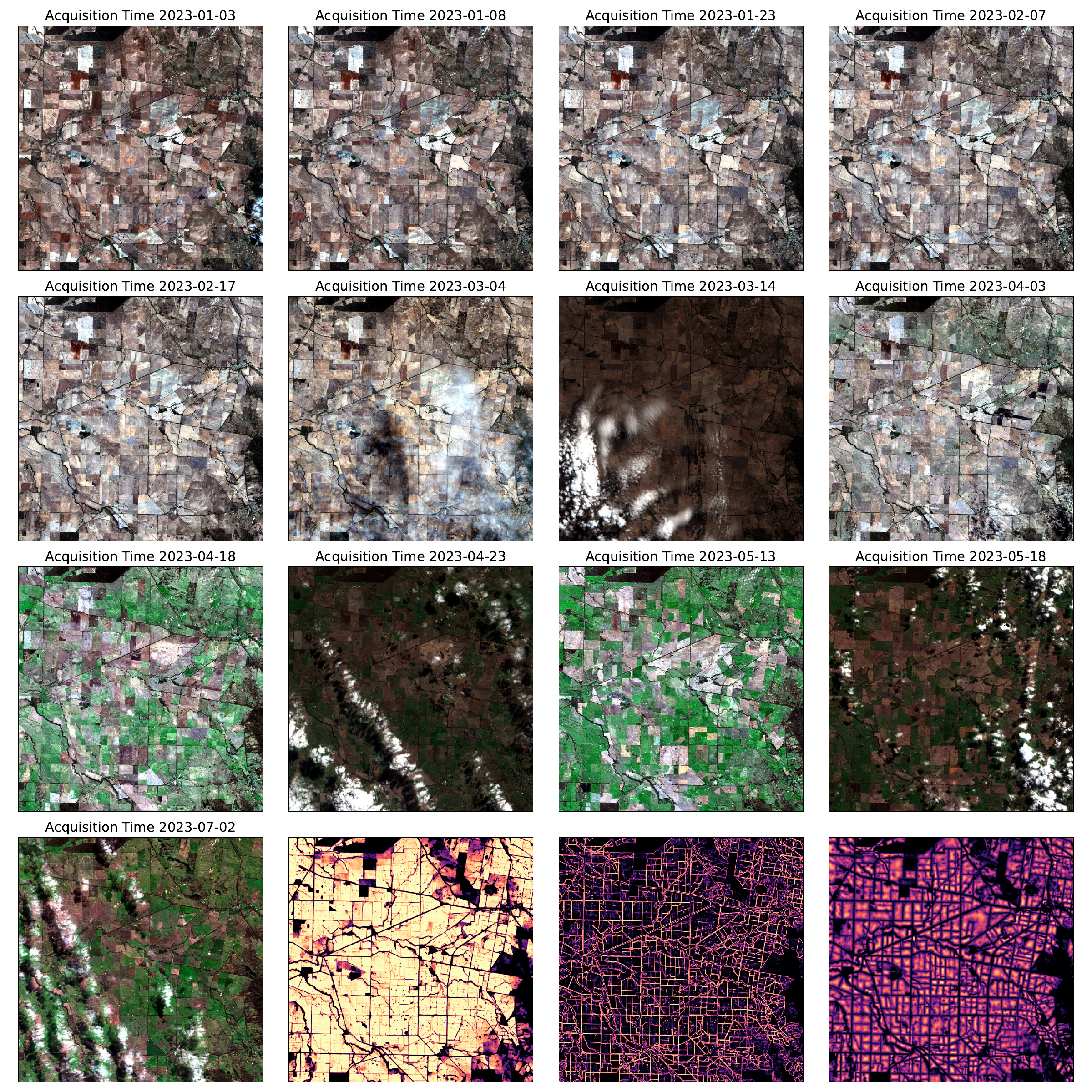}
\end{center}
\caption{Example of inference on partially clouded imagery (selection of S2A imagery with cloud tollerance $\leq 10\%$), natural color, ePaddocks dataset.  \textcolor{red}{The last three subplots correspond to the raster field extent, boundaries and distance transform that our algorithm predicts \citep{DIAKOGIANNIS202094,WALDNER2020111741,rs13112197}}.}
\label{Main-herecomescloud}
\end{figure}

\begin{figure*}[!tp]
\begin{center}
\includegraphics[clip, trim=0.025cm .5cm 0.05cm 0.01cm,width=\textwidth,height=0.72\textheight,keepaspectratio]{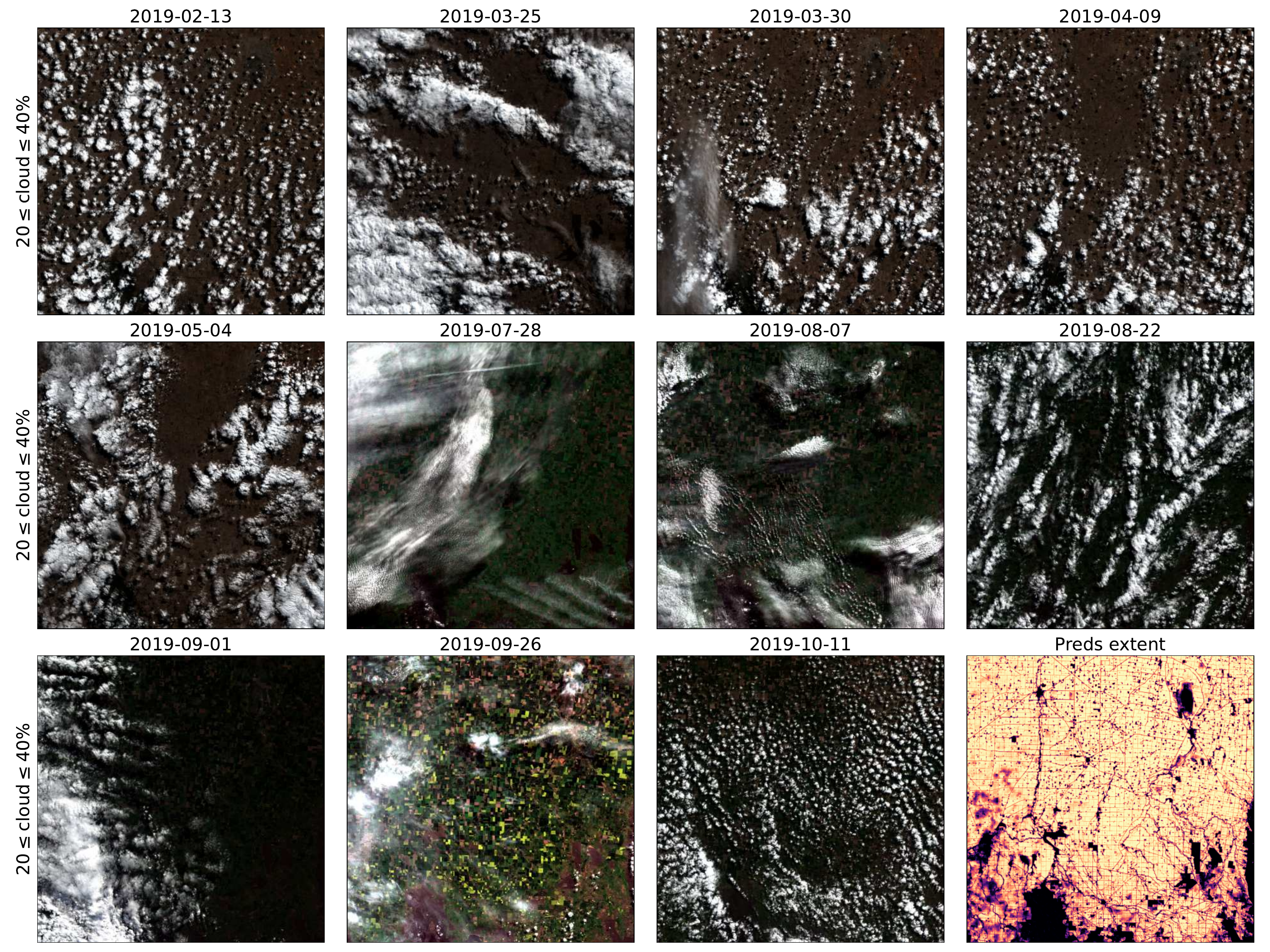}
\end{center}
\caption{Example of sourced imagery with up to 20\% to 40\% cloud tolerance. We also visualize the extent predictions, ePaddocks dataset.}
\label{Main-54hxe_cloud_40}
\end{figure*}

\subsection{Tackling sparse clouds}
\label{Main-tfcl_general}
In this section we are trying to address the question qualitatively using sourced imagery over Australia. Given that the way we source data through the Open Data Cube (ODC) API we can only select a cloud percentage, that relates to the whole scene coverage and not individual pixels, as well as the fact that deep learning models see a wide field of view (not just pixels) to make a decision, it is very hard to quantify precisely contaminated pixels in the field of view. Our best hope is to show examples where the requested input imagery (for the case of optical data) was selected with a cloud coverage tolerance. We note that different cloud cover tolerances give in general different number of time series of observations. This affects the overall performance, as the general rule is that the more time observations we use the better the performance \citep{DIAKOGIANNIS202444}.  In our approach, we request through the ODC API up to a maximum number of S2 tiles, within a given cloud tolerance, and we keep the ones that have the least cloud contamination and nodata ratio greater than 70\%. The key idea in going around the problem of sparse cloud coverage is that clouds move, and within an extended time series, all ground surface regions will be revealed irrespective of partial occlusion due to cloud contamination in individual time instances.

\subsubsection{S2 solution}
\label{Main-tfcl_s2_solution}

Here, we show two qualitative examples of inference, on a time series of 13 S2A tiles. These where sourced by requesting cloud coverage tolerance up to 10\% for the whole scene.  In figures  Fig.~\ref{Main-ptavit3d_cloud_demo} and Fig.~\ref{Main-herecomescloud} we show inference on a zoomed in area that has partially clouded images. 
By visual inspection of the predictions in these two cases, we see that there is no visible effect of the partially clouded areas on the predictions masks.  
We note that this is in stark contrast with the work of \citep{9150690} where the Authors there used also time series of input images into their spatio-temporal modeling: they report performance degradation in areas obscured by cloud. This can potentially be attributed in their treatment of time series of S2 images as well as the small number of time instances (they used three instances). In their work they consume time series via concatenation of the input images along the channel dimension, i.e. there is no explicit 3D structure into the deep learning modeling approach.

\emph{How much cloud is too much cloud?} To address this question, we sourced imagery for the test set of 67 tiles across the cropping area in Australia, using the Open Data Cube (ODC) API at varying levels of cloud coverage tolerance. The API allows specifying cloud coverage within intervals, so we selected $[0,20]\%$, $[0,50]\%$, and $[20,50]\%$. We also set upper bounds on the number of time instances used, at 16 and 32.

Table \ref{Main-how_much_cloud} shows the numerical performance in terms of IoU on the raster predictions. We observe a degradation in IoU performance between the $[0,20]\%$ and $[20,50]\%$ cloud coverage intervals when the maximum number of time instances is limited to 16, with IoU dropping from $\sim 0.8\pm 0.2$ to $\sim 0.6\pm 0.3$. This pattern is consistent whether the maximum number of time instances is 16 or 32. However, when the lower cloud coverage is fixed at 0, the performance between the $[0,20]\%$ and $[0,50]\%$ intervals remains nearly identical (within error bars).

The cloud-tolerance analysis in \textcolor{red}{Table~\ref{Main-how_much_cloud}} should be interpreted as an \emph{operational sourcing experiment} rather than a controlled per-pixel cloud ablation. In our pipeline, Sentinel-2 observations are requested through the Open Data Cube using a scene-level cloud-cover tolerance, after which observations with insufficient valid-data coverage are discarded and the remaining dates are reduced to at most 16 approximately evenly spaced time instances. Broader cloud-tolerance settings therefore define nested candidate pools of observations, but they do not correspond to independent groups with a fixed pixel-level cloud percentage over the model field of view. As a result, the observed performance reflects a trade-off between admitting cloudier observations and preserving a more complete temporal trajectory.

\textcolor{red}{The complementary controlled experiment fixes a 16-date PASTIS registry and applies synthetic thick cloud, thin cloud, cloud shadow, and mixed cloud--shadow at known temporal and spatial support fractions (Supplementary Sections~\ref{SM-app:controlled-cloud-ablation}--\ref{SM-app:regional-cloud-stress}). Cloud-augmented PTAViT3D reduces severe thin-cloud IoU loss from 0.074 to 0.031 and severe mixed cloud--shadow IoU loss from 0.032 to 0.015 while preserving clean-input IoU.}

\begin{figure*}[!tp]
\begin{center}
\includegraphics[clip, trim=0.025cm 3.0cm 0.05cm 0.01cm,width=\textwidth,height=0.72\textheight,keepaspectratio]{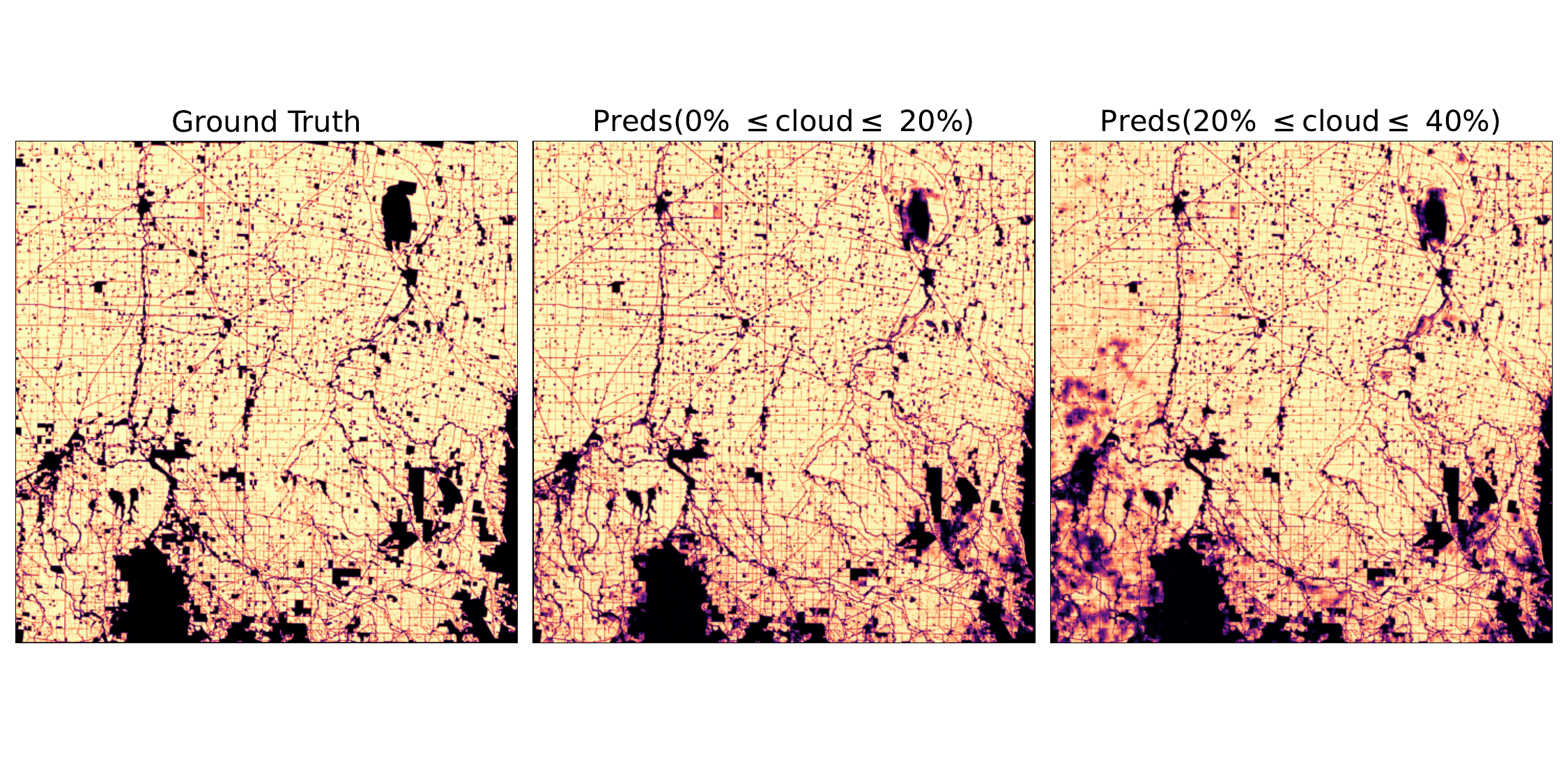}
\end{center}
\caption{Comparison of inference for sourced time series of S2 at varying levels of cloud tolerance.}
\label{Main-54hxe_preds_wth_cloud}
\end{figure*}

To fully appreciate the effectiveness of our algorithm's cloud tolerance, we also visualize sourced imagery from S2 tile 54HXE within the
$[20,40]\%$ (Fig \ref{Main-54hxe_cloud_40}) interval, alongside the corresponding predictions (Fig \ref{Main-54hxe_preds_wth_cloud}). As seen, there is some degradation when input imagery is restricted to 20-40\% cloud coverage, but the results remain practically useful.

\begin{table*}[!tp]
\fontsize{8}{10}\selectfont
\centering
\begin{tabular}{ccc}
\toprule
\textbf{Cloud Coverage (\%)} & $\max$ \textbf{Time Instances} &$\langle \text{\bf IoU} \rangle$ \\
\hline
$[0, 20]$  & 16 & $0.79 \pm 0.21$ \\
$[0, 50]$  & 32 & $0.77 \pm 0.22$ \\
$[20, 50]$ & 16 & $0.5818 \pm 0.3005$ \\
$[20, 50]$ & 32 & $0.5824 \pm 0.2999$ \\
\bottomrule
\end{tabular}
\caption{Segmentation IoU values for different cloud coverage tolerance intervals.}
\label{Main-how_much_cloud}
\end{table*}

\subsubsection{S1 solution}
\label{Main-tfcl_s1_solution}

\begin{figure*}[!tp]
\begin{center}
\includegraphics[clip, trim=0.25cm .25cm .028cm 0.25cm,width=\textwidth,height=0.72\textheight,keepaspectratio]{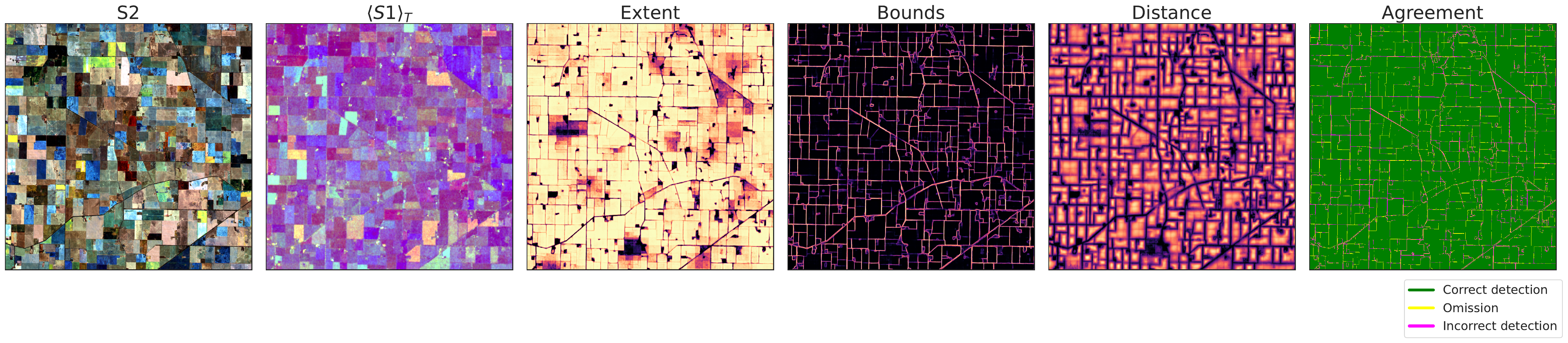}
\end{center}
\caption{ Example inference results from the PTAViT3D model using Sentinel-1 (S1) time series as input (16 temporal instances). The figure illustrates model outputs for extent, bounds, and distance layers. The final panel (``Agreement'') explicitly compares the inferred field boundaries against the ground truth, highlighting areas of correct detection, omissions, and incorrect detections.}
\label{Main-example_s1_agreement}
\end{figure*}

Here we again utilize the ePaddocks dataset. In areas that are mostly covered with clouds and it is not realistic to anticipate partial cloud cover, our S1 proposed algorithmic pipeline offers competitive results and a viable solution. Indeed our new model, PTAViT3D that consumes S1 time series performs on par with the baseline 2D FracTAL algorithm (Table \ref{Main-tab:evaluation_metrics}) that was trained on higher spatial resolution of optical S2 imagery. In Figure \ref{Main-example_s1_agreement} we visualize inference examples for the Algorithm PTAViT3D trained on time series of S1 images (16 time instances). From left to right we visualize the corresponding S2 scene, the S1 scene (bands $VH$, $VV$, anisotropy $A$, average across all times), the extent density prediction, $e$, the bounds density prediction, $b$, the Distance transform, $d$, as well as the agreement with the ground truth fields.

\begin{figure*}[!tp]
\begin{center}
    \begin{minipage}{0.45\textwidth}
        \includegraphics[clip, trim=0.25cm 0.0cm 0.05cm 0.1cm, width=\textwidth,height=0.72\textheight,keepaspectratio]{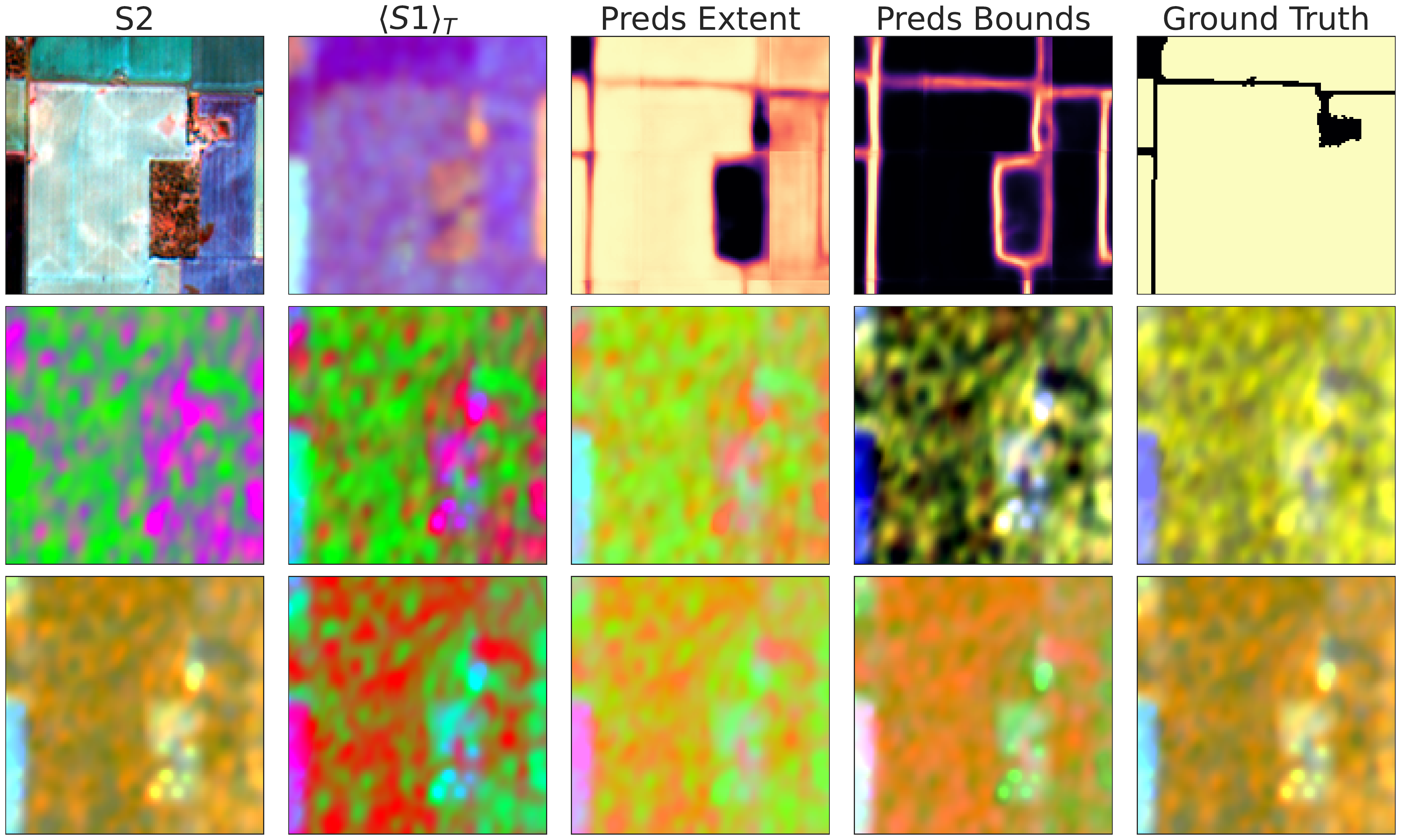}
        \label{Main-s1_upscale_example3}
    \end{minipage}
    \hfill
    \begin{minipage}{0.45\textwidth}
        \includegraphics[clip, trim=0.25cm 0.0cm 0.05cm 0.1cm, width=\textwidth,height=0.72\textheight,keepaspectratio]{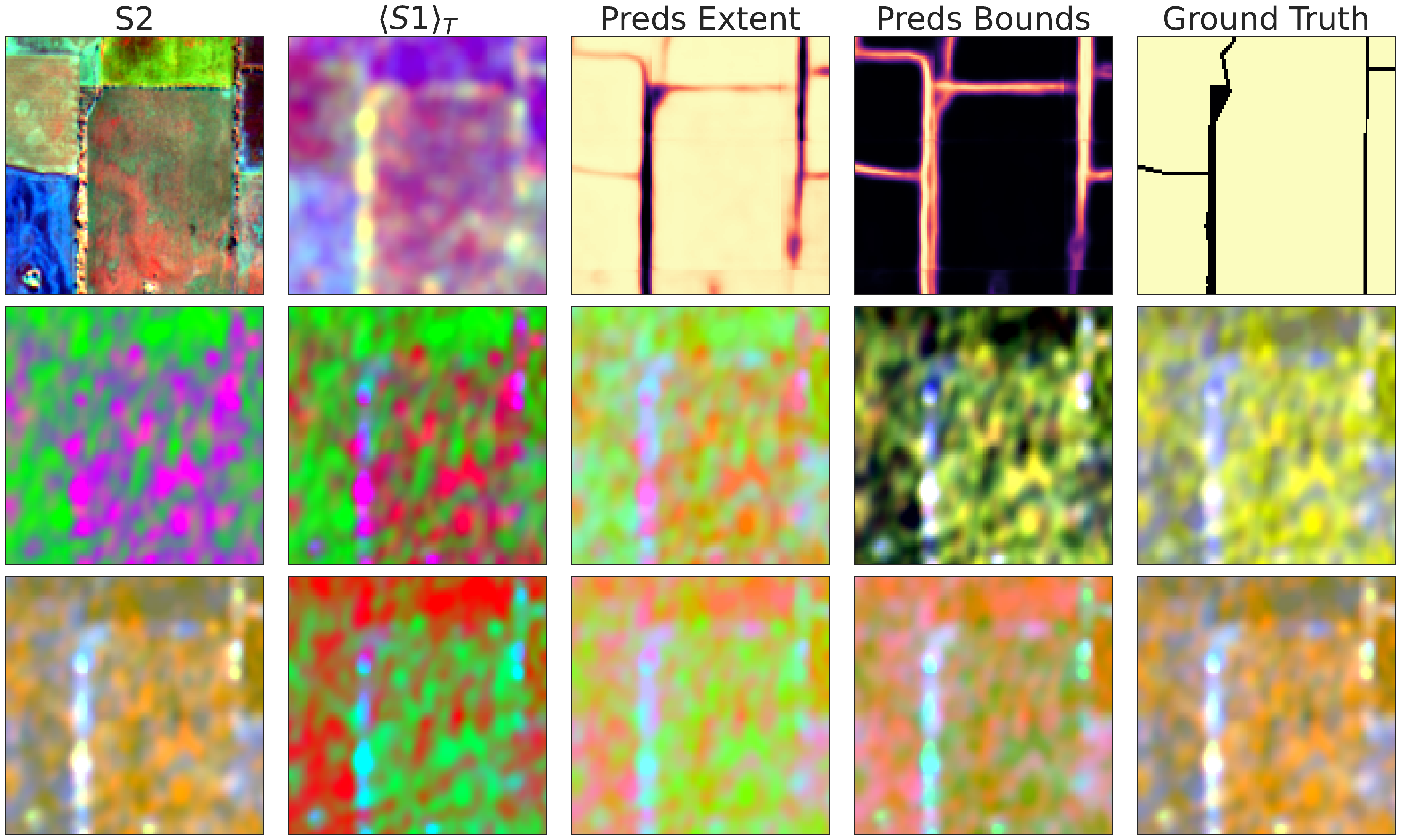}
        \label{Main-s1_upscale_example4}
    \end{minipage}
\end{center}
\caption{
Illustration of the spatial-resolution enhancement achieved by PTAViT3D predictions from Sentinel-1 (S1) time series imagery. The top row shows, from left to right, the reference Sentinel-2 (S2) imagery (not used by the model, provided here solely for visual comparison), the temporal average of S1 bands (VH,VV,A), the predicted extent and boundary layers, and the ground truth field boundaries at S2 resolution. The next two rows visualize all possible combinations of triplets from the five available S1 bands, specifically for the last date in the time series, highlighting that no single-date or averaged snapshot matches the detailed spatial information of the temporally integrated predictions.
}
\label{Main-s1_upscale_example1}
\end{figure*}

It can be seen (Figures 
 \ref{Main-example_s1_agreement}, \ref{Main-s1_upscale_example1}) that the quality of the predictions is of high standard following closely the algorithmic results of the algorithm trained on the higher spatial resolution S2 data. In Figure \ref{Main-example_s1_agreement} we visualize the agreement between the predicted field boundaries and the ground truth.

One very interesting result is that the spatial resolution of the predicted labels, is not close to the static single date input imagery as can be seen in Figure \ref{Main-s1_upscale_example1}. That is, the algorithm, through the inclusion of additional time information, learns to effectively upscale the predicted labels to the level of the ground truth pixel resolution. 
In Figure \ref{Main-s1_upscale_example1} we show the performance on the scale of a single chip, of shape 128x128. In the top row, from left to right we provide the corresponding S2 imagery (that is not seen from the PTAViT3D S1 algorithm), the average across all times S1 (bands $VH$, $VV$ and $A$), the predicted extent and boundary layers as well as the ground truth (that corresponds to the S2 image resolution). In the next 2 rows we visualize all possible combinations of triplets for the 5 S1 bands, for the last date. It can be seen that the detail level of the ground truth masks cannot be deduced from a single image, nor the time averaged signal.

We also note that the visualization quality of the S1 predictions can be improved significantly by utilizing more data. The visualizations shown here where created by training on only two scenes, and testing on a separate test tile.

\begin{figure}[!htbp]
\begin{center}
\includegraphics[clip, trim=0.15cm 0.05cm 0.05cm 0.05cm,width=\columnwidth,height=0.72\textheight,keepaspectratio]{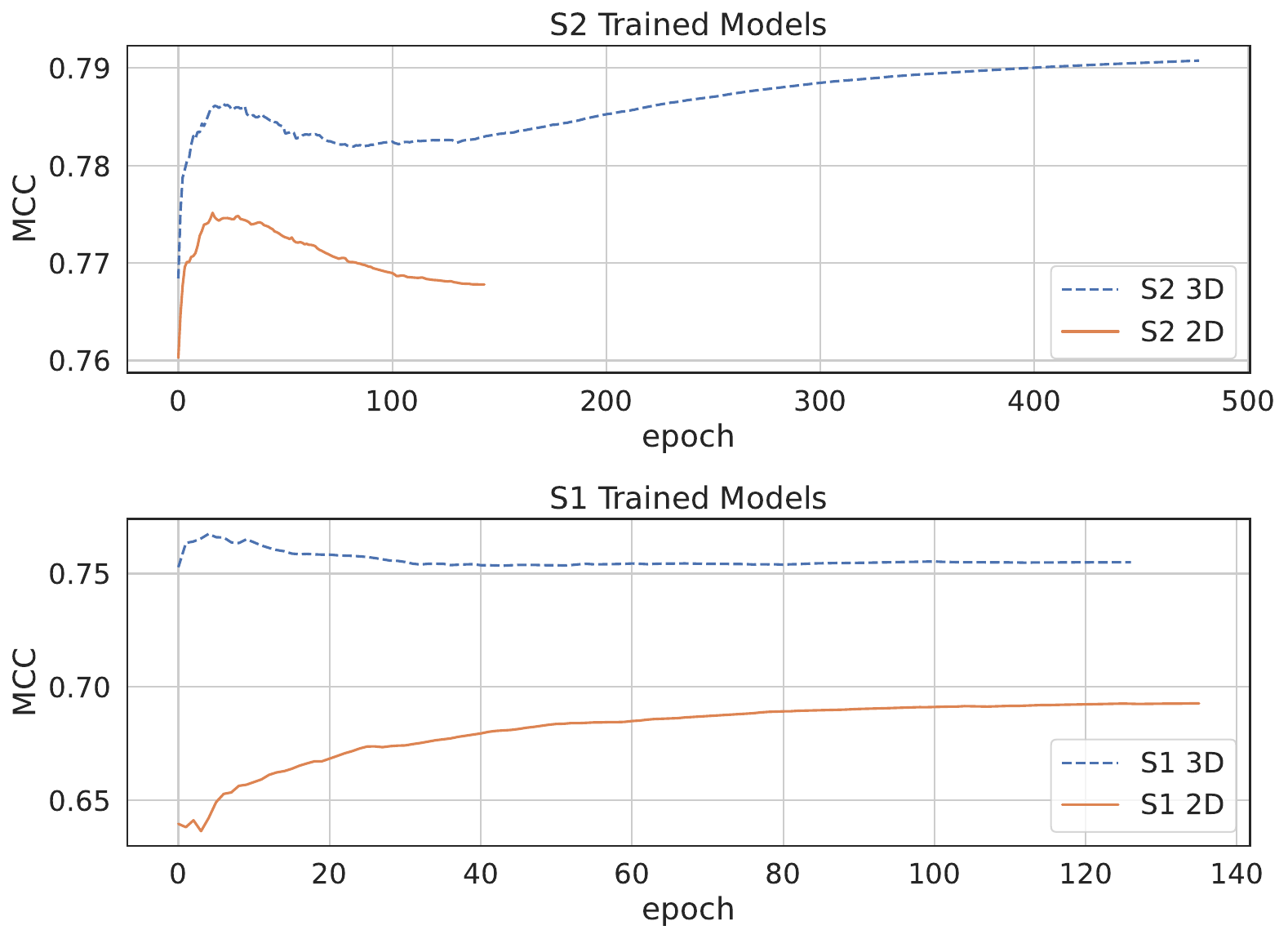}
\end{center}
\caption{Validation MCC evolution. The 3D models are evaluated with 4 time inputs. The focus is comparing S2 and S1 quality as discriminative features.}
\label{Main-compare_2D_vs_3D}
\end{figure}

\begin{figure}[!htbp]
\begin{center}
\includegraphics[clip, trim=0.05cm 0.05cm 0.05cm 0.05cm,width=\columnwidth,height=0.72\textheight,keepaspectratio]{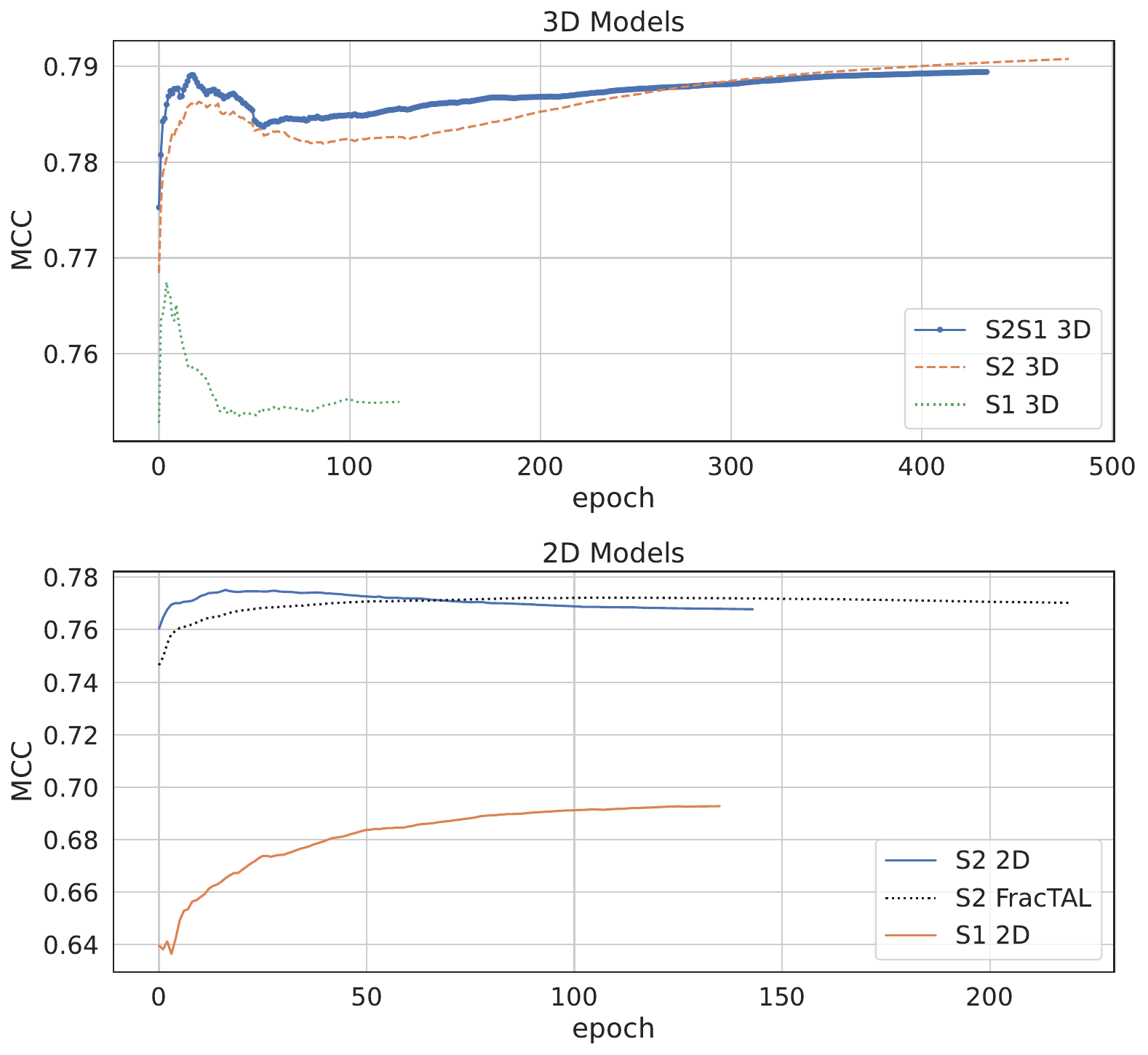}
\end{center}
\caption{As in Fig \ref{Main-compare_2D_vs_3D} with focus on comparing 3D architectures with 2D ones.} 
\label{Main-compare_all_models}
\end{figure}

\subsection{Sensor modality analysis}
How does model performance is affected with different kind of inputs. Here we analyse S2 vs S1 inputs performance, 3D vs 2D  vs joint S2\&S1 training.

\subsubsection{S2 vs S1 inputs for field delineation}
\label{Main-tfcl_sensor_mod_s2vss1}

We compare the performance between S2 and S1 models using the ePaddocks dataset. 
In Fig. \ref{Main-compare_2D_vs_3D} we plot validation metrics for both 3D and 2D models trained on only S2 imagery (top panel) as well as on only S1 imagery (bottom panel). For the S2 imagery the 3D model consumes 4 cropping season time observations, while the  S1 model consumes 16 time instances within the same time span. In both 2D and 3D cases the Sentinel 2 inputs provide a stronger signal for field boundary extraction. It is to be anticipated that due to the higher resolution S2 would perform better, however the increased time resolution of S1 could provide a counter balance in performance as we know that performance increases with the number of time steps \citep{DIAKOGIANNIS202444}. Despite the lower score, the 3D S1 model achieves excellent performance to be useful as a tool in areas where there is dense throughout the year cloud coverage, but it lags behind the S2 observations. We note that the performance of the PTAViT3D model with S1 inputs is on par with the FracTAL ResNet baseline (Table \ref{Main-tab:evaluation_metrics}) which is trained on 2D cloud free S2 images.

\subsubsection{3D vs 2D vs joint S1\&S2}
\label{Main-tfcl_sensor_modality_s2s1joint}

For the case of the ePaddocks dataset, in Fig. \ref{Main-compare_all_models} we provide evolution plots on the validation split of the dataset for 3D models (top panel) as well as 2D models (bottom panel). 
The S2 and S2\&S1 3D models are evaluated with 4 time inputs, while the S1 3D model with 16 time inputs. 
As expected, the time series (3D) models outperform their 2D counterparts in all cases. The evolution of the model S2S1 3D can be seen in the same Figure. 

We observe that the model that consumes both S2 and S1 features converges faster to optimality, however after approximately 200 epochs the two models become similar in performance, showing that the 3D model trained on S2 only is a practical cost effective alternative to the S2\& S1. The choice between the two boils down to the percentage of cloud coverage as well as computational resources available, given the S2\& S1 converges faster to higher performance.

Finally in Table \ref{Main-tab:evaluation_metrics} we present the numerical values from all the models, where it can be seen that the ranking is S2\&S1 3D $\geq$ S2 3D $\geq$ S2 2D $\geq$ S1 3D $\approx$ FracTAL S2  $\ge$ S1 2D.

\textcolor{red}{Supplementary Material, Section~\ref{SM-app:temporal-2d-baselines}, extends this comparison with channel-stacked, four-date, and per-date-averaged 2D controls on PASTIS. Under the severe cloud/shadow summary, PTAViT3D achieves object F1 of 0.698 and PoLiS of 43.93~m, compared with 0.693 and 46.34~m for the strongest stacked-2D control and 0.525 and 73.27~m for per-date averaging.}

\begin{table*}[!tp]
\fontsize{8}{10}\selectfont
\centering
\begin{tabular}{|l|c|c|c|c|c|c|c|c|}
\hline
\textbf{Model} & \textbf{NTimes} & \textbf{Epoch} & \textbf{MCC} &  \textbf{Precision} & \textbf{Recall} & \textbf{F1} & \textbf{IoU} \\
\hline
S2S1 3D & 4 & 434 & \textbf{0.79}  & \textbf{0.89} & 0.91 & \textbf{0.89} & \textbf{0.81} \\[5pt]
S2 3D & 4 & 477 & \textbf{0.79}  & 0.88 & 0.92 & \textbf{0.89} & \textbf{0.81} \\
S1 3D & 16 & 4 & 0.77  & 0.84 & \textbf{0.95} & 0.88 & 0.79 \\[5pt]
S2 2D & 1 & 16 & 0.78  & 0.84 & \textbf{0.95} & 0.88 & 0.79 \\
S1 2D & 1 & 104 & 0.69  & 0.82 & 0.88 & 0.84 & 0.73 \\[5pt]
FracTAL S2 & 1 & 108 & 0.77  & 0.84 & \textbf{0.95} & 0.88 & 0.79 \\
\hline
\end{tabular}
\caption{Evaluation Metrics for different models  (i.e. different sensor modalities) on the ePaddocks validation set (Australia). The best numerical values are designated with \textbf{bold} font.}
\label{Main-tab:evaluation_metrics}
\end{table*}

\begin{table*}[!tp]
\centering
\caption{Comparison of pixel-level mIoU scores between PTAViT3D and FoTW 2-class baseline model across 25 countries. \textcolor{red}{Published baseline values are available for 21 countries.} The best score per country is shown in \textbf{bold}.}
\begin{tabular}{lcc}
\toprule
\textbf{Test Country} & \textbf{FoTW} & \textbf{PTAViT3D} \\
\midrule
Austria       & 0.77 & \textbf{0.87} \\
Belgium       & 0.80 & \textbf{0.89} \\
Brazil        & --   & \textbf{0.60} \\
Cambodia      & \textbf{0.76} & 0.67 \\
Corsica       & 0.49 & \textbf{0.74} \\
Croatia       & 0.76 & \textbf{0.85} \\
Denmark       & 0.84 & \textbf{0.85} \\
Estonia       & 0.81 & \textbf{0.90} \\
Finland       & 0.87 & \textbf{0.93} \\
France        & 0.82 & \textbf{0.84} \\
Germany       & 0.80 & \textbf{0.86} \\
India         & --   & \textbf{0.59} \\
Kenya         & --   & \textbf{0.50} \\
Latvia        & 0.84 & \textbf{0.91} \\
Lithuania     & 0.78 & \textbf{0.87} \\
Luxembourg    & 0.85 & \textbf{0.89} \\
Netherlands   & 0.79 & \textbf{0.91} \\
Portugal      & 0.29 & \textbf{0.56} \\
Rwanda        & --   & \textbf{0.57} \\
Slovakia      & 0.93 & \textbf{0.94} \\
Slovenia      & 0.69 & \textbf{0.85} \\
South Africa  & \textbf{0.82} & 0.81 \\
Spain         & 0.84 & \textbf{0.88} \\
Sweden        & 0.83 & \textbf{0.89} \\
Vietnam       & 0.67 & \textbf{0.73} \\
\midrule
\textcolor{red}{\textbf{Matched-country mean ($n=21$)}} & \textcolor{red}{0.764} & \textcolor{red}{\textbf{0.840}} \\
\textcolor{red}{\textbf{Matched-country std}} & \textcolor{red}{0.137} & \textcolor{red}{\textbf{0.091}} \\
\textcolor{red}{\textbf{Matched-country min}} & \textcolor{red}{0.29} & \textcolor{red}{\textbf{0.56}} \\
\bottomrule
\end{tabular}
\label{Main-FoTW:miou_comparison}
\end{table*}

\begin{figure}[!htbp]
\begin{center}
\includegraphics[clip, trim=0.325cm 0.30cm 0.35cm 0.13cm,width=\columnwidth,height=0.72\textheight,keepaspectratio]{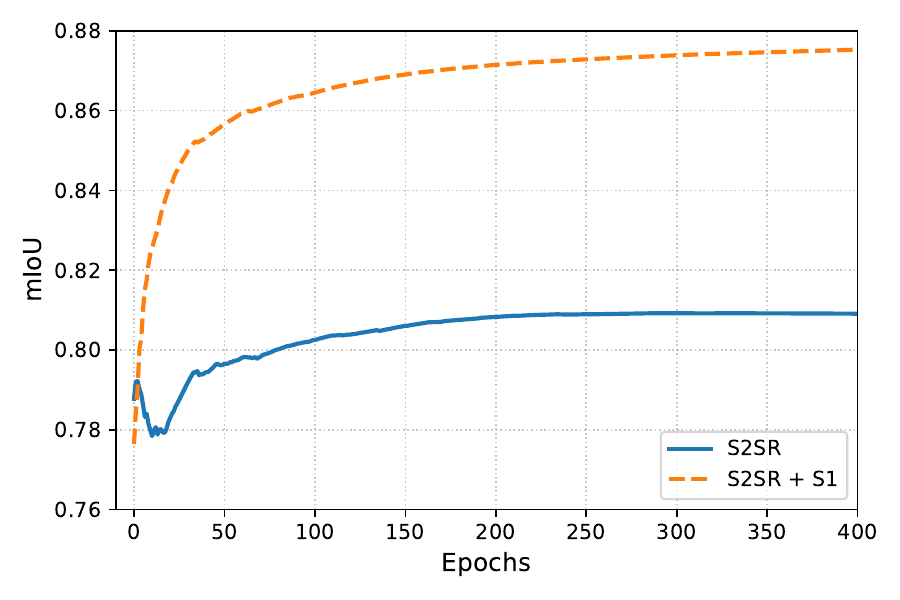}
\end{center}
\caption{Performance comparison for the \texttt{AI4SmallFarms} dataset of models trained with a single S2 super resolution image (S2SR) vs S2SR in combination with a time series of S1 images.}
\label{Main-AI4SmallFarmsS2S1Performance}
\end{figure}

For the case of the \texttt{AI4SmallFarms} dataset, that consumes a single S2-SR input image and a time series of S1 models, the situation of performance is different, indicating that the S1 time series can add significant value to the modeling. Fig.~\ref{Main-AI4SmallFarmsS2S1Performance} contrasts the validation set performance during  training: adding S1 (dashed/orange) sharpens convergence and lifts mIoU by $+6$ to $+7$pp over the S2-SR-only baseline. 

\subsection{Transferability across regions}
\label{Main-tfcl_across_world}

\subsubsection{Performance on Fields of the World (FoTW) Benchmark}
\label{Main-FoTWAppendix}

\begin{figure*}[!tp]
\begin{center}
\includegraphics[clip, trim=0.025cm 0.0cm 0.05cm 0.01cm,width=\textwidth,height=0.72\textheight,keepaspectratio]{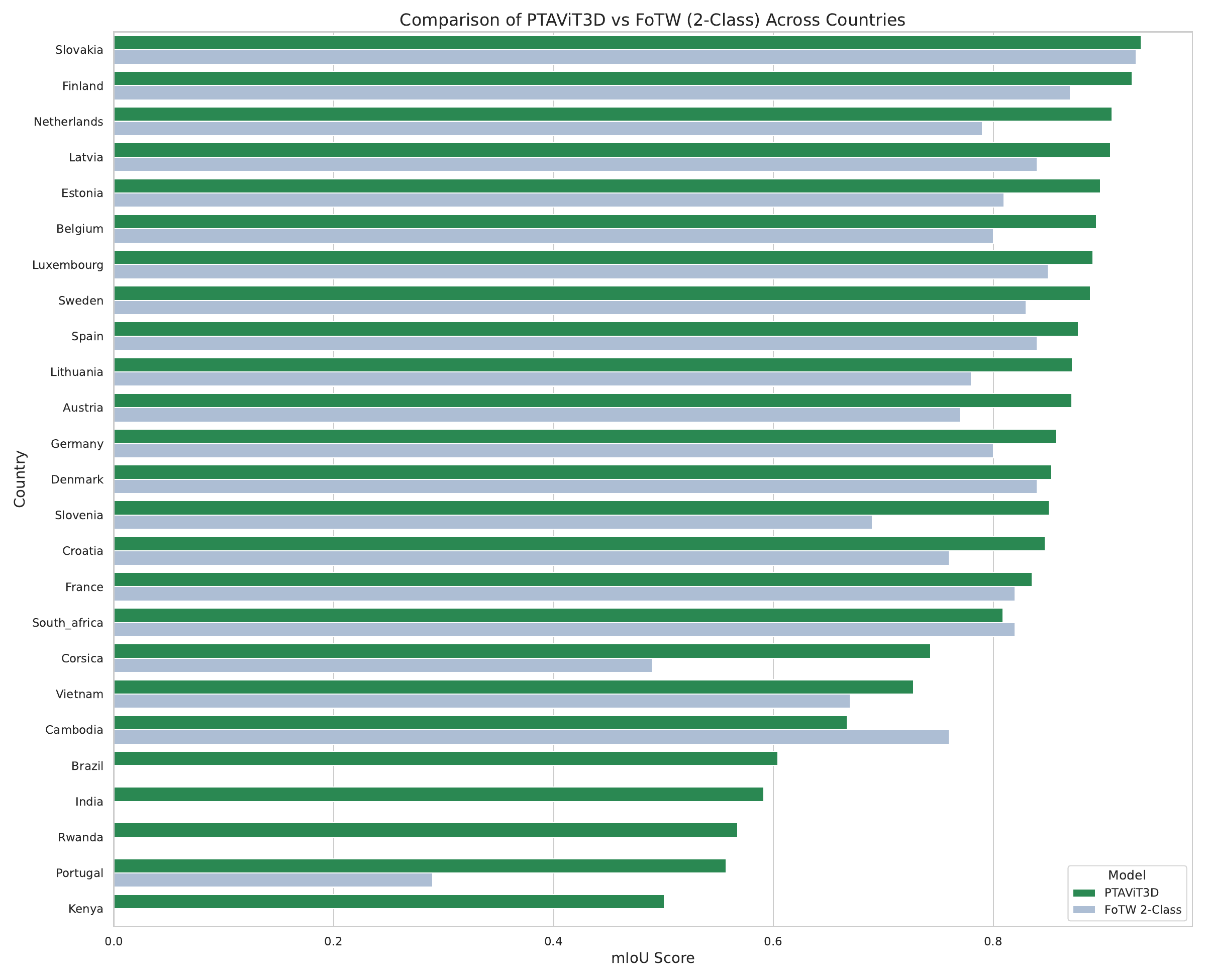}
\end{center}
\caption{Performance on FoTW in comparison with Baseline for each country.}
\label{Main-FoTWComparison}
\end{figure*}

We compared our model to the 2-class baseline published with the FoTW benchmark \citep{kerner2024fields}, which uses a U-Net with an EfficientNet-B3 backbone and was trained under identical input conditions (i.e., same number of bands and timestamps, identical splits on training/validation test folds of the FoTW dataset). Table~\ref{Main-FoTW:miou_comparison} and Fig.~\ref{Main-FoTWComparison} summarize the per-country results in terms of pixel-level mIoU.

Despite the fact that PTAViT3D was developed and finetuned within the context of the Australian cropping system, it performs competitively or better than the FoTW baseline in the majority of evaluated countries. \textcolor{red}{Among the 21 countries for which both scores are available, PTAViT3D achieves higher mIoU in 19.} Notably, it consistently outperforms the baseline in diverse settings such as Finland, Slovakia, Germany, Estonia, and Sweden—regions with varied field morphologies and parcel complexities.

\textcolor{red}{Across the 21 matched countries, PTAViT3D increases mean mIoU from 0.764 to 0.840, reduces the standard deviation from 0.137 to 0.091, and raises the minimum from 0.29 to 0.56. These matched-country statistics demonstrate both higher accuracy and more consistent performance than the published FoTW baseline.}

While the architectural design of PTAViT3D was not specifically fine-tuned on FoTW data, its competitive performance underlines the effectiveness of its spatiotemporal encoding and attention mechanisms. These results collectively suggest that PTAViT3D generalizes well across regions and can serve as a strong baseline for future research in global field boundary segmentation.

\subsubsection{Performance on the PASTIS benchmark}
\label{Main-PastisEval}

\begin{table*}[!tp]
    \centering
    \caption{Semantic–segmentation results on the PASTIS benchmark
             (all 19 classes, background included). We report also the results for U-TAE replicated from \cite{garnot2021panoptic}. We used 24 random samples in Time over all time series elements. The best score is shown in \textbf{bold}.}
    \vspace{2mm}
    \begin{tabular}{lccc}
        \toprule
        \textbf{Model} & \textbf{OA} & \textbf{mIoU} & \textbf{MCC} \\
        \midrule
        U-TAE & 0.832 & 0.631 & -- \\
        PTAViT3D (ours) & \textbf{0.847} & \textbf{0.649} & \textbf{0.796} \\
        \bottomrule
    \end{tabular}
    \label{Main-tab:pastis_results_bg}
\end{table*}

\begin{table*}[!tp]
    \centering
    \caption{Performance on the PASTIS dataset as a function of the number of Sentinel-2 acquisitions used
             (all 19 classes, background included). The best score  is shown in \textbf{bold}.}
    \vspace{2mm}
    \begin{tabular}{lcccc}
        \toprule
        \textbf{Metric} & \textbf{16$\times$} & \textbf{24$\times$} & \textbf{32$\times$} & \textbf{48$\times$} \\
        \midrule
        Overall accuracy & 0.842 & \textbf{0.847} & 0.846 & 0.844 \\
        mIoU            & 0.633 & \textbf{0.649} & 0.647 & 0.639 \\
        MCC             & 0.790 & \textbf{0.796} & 0.796 & 0.793 \\
        \bottomrule
    \end{tabular}
    \label{Main-tab:time_vs_scores}
\end{table*}

\begin{figure*}[!tp]
\begin{center}
\includegraphics[clip, trim=0.025cm 0.0cm 0.05cm 0.01cm,width=\textwidth,height=0.72\textheight,keepaspectratio]{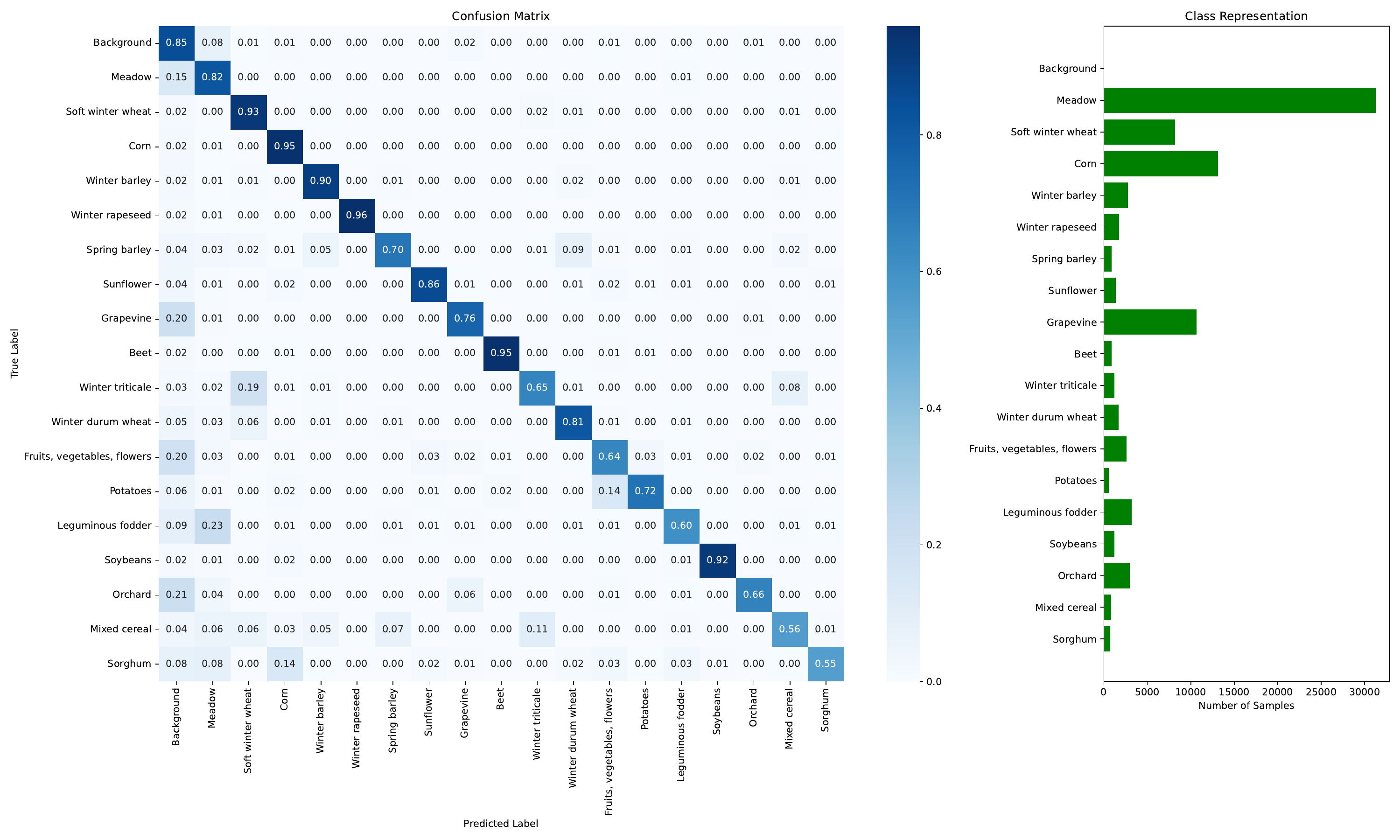}
\end{center}
\caption{Performance over 19 classes on PASTIS dataset using randomly sampled 24 instances of S2 time images.}
\label{Main-PastisPerformance}
\end{figure*}

To demonstrate that PTAViT3D is not limited to field-boundary tasks, we evaluated it on the full \textsc{PASTIS} land-cover benchmark \citep{garnot2021panoptic}.  This dataset comprises 38--61 Sentinel-2 (S2) acquisitions per pixel; even after discarding scenes with $>90\%$ cloud, roughly \textbf{28\,\%} of the remaining images still contain partial cloud cover \citep{garnot2021panoptic}.  We adhered to the official five-fold train/val/test split and trained with \emph{only eight} randomly sampled acquisitions per pixel.  

At inference, we swept the number of snapshots, \(N \in \{16,24,32,48\}\) (Table~\ref{Main-tab:time_vs_scores}). Accuracy peaks at \(N=24\), reaching \textbf{0.847} OA, \textbf{0.649} mIoU, and \textbf{0.796} MCC, while varying by at most \(0.01\) OA even when the time series is doubled.

Using 24 frames for PTAViT3D, Table~\ref{Main-tab:pastis_results_bg} shows that PTAViT3D improves on U-TAE by  
\(+1.5\) pp OA (0.847 vs 0.832) and \(+1.8\) pp mIoU (0.649 vs 0.631) while also providing a calibrated MCC of 0.796.

The confusion matrix in Fig.~\ref{Main-PastisPerformance} reveals that 11 of the 19 classes (including background) achieve \(\ge 0.75\) recall.  Notable strong classes include \textit{soft-winter wheat} (0.93 recall) and \textit{corn} (0.95 recall).  Most errors occur among botanically similar cereals with few training samples--e.g.\ \textit{mixed cereal} (0.56 recall) and \textit{winter triticale} (0.65 recall).  

The accompanying bar-plot (right panel, Fig.~\ref{Main-PastisPerformance}) highlights the extreme skew: \textit{meadow} contributes \(\approx30\,000\) instances whereas seven classes have fewer than 2 000.  Achieving high mIoU under this imbalance -- without focal loss or re-weighting -- shows that PTAViT3D’s patch-wise Tanimoto attention remains robust on dense, cloud-affected time series.

\subsubsection{Smallholder rice paddies: \textsc{AI4SmallFarms}}
\label{Main-AI4SmallFarms}

\begin{table*}[!tp]
\centering
\caption{Evaluation on the \textsc{AI4SmallFarms} \emph{test} split.  
Top block in each region: \textbf{pixel-level} precision (P), recall (R), $F_{1}$ and IoU for field extent and boundary.  
Bottom block: \textbf{polygon-level} mean PoLiS, (symmetric) Hausdorff distance and mean surface distance (MSD).  
Best score in each sub–category, where baselines exist for comparison, is in \textbf{bold}.}
\label{Main-tab:ai4sf_metrics}
\renewcommand{\arraystretch}{1.05}
\begin{tabular}{@{}llcccc@{}}
\toprule
\multicolumn{2}{c}{\textbf{Region / Mask}} & \textbf{P} & \textbf{R} & \textbf{$F_{1}$} & \textbf{IoU} \\ \midrule
\multirow{8}{*}{\textbf{Global}}
    & Extent (2.5m-ours)               & 0.933 & 0.874 & 0.902 & 0.822 \\[-3pt]
    & Extent (10m-ours)                & 0.935 & 0.933 & 0.934 & 0.876 \\ \cmidrule{2-6}
    & Boundary (2.5m-ours)             & 0.193 & 0.135 & 0.159 & 0.086 \\[-3pt]
    & Boundary (10m-ours)              & \textbf{0.567} & 0.480 & \textbf{0.520} & \textbf{0.351} \\[-3pt]
    & Boundary (10m-\citet{10278130}) & 0.480 & 0.330 & 0.390 & -- \\ 
    & Boundary (0.5m-\citet{10278130}) & 0.47 & \textbf{0.52} & 0.49 & -- \\ \cmidrule{2-6}
    & PoLiS (2.5m-ours)                & \multicolumn{4}{c}{\bf 14.35m} \\[-3pt]
    & PoLiS (10m-ours)                 & \multicolumn{4}{c}{15.06m} \\[-3pt]
    & PoLiS (10m-\citet{10278130})     & \multicolumn{4}{c}{26.7m} \\ 
    & PoLiS (0.5m-\citet{10278130})     & \multicolumn{4}{c}{20.3m} \\[-3pt] \midrule
\multirow{8}{*}{\textbf{Cambodia}}
    & Extent (2.5m-ours)               & 0.962 & 0.894 & 0.926 & 0.863 \\[-3pt]
    & Extent (10m-ours)                & 0.962 & 0.959 & 0.961 & \textbf{0.925} \\[-3pt]
    & Extent (10m-\citet{kerner2024fields}) & -- & -- & -- & 0.760 \\ \cmidrule{2-6}
    & Boundary (2.5m-ours)             & 0.187 & 0.143 & 0.162 & 0.088 \\[-3pt]
    & Boundary (10m-ours)              & 0.566 & 0.502 & 0.532 & 0.362 \\ \cmidrule{2-6}
    & PoLiS (2.5m-ours)                & \multicolumn{4}{c}{12.01m} \\
    & PoLiS (10m-ours)                 & \multicolumn{4}{c}{12.73m} \\[-3pt]\midrule
\multirow{8}{*}{\textbf{Vietnam}}
    & Extent (2.5m-ours)               & 0.884 & 0.839 & 0.861 & 0.756 \\[-3pt]
    & Extent (10m-ours)                & 0.888 & 0.888 & 0.888 & \textbf{0.799} \\[-3pt]
    & Extent (10m-\citet{kerner2024fields}) & -- & -- & -- & 0.670 \\ \cmidrule{2-6}
    & Boundary (2.5m-ours)             & 0.204 & 0.123 & 0.154 & 0.083 \\[-3pt]
    & Boundary (10m-ours)              & 0.569 & 0.450 & 0.503 & 0.336 \\ \cmidrule{2-6}
    & PoLiS (2.5m-ours)                & \multicolumn{4}{c}{ 19.54m} \\[-3pt]
    & PoLiS (10m-ours)                 & \multicolumn{4}{c}{20.25m} \\
    \bottomrule
\end{tabular}
\end{table*}

The \textsc{AI4SmallFarms} tiles for \textbf{Cambodia} and \textbf{Vietnam} are also present in \textsc{FoTW}, yet the two repositories differ in both \emph{spectral} and \emph{temporal} content:  
\textsc{FoTW} provides \emph{two} cloud–free Sentinel-2 (S2) composites at 10m, whereas the original \textsc{AI4SmallFarms} dataset offers only \emph{one} S2 composite--but with access to the full reference polygon files. In addition we have acquired the Sentinel-1 (S1) time series for the cropping season that refers to the data.   
These two regions therefore combine (i) \textbf{persistent cloud cover} and (ii) \textbf{sub-hectare paddies}, making them an ideal stress-test for PTAViT3D-CA with S2 and SAR fusion. Given the extra challenge of sub-hectare paddies, and since we have already modeled with original S2 resolution this dataset (FoTW), we resort to super resolution (SR) for upscaling the original S2 image. This upscaled S2 image will be fused with SAR for modeling with PTAViT3D-CA.

To accommodate the asymmetric inputs -- one super-resolved S2 image versus a multi-temporal S1 stack -- we remove the fusion branch of the dual encoder PTAViT3D-CA.  First, each 10m Sentinel-2 composite is up-sampled offline by a factor of four with the open sourced LDSR-S2 model~\citep{10887321}, yielding a single 2.5m S2-SR tile.  The complete Sentinel-1 time series (upscaled with linear interpolation to match the spatial dimensions of S2-SR) forms the secondary branch; its features interact with those of S2 solely through Patch Tanimoto Attention 3D, avoiding any explicit feature concatenation and thus sidestepping temporal mis-alignment issues.  At every encoder depth, the S2-SR feature map \(f_{S_2^{\mathrm{SR}}}\in\mathfrak{R}^{B\times N\times 1\times H\times W}\) is enriched by the attention map computed against the S1 features \(f_{S_1}\in\mathfrak{R}^{B\times N\times T\times H\times W}\):
\[
f_{S_2^{\mathrm{SR}}} \;\leftarrow\; f_{S_2^{\mathrm{SR}}}\bigl(1+\mathcal{A}(f_{S_1},f_{S_2^{\mathrm{SR}}},f_{S_2^{\mathrm{SR}}})\bigr),
\]
where \(\mathcal{A}(\cdot)\) denotes PTA3D (Eq. \eqref{Main-Attention_map}).  We note that the S2-SR image features are used as key-value pairs. Given that \(\mathcal{A}\) contracts the time dimension of the query features \(f_{S_1}\) during the query–key interaction, its output is dimensionally compatible with \(f_{S_2^{\mathrm{SR}}}\) (see Fig.~\ref{Main-mantis_v2_base_arch}).  In practice this design lets the cloud-immune temporal cues in S1 refine the high-resolution spatial detail recovered from S2-SR.

Beginning with the native 10m Sentinel-2 composites, we first apply LDSR-S2 to obtain a $4\times$ super-resolved image at 2.5m.  The ground-truth parcel polygons are rasterised on this high-resolution grid, and we again construct the extent, bounds and distance transform (per parcel).  We then feed the resulting S2-SR image, together with the full Sentinel-1 time series, into the joint S2-SR\& S1 variant of PTAViT3D-CA , whose multitask heads output \emph{extent}, a distance-transform map and auxiliary edge logits. We also train a pure PTAViT (2D) model with the S2-SR input for comparison with the fused model S2-SR\&\,S1.   

During inference, the predictions logits are used in two ways, either in the predicted 2.5m resolution or they are down-sampled to the original 10m resolution—using the mean for the \textit{extent} channel and the maximum for the \textit{boundary} channel—so that predictions remain directly comparable with the pixel-level evaluation metrics.

Given the clear superiority of S2-SR\& S1 model over the pure S2-SR one (Fig.~\ref{Main-AI4SmallFarmsS2S1Performance}), we proceed in test set evaluation only with the former (Table~\ref{Main-tab:ai4sf_metrics}). 
The S2-SR\&\,S1 model achieves IoU $92.5\%$ for Cambodia and $79.9\%$ for Vietnam. The corresponding boundary F1 score (10m resolution) of 52.0\% improves  on the baseline ($39\%$) reported by \citet{10278130} by 13.0 percentage points.
For the case of the 2.5m resolution (super resolution imagery) the Extent metrics remain all high, however the Boundary metrics (corresponding to a single pixel thickness) fall to low values, much lower than the corresponding 0.5m Google Maps (GM) results reported from  \citet{10278130}.  However, the PoLiS distance remains high both in 10m and 2.5m resolution, surpassing the \citet{10278130} baseline, suggesting that the problem with the low score in boundary pixels is mainly owed to our model missing the 1 pixel width of boundaries for the 2.5m case. Indeed during training the reference boundaries are deliberately rasterised with a width of three pixels to stabilise optimisation. When these wide predictions are morphologically thinned back to a one-pixel skeleton for evaluation, any localisation error larger than half a pixel propagates outward, effectively inflating the positional bias at the higher resolution. The fact that parcel \emph{area} metrics remain excellent at 2.5m supports the view that the model is operating in an error-tolerant regime with respect to boundary thickness, rather than misclassifying large regions.

\begin{figure*}[!tp]
\begin{center}
\includegraphics[clip, trim=0.0125cm 0.0130cm 0.0135cm 0.013cm,width=\textwidth,height=0.72\textheight,keepaspectratio]{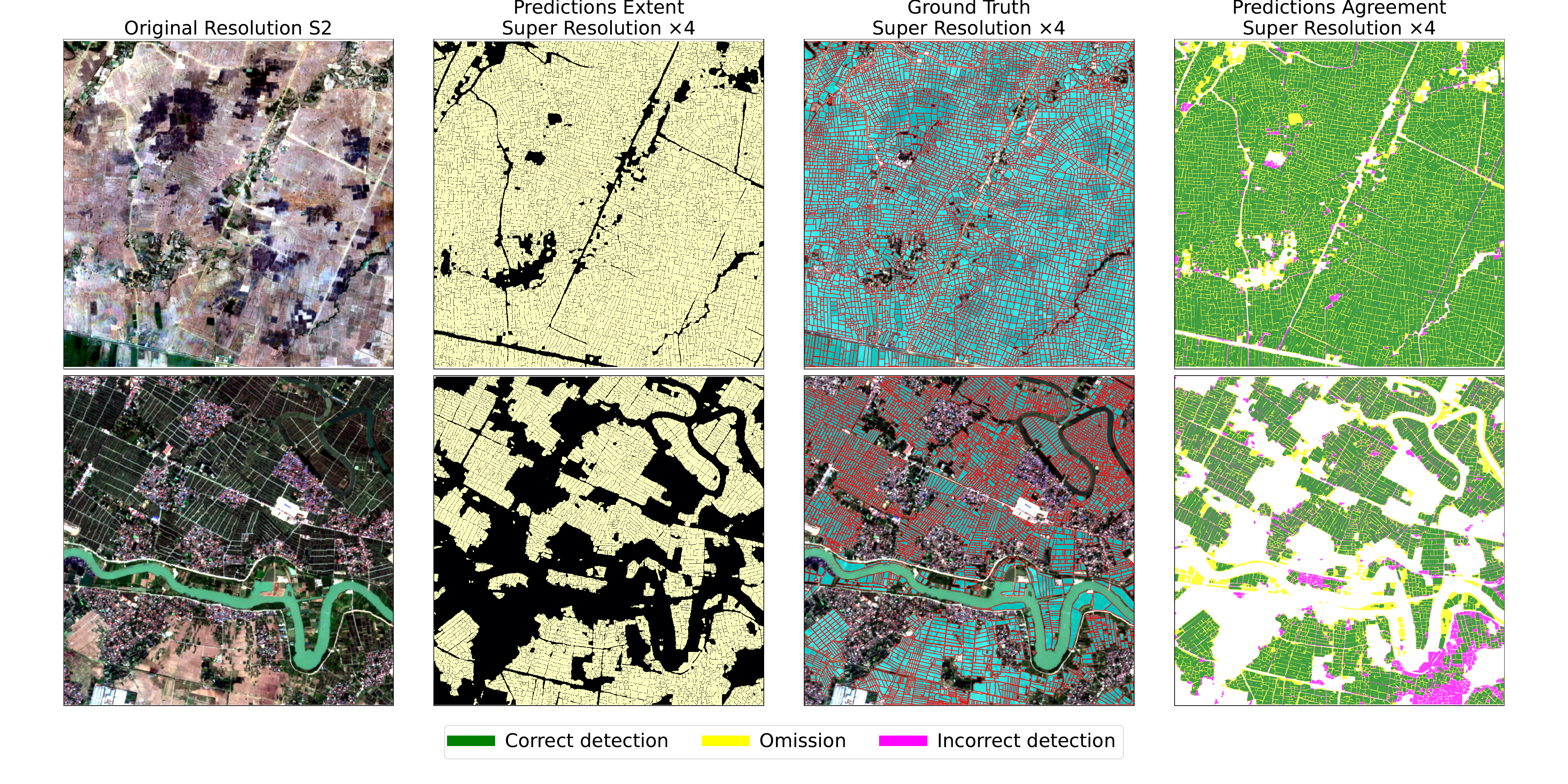}
\end{center}
\caption{Sample performance from the test set on tiles Cambodia 12 (top) and Vietnam 27 (bottom).}
\label{Main-AI4SmallFarmsS2S1Performance_VietCamb}
\end{figure*}

\begin{figure*}[!tp]
\begin{center}
\includegraphics[clip, trim=0.0125cm 0.0130cm 0.0135cm 0.013cm,width=\textwidth,height=0.72\textheight,keepaspectratio]{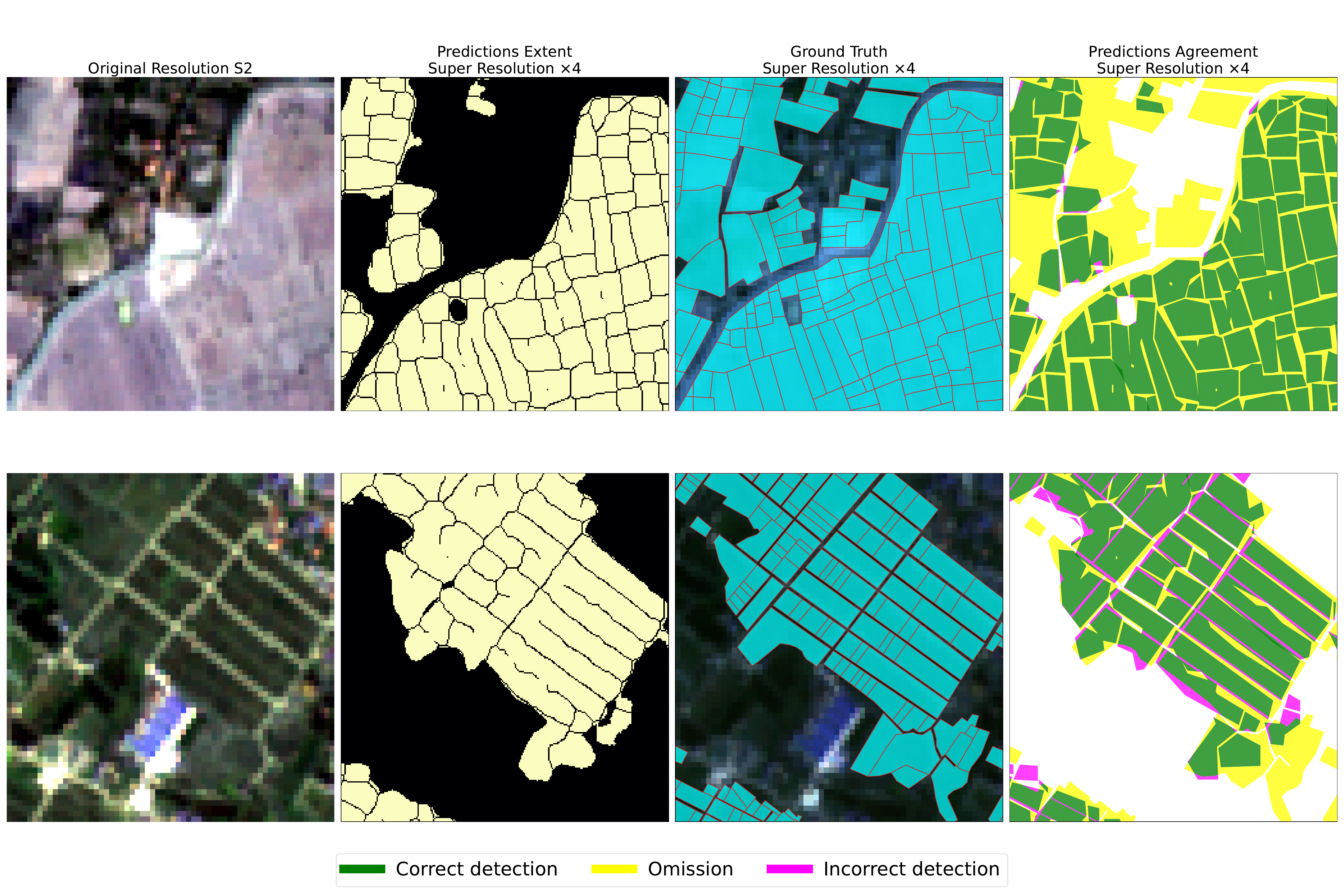}
\end{center}
\caption{As Fig.~\ref{Main-AI4SmallFarmsS2S1Performance_VietCamb}, zoomed in random regions on tiles Cambodia 12 (top) and Vietnam 27 (bottom).}
\label{Main-AI4SmallFarmsS2S1Performance_VietCambZOOM}
\end{figure*}

For the  case of the S2-SR\&\,S1 model, in Fig.~\ref{Main-AI4SmallFarmsS2S1Performance_VietCamb} we present qualitative results for the two test tiles of Cambodia 12 (top) and Vietnam 27 (bottom). From left to right we plot for each row: the Original resolution S2 (10m) image, the refined extent masks that our model predicts, the ground truth at super resolution of 2.5m of polygons and finally, the agreement between predicted polygons and ground truth. Fig.~\ref{Main-AI4SmallFarmsS2S1Performance_VietCambZOOM} is the same, however with zoomed in regions so interested readers can evaluate qualitatively the boundaries results.

\subsection{PTAViT3D continental scale application: the ePaddocks National product of Australia}
\label{Main-ptavit3d_epaddocks}

\begin{figure*}[!tp]
\begin{center}
\includegraphics[clip, trim=0.025cm 0.0cm 0.05cm 0.01cm,width=\textwidth,height=0.72\textheight,keepaspectratio]{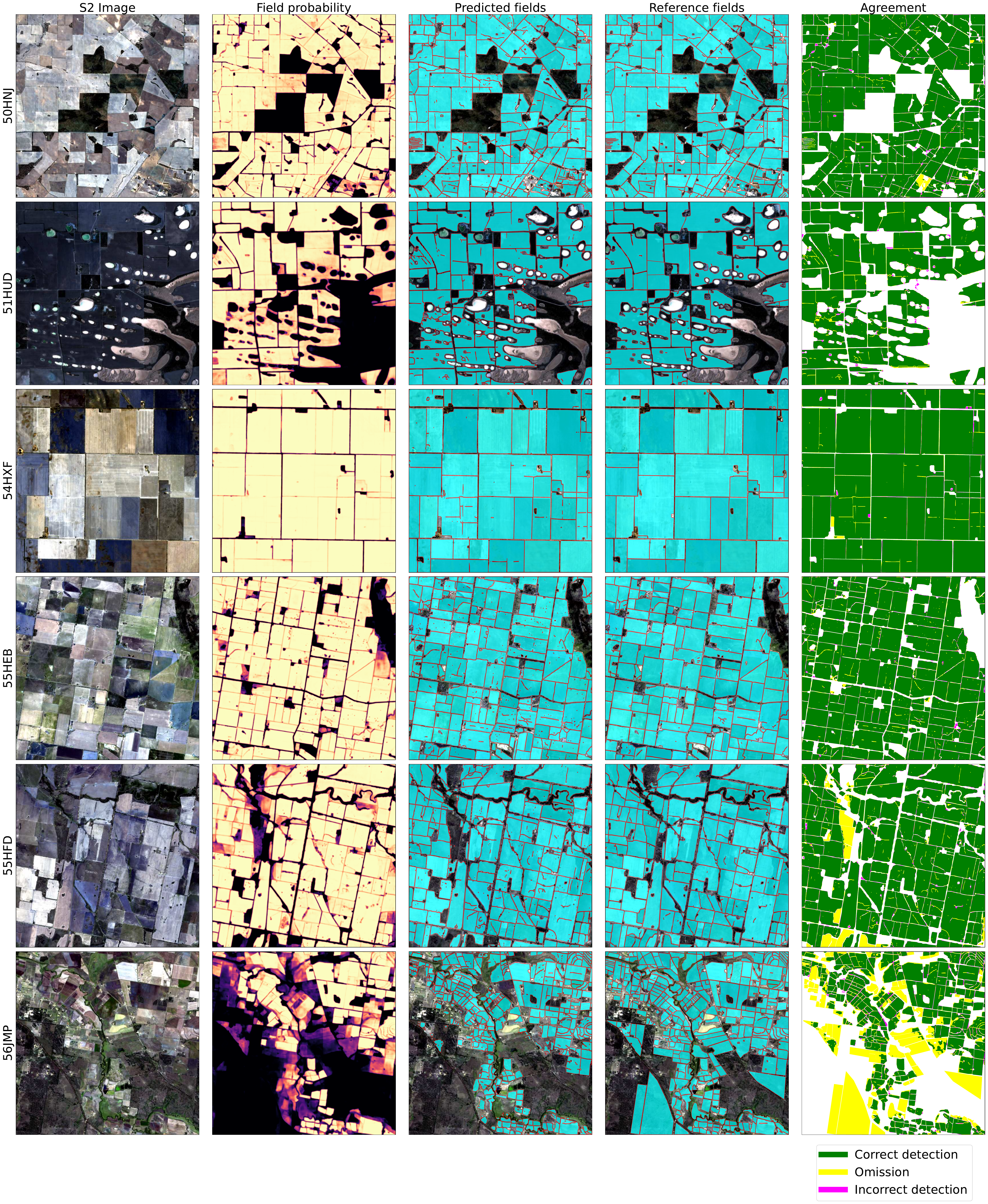}
\end{center}
\caption{Example of Instance Segmentation for the field boundaries in six test sites (year 2023). In the agreement column, white space indicates agreement in background.}
\label{Main-TestTiles_instanceSegm}
\end{figure*}

\begin{figure}[!htbp]
\begin{center}
\includegraphics[clip, trim=0.025cm 0.0cm 0.05cm 0.01cm,width=\columnwidth,height=0.72\textheight,keepaspectratio]{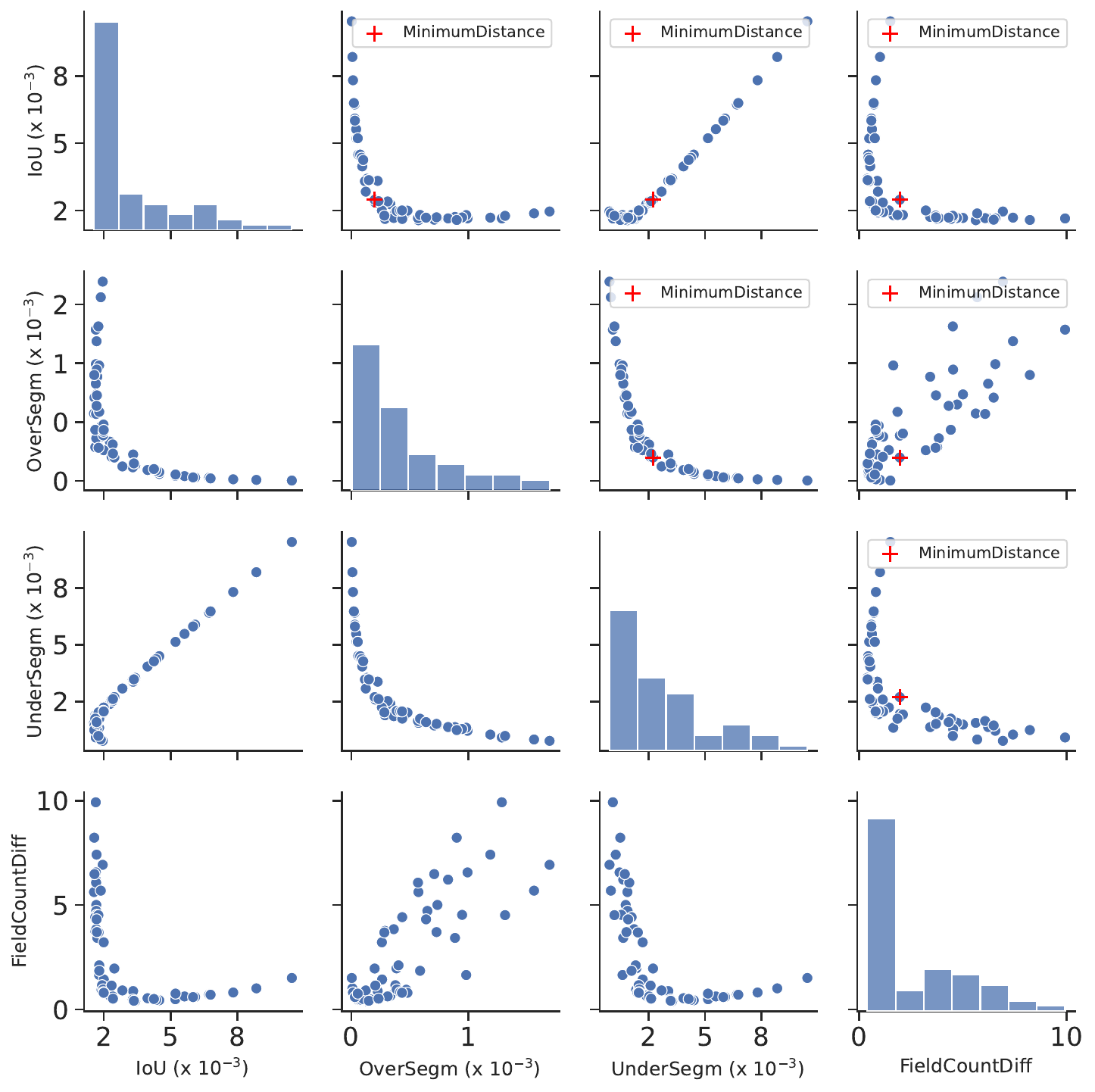}
\end{center}
\caption{Pareto front of the multiobjective optimization for the estimation of instance segmentation hyperparameters.}
\label{Main-pareto_front}
\end{figure}

In this section we provide a detailed roadmap on applying the PTAViT3D algorithm at continental scale. We also utilize the results to compare with previous established baselines \citep{WALDNER2020111741,rs13112197}. To construct the National Product (ePaddocks) of Australia for 2023, we followed these steps:
\begin{enumerate}
    \item We initially pretrained the PTAViT3D on six tiles from Australia using 2019 data, the same data used for creating ePaddocks 2019 (Figure \ref{Main-TrainingSites}).
    \item We then sourced new imagery for the same date range (2019) using Open Data Cube (ODC\footnote{\href{https://www.opendatacube.org/}{https://www.opendatacube.org/}}) and fine-tuned the algorithm on these new images, ensuring they had $\leq$10\% cloud cover. This mimicked the expected conditions during inference, as opposed to the clear rasters used for initial pretraining. It also enriches the algorithm by exposing it to greater variability of input imagery during training.
    \item Inference for the year 2023 was performed across 293 tiles, as shown in Figure \ref{Main-TrainingSites}. This process created raster products that combined field extent, boundaries, and distance transform.
    \item To transition from semantic segmentation to instance segmentation, we post-processed the data (Section \ref{Main-post_processing}) using set operations, thresholding, and extracted polygons with the \texttt{rasterio} \citep{gillies_2019} features module. 
We identified optimal thresholds for the raster to polygon algorithm (\textcolor{red}{Supplementary Material, Section~\ref{SM-alg:raster2polygons}}) using      the pareto front for the following metrics, $IoU$, $FDR$ (oversegmentation), $FOR$ (undersegmentation) and the difference in counts of total number of fields. The optimum choice (Figure \ref{Main-pareto_front}) was selected for the point of the pareto front closest to the origin ($\mathbf{0}$) of the coordinate system, that represents the ideal minimum. The optimization was performed with the library \texttt{optuna} \citep{10.1145/3292500.3330701} and the best choice of thresholds was found to be $t_b\approx 0.2$ and $t_e \approx 0.4$. 
    \item Further post-processing was done to remove inference results smaller than 100m$^2$ and to simplify polygons with a 10m (1 pixel) tolerance.
    \item The test set was validated on 67 small raster patches, each approximately 100km$^2$, which were manually annotated by our internal CSIRO team.
\end{enumerate}

Qualitative indicative performance on the test set of 2023 can be seen in Figure \ref{Main-TestTiles_instanceSegm}, where we visualize from left to right, a single date of S2 imagery, the field pseudo-probability, the predicted polygon fields, the ground truth fields and the agreement between them.

\subsubsection{ePaddocks test set evaluation (product-level raster \& polygon metrics)}
\label{Main-ePaddocks_walkthrough_test_Set_polis}

The ePaddocks test set evaluation was conducted using 67 test tiles, each approximately 100km\(^2\), across the cropping zones of Australia (Section \ref{Main-s2_section_data}). The evaluation is reported in two complementary views: (i) raster segmentation performance and (ii) polygon-level geometric agreement after raster-to-polygon conversion.

This evaluation follows the metric definitions introduced in Section~\ref{Main-evaluation_methods_all}. Specifically, we report raster-level semantic segmentation metrics (IoU, FDR, FOR; Section~\ref{Main-segmentation_metrics_section}--\ref{Main-segmentation_for_fdr_metrics_section}) computed from the pixel confusion matrix, and we additionally report polygon-level boundary agreement metrics (MSD and Hausdorff; Section~\ref{Main-boundary_metrics_section}) computed after vectorization and one-to-one polygon matching. We separate these two views because the ePaddocks product is delivered as a vector polygon dataset: raster metrics quantify segmentation accuracy, while polygon metrics quantify geometric fidelity of the final cartographic output.

\begin{figure}[!htbp]
\begin{center}
\includegraphics[clip, trim=0.25cm 0.0cm 0.05cm 0.1cm,width=\columnwidth,height=0.72\textheight,keepaspectratio]{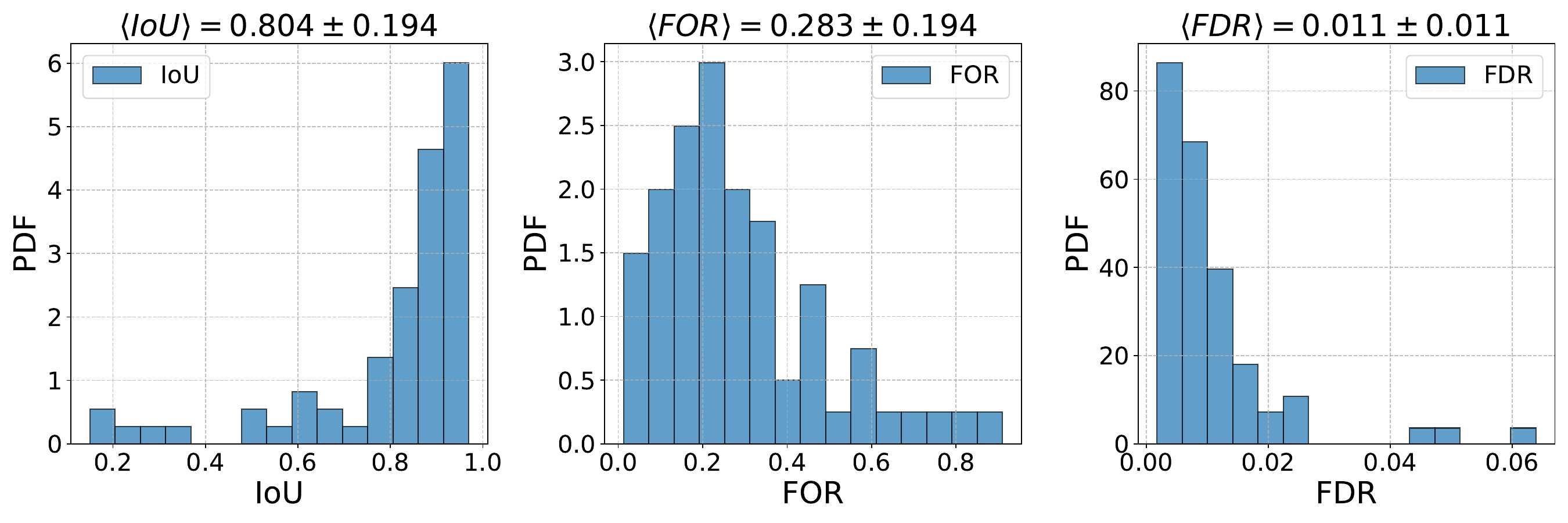}
\end{center}
\caption{Segmentation performance on the Test set, model trained on S2 time series. These metric are evaluated from raster data, not polygons.}
\label{Main-segm_metrics}
\end{figure}

\paragraph{Segmentation Performance (Raster Data)}
The segmentation performance was assessed using raster data with the following metrics (Figure \ref{Main-segm_metrics}):

\textbf{Intersection over Union (IoU):} The mean IoU was $0.804 \pm 0.194$, indicating a high degree of overlap between the predicted and actual field boundaries.

\textbf{False Omission Rate (FOR):} The mean FOR was $0.283 \pm 0.194$, reflecting the extent of under-segmentation in the predictions. The increased omission rate primarily stems from the 1-pixel buffer in the boundaries we evaluated based on ground truth. This buffer leads to more conservative field boundaries (Fig. \ref{Main-TestTiles_instanceSegm}), which is a result of the raster-to-polygon strategy we followed. Although this strategy has the drawback of more conservative boundaries, it also ensures that boundary fields do not intersect.

\textbf{False Discovery Rate (FDR):} The mean FDR was $0.011 \pm 0.011$, showcasing an extremely low rate of false positives. This indicates that the model rarely identified non-field areas as fields, resulting in excellent performance for false discovery.

These metrics demonstrate that the segmentation model performs well, with high accuracy in identifying field boundaries and a minimal number of false positives. The conservative boundaries strategy leads to higher omission rates but prevents overlapping boundaries.

\begin{figure}[!htbp]
\begin{center}
\includegraphics[clip, trim=0.25cm 0.0cm 0.05cm 0.1cm,width=\columnwidth,height=0.72\textheight,keepaspectratio]{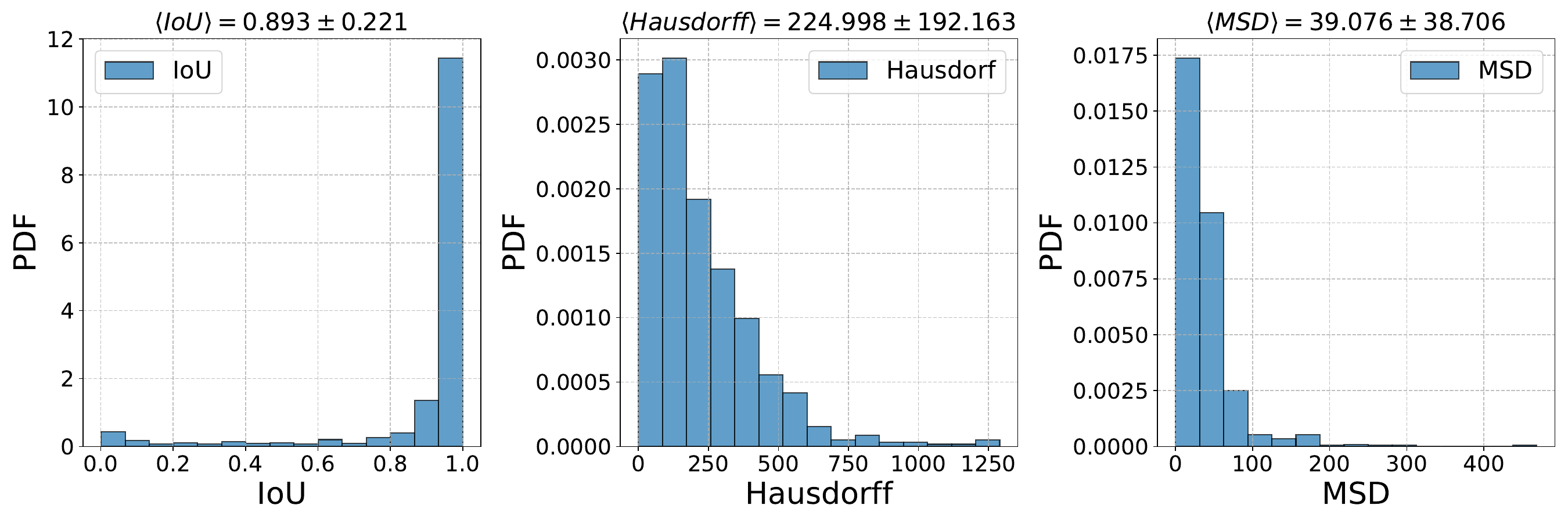}
\end{center}
\caption{Segmentation performance on the Test set, model trained on S2 time series. These metric are evaluated from Polygons not rasters.}
\label{Main-polygon_metrics}
\end{figure}

\paragraph{Polygon Matching Performance}
The performance of the polygon matching was evaluated for polygons with an IoU of at least 0.001 using the following metrics (Figure \ref{Main-polygon_metrics}):

\textbf{Intersection over Union (IoU):} The mean IoU was $0.893 \pm 0.221$, indicating a higher degree of overlap when compared to the raster evaluation. The difference between the IoU value estimated from segmentation predictions and polygons (0.893 vs 0.804 for rasters) is due to two main reasons: (a) the IoU for polygons is computed as the average IoU across individual polygons rather than across tiles, resulting in a different computation as it is per object; (b) there is a small selection bias, as matches with an IoU smaller than $10^{-3}$ were excluded to ensure meaningful matches. 

\textbf{Hausdorff Distance:} The mean Hausdorff Distance was 
$224.998 \pm 192.163$, measuring the greatest distance between the predicted and actual field boundaries. This large variance reflects the variability in the complexity of field shapes across the test set.

\textbf{Mean Surface Distance (MSD):} The mean MSD was $39.076 \pm 38.706$, which provides an average measure of the distance between the surfaces of the predicted and actual polygons. This is approximately 4 pixels, or 40m on average. 

These metrics indicate that the model's polygon predictions align closely with the actual field boundaries, although the Hausdorff Distance suggests some variability due to the diverse shapes of the fields.

\begin{figure}[!htbp]
\begin{center}
\includegraphics[clip, trim=0.025cm 0.0cm 0.05cm 0.01cm,width=\columnwidth,height=0.72\textheight,keepaspectratio]{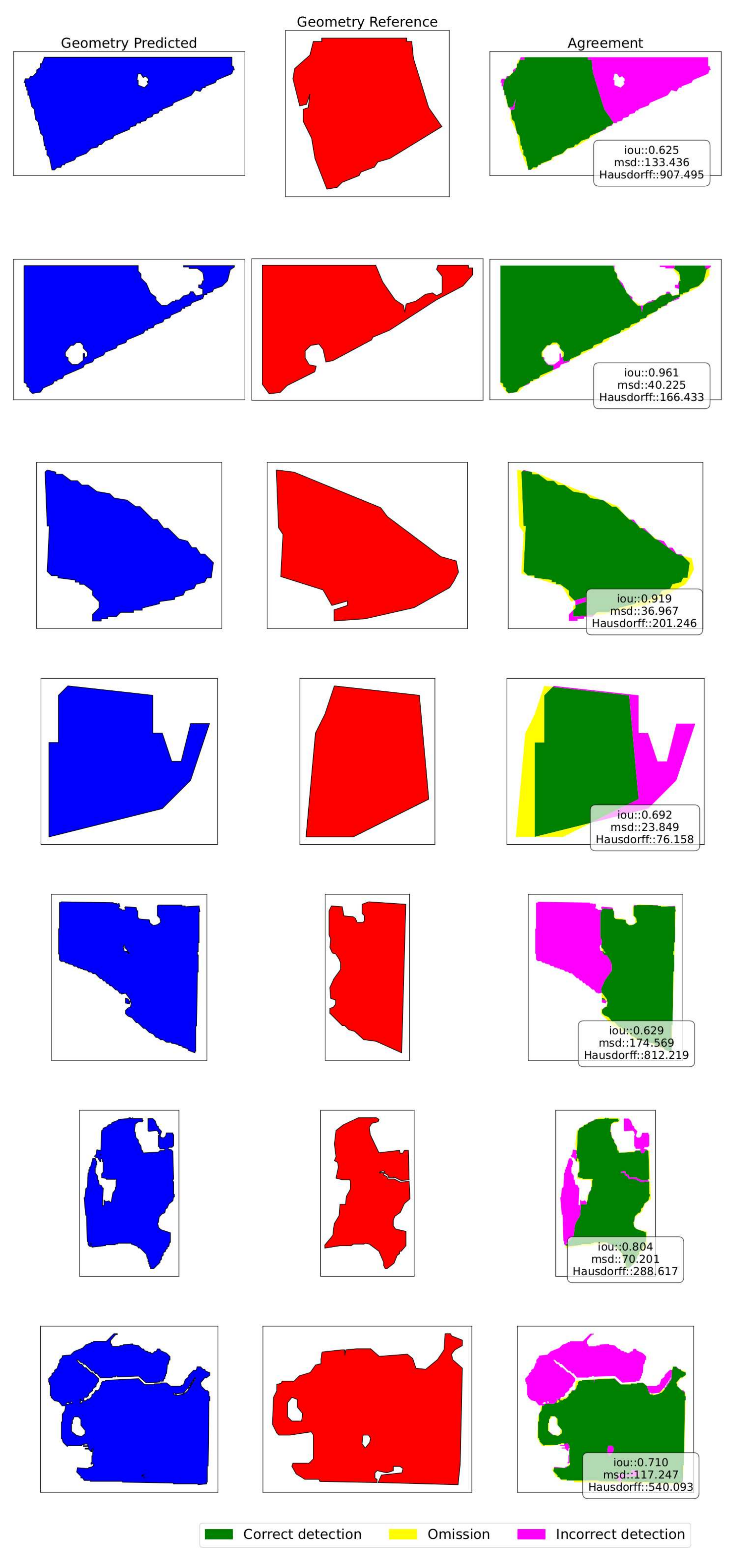}
\end{center}
\caption{Agreement per field example with representative boundary metrics}
\label{Main-agreement_per_field}
\end{figure}

Figure \ref{Main-agreement_per_field} (see also Figure \ref{Main-TestTiles_instanceSegm}) illustrates the agreement per field example with representative boundary metrics.  These examples underscore the model's ability to accurately delineate field boundaries while minimizing errors, as evidenced by the low FDR.

The ePaddocks test set evaluation confirms the robustness and accuracy of the proposed model in detecting field boundaries from Sentinel-2 time series imagery. The results demonstrate that the model performs effectively in both raster-based segmentation and polygon-based boundary delineation, making it a reliable tool for field boundary mapping in diverse agricultural landscapes. The conservative boundary delineation strategy results in higher omission rates but ensures non-intersecting fields, which is a significant advantage for practical applications.

\section{Discussion}

This study introduced and evaluated PTAViT3D and PTAViT3D-CA for field boundary delineation from Sentinel-1 (S1) and Sentinel-2 (S2) satellite image time series. The main result is that treating the problem as explicitly spatio-temporal is beneficial in practice. Rather than relying on a single optical composite or on manual temporal consolidation, the proposed models consume the image time series directly and learn how to use the available observations. This is particularly relevant in agricultural settings, where planting and harvest timing, crop-dependent growth, and management effects often make the boundary signal inherently temporal.

The benchmark experiments further show that the proposed approach generalizes across different agricultural systems and observation conditions. On FoTW, PTAViT3D performs strongly across a large and geographically diverse benchmark. On PASTIS, it achieves competitive performance with only a fraction of the available S2 observations. On AI4SmallFarms, multimodal fusion is particularly useful in fragmented smallholder environments with persistent cloud cover. Taken together, these results support the view that the proposed models are not tied to a single geography or patch configuration, but remain effective across different farm structures, climates, and sensing conditions.

Taken together, Table~\ref{Main-how_much_cloud}, Table~\ref{Main-tab:evaluation_metrics}, Fig.~\ref{Main-compare_all_models}, and Fig.~\ref{Main-AI4SmallFarmsS2S1Performance} clarify the contribution of different input modalities under practically relevant cloud regimes. Under moderate cloud contamination, Sentinel-2 time series remains the strongest source of information. As optical coverage deteriorates, Sentinel-1 becomes increasingly valuable, either as a standalone input or in fusion with limited Sentinel-2 observations. This is evident on ePaddocks, where S1-only PTAViT3D remains competitive with the established S2 baseline, and on AI4SmallFarms, where adding a Sentinel-1 time series to a single Sentinel-2 image improves validation mIoU by about 6--7 points. These results should therefore be interpreted as modality-level contributions under operational cloud regimes, rather than as a controlled feature-attribution study with cloud level varied independently of temporal coverage.

\textcolor{red}{The controlled companion experiments reinforce the benefit of joint spatio-temporal modelling. Under the severe cloud/shadow summary, cloud-augmented PTAViT3D attains object F1 of 0.698, recall of 0.816, and PoLiS of 43.93~m; the 16-date stacked 2D control attains 0.693, 0.758, and 46.34~m, respectively, while per-date averaging falls to object F1 of 0.525 and PoLiS of 73.27~m (Supplementary Table~\ref{SM-tab:app-2d3d-severe}). In the full FoTW attention replacement, fresh native-resolution validation gives PTAViT3D an mIoU of 0.8786 versus 0.8751 for support-matched window attention, while the evaluated configuration packs the same effective training batch of 32 on two rather than eight H100 GPUs (Supplementary Tables~\ref{SM-tab:supp-attention-fullmodel-native256}--\ref{SM-tab:supp-attention-fullmodel-resources}).}

\textcolor{red}{Applying the same cloud/shadow perturbation schedule to four PASTIS source tiles separates regional sensitivity from perturbation severity. Thin cloud is the most damaging severe perturbation in every tile, and mean \(\Delta_{\mathrm{cloud}}\mathrm{IoU}\) across the four severe endpoints decreases from 0.036 for native training to 0.015 with cloud-augmented training (Supplementary Table~\ref{SM-tab:app-regional-cloud-stress}). This is a controlled regional response to simulated within-sequence contamination frequency, not an estimate of regional cloud climatology.}

\textcolor{red}{The controlled perturbations and operational Open Data Cube experiment provide complementary evidence: the former isolates cloud and shadow effects under a fixed temporal registry, while the latter evaluates real imagery in which scene contamination and temporal availability vary together. Synthetic perturbations do not reproduce every persistent-cloud morphology or regional acquisition pattern, so new deployment domains still require local validation; where usable optical observations are insufficient, S1 provides the all-weather input alternative evaluated here.}

The continental-scale ePaddocks application is also an important part of the contribution. The manuscript does not only report benchmark results, but an operational deployment for the generation of the 2023 ePaddocks product across Australia's cropping regions. This demonstrates that the proposed models can be used in a full production workflow, from time-series sourcing and model inference to post-processing into polygon instances. In that sense, the work contributes not only a model, but also a practical roadmap for large-scale deployment.

\textcolor{red}{The acquisition window should span the relevant cropping cycle because the model uses chronological order rather than explicit elapsed-time features. It predicts one parcel configuration per stack: multiple crops are compatible while parcel geometry remains unchanged, whereas a boundary reconfiguration requires a separate season-consistent stack and reference label.}

Several avenues remain for future work. These include end-to-end polygon inference to reduce post-processing, improved geographical conditioning, and further integration of super-resolution methods in smallholder settings. It would also be useful to explore the same spatio-temporal formulation for related tasks such as crop-type classification and broader agricultural monitoring problems.

\section{Conclusion}

We presented PTAViT3D and PTAViT3D-CA, two deep learning architectures for field boundary delineation from Sentinel-1 and Sentinel-2 image time series. The central idea is to process the temporal stack directly, instead of relying on single composites or manual temporal consolidation. Across ePaddocks, FoTW, PASTIS, and AI4SmallFarms, the proposed models showed robust and competitive performance under diverse agricultural and observational conditions.

The results also clarify the relative role of the input sources. Sentinel-2 time series remains the strongest source when optical observations are sufficiently available, while Sentinel-1 becomes increasingly important as optical coverage becomes sparse or unreliable. In this setting, multimodal fusion provides a practical advantage, especially in cloud-affected environments.

Finally, the 2023 ePaddocks application shows that the method is not only effective on public benchmarks, but also suitable for continental-scale operational deployment. We therefore view PTAViT3D as a practical step toward scalable and robust field boundary delineation from satellite image time series. Future work will focus on polygon-level inference, geographical feature integration, and broader applications in agricultural and environmental monitoring.

\section*{Acknowledgments} 
This project was supported by resources and expertise provided by CSIRO IMT Scientific Computing.

\FloatBarrier
\bibliography{AI_BIB}
\clearpage

\clearpage
\onecolumn

\fontsize{11}{13.2}\selectfont
\setcounter{section}{0}
\setcounter{subsection}{0}
\setcounter{footnote}{0}
\renewcommand{\theHsection}{supp.\arabic{section}}
\renewcommand{\theHfigure}{supp.\arabic{figure}}
\renewcommand{\theHtable}{supp.\arabic{table}}
\renewcommand{\theHequation}{supp.\arabic{equation}}
\renewcommand{\theHfootnote}{supp.\arabic{footnote}}
\section*{Supplementary material}
\addcontentsline{toc}{section}{Supplementary material}

\renewcommand{\thefigure}{S\arabic{figure}}
\renewcommand{\thetable}{S\arabic{table}}
\renewcommand{\thesection}{S\arabic{section}}
\renewcommand{\thesubsection}{S\arabic{section}.\arabic{subsection}}
\renewcommand{\thesubsubsection}{S\arabic{section}.\arabic{subsection}.\arabic{subsubsection}}
\renewcommand{\theequation}{S\arabic{equation}}
\setcounter{figure}{0}
\setcounter{table}{0}
\setcounter{equation}{0}

\section*{Overview}
This supplementary material contains the additional processing details, controlled ablation experiments, model characteristics, computational notes, and algorithm listings that support the main manuscript. The controlled ablation study is separated from the operational experiments in the main manuscript so that cloud contamination, regional consistency, two-dimensional temporal alternatives, per-date aggregation baselines, attention replacement, and loss-function controls can be read as one matched experimental protocol.

\section{Sentinel-1 processing}
\label{SM-s1_processing_appendix}

The Entropy/alpha decomposition algorithm was originally developed to simplify the multi-parameter depolarisation issues using statistical method for quad-pol radar backscatter. It can also be applied to the simpler case of dual polarisation. In this latter scenario the radar transmits only a single polarisation and receives, either coherently or incoherently, two orthogonal components of the scattered signal. This corresponds (in the coherent case) to a measurement of the full state of polarisation of the scattered signal for fixed illumination. The current space-borne SAR sensors including Sentinel-1 provide at least a “partial polarimetric mode”, acquiring only 2 of the 4 elements of the Sinclair matrix, for example, HH and HV or VV and VH:
\begin{equation}
\mathbf{S} = 
\begin{bmatrix}
    S_{HH} & S_{HV} \\
    S_{VH} & S_{VV}
\end{bmatrix}.
\end{equation}
  The coherent dual-pol Sentinel-1 provides the potential to investigate the development and application of a dual polarised entropy/alpha technique that can be used to take advantage of such coherent dual polarised systems. These radars can be used to estimate the 2x2 wave coherency matrix \citep{2007ESASP.644E...2C}:
\begin{equation}
\mathbf{J}_H = 
\begin{bmatrix}
    \left\langle S_{HH} S_{HH}^\ast \right\rangle & \left\langle S_{HH} S_{HV}^\ast \right\rangle \\
    \left\langle S_{HV} S_{HH}^\ast \right\rangle & \left\langle S_{HV} S_{HV}^\ast \right\rangle
\end{bmatrix}
\end{equation}
or 
\begin{equation}
\mathbf{J}_V = 
\begin{bmatrix}
    \left\langle S_{VV} S_{VV}^\ast \right\rangle & \left\langle S_{VV} S_{VH}^\ast \right\rangle \\
    \left\langle S_{VH} S_{VV}^\ast \right\rangle & \left\langle S_{VH} S_{VH}^\ast \right\rangle
\end{bmatrix}.
\end{equation}
Using the standard interpretation of normalised eigenvalues, $P_i=\lambda_i/(\lambda_1+\lambda_2)$, of $\mathbf{J}$ as probabilities, together with the fact that in 2$\times$2 problems the second eigenvector can be derived from the principal eigenvector using orthogonality, we obtain an entropy/alpha parameterisation of the wave coherency matrix, $\mathbf{J}$, as shown in Equations \ref{SM-ZS_eq41} and \ref{SM-ZS_eq42}: 
\begin{align}
\notag
\mathbf{J} &= 
\begin{bmatrix}
    J_{xx} & J_{xy} \\
    J_{xy}^* & J_{yy}
\end{bmatrix} \Longrightarrow \\
\mathbf{U}_2 &= 
\begin{bmatrix}
    \cos \alpha & -\sin \alpha e^{-i\delta} \\
    \sin \alpha e^{i\delta} & \cos \alpha
\end{bmatrix}
\notag
\\
\notag
\mathbf{D} &= (\lambda_1+\lambda_2)
\begin{bmatrix}
    P_1 & 0 \\
    0 & P_2
\end{bmatrix}
\end{align}
Expanding the last two equations further yields: 
\begin{align}
\label{SM-ZS_eq41}
\bar{\alpha}_2 &= \alpha (P_1-P_2) + P_2 \frac{\pi}{2},\\
H_2 &= -\sum_{i=1}^2 P_i\log_2 P_i.
\label{SM-ZS_eq42}
\end{align}
Here $\bar{\alpha}_2$ and $H_2$ are the mean scattering angle and entropy for the dual-polarised case, respectively.

ESA’s SNAP software was used to pre-process all Sentinel-1 products. 
IW dual-pol images, in both SLC and GRD formats, were processed into analysis-ready data, including normalized radar backscatter coefficients ($\Gamma_0$) and polarimetric decomposition parameters, on CSIRO’s HPC systems \citep{data4030100,zhou2023polarimetric}. The following steps were applied to each dual-pol VV+VH GRD data to generate normalized radar backscatter products: applying the orbit file, removing thermal noise, removing GRD border noise, radiometric calibration, terrain flattening, speckle filtering, multi-looking, and terrain geometric correction. Concurrently, the following steps were implemented for each dual-pol VV+VH SLC data to generate polarimetric decomposition products: applying the orbit file, removing thermal noise, performing radiometric calibration, executing S1 TOPS deburst and merge, calculating polarimetric metrics, applying polarimetric speckle filtering, multi-looking, performing polarimetric decomposition, and terrain geometric correction. The complete workflows for both backscatter and decomposition processing using SNAP on a high-performance computing platform are described in \citep{zhou2023polarimetric}. In particular, the implementation of thermal noise removal  \citep{2022_Mascolo}  and the use of the Copernicus DEM within the data processing have significantly improved the quality of the data for analysis purposes.

\begingroup
\section{Controlled ablation study}
\label{SM-app:controlled-ablation-study}

This controlled ablation study complements the main manuscript's operational experiments. The operational cloud-tolerance experiment (Section~\ref{Main-tfcl_general}) uses imagery available from the Open Data Cube under scene-level cloud-cover constraints. Here, the benchmark, temporal sample, perturbation registry, and evaluation protocol are fixed so that contamination frequency and severity, regional sensitivity, 2D temporal alternatives, per-date aggregation, attention replacement, and loss-function effects can be compared under controlled conditions. Sentinel-1-only and Sentinel-1/Sentinel-2 fusion controls remain in the main manuscript's sensor-modality analysis (Section~\ref{Main-tfcl_sensor_mod_s2vss1}), where both modalities are available; they are not duplicated in this Sentinel-2-only PASTIS study. The attention replacement instead uses FoTW, as described in Subsection~\ref{SM-supp:pta-vs-window-attention}.

PASTIS was originally developed for panoptic crop mapping and is repurposed here as a field-boundary benchmark because it provides dense Sentinel-2 time series and parcel-instance annotations \citep{garnot2021panoptic}. Crop identities are discarded and each parcel raster is converted into a binary, instance-aware target. Valid non-background, non-void pixels define field extent; transitions between parcel identifiers, including field--background transitions, define field boundaries. The supervision stack contains background and foreground channels for extent, boundary, and distance transform. Void pixels are excluded with the PASTIS valid mask. All PASTIS controls use the official five-fold split, a fixed 16-acquisition sample for each patch, and the same 2,433-patch evaluation registry per condition.

Throughout this supplementary material, \emph{native PASTIS} means the unmodified PASTIS input sequence, without an additional synthetic perturbation. It does not mean cloud-free imagery: the released PASTIS time series retain naturally cloud-affected acquisitions, and approximately 28\% of acquisitions remaining after the dataset's $>90\%$ cloud filter contain partial cloud cover \citep{garnot2021panoptic}. The controlled experiment therefore measures the incremental effect of SCG perturbations relative to an already realistic native input, rather than relative to an idealized cloud-free reference.

Synthetic optical contamination is generated in reflectance space with Satellite Cloud Generator version~0.4 \citep[SCG;][]{rs15174138}. Normalized inputs are returned to reflectance, divided by 10,000 and clipped to $[0,1]$ before perturbation, then rescaled and normalized for inference. The experimental parameter \texttt{spatial\_coverage}, rather than an SCG package argument, enforces spatial support. The temporal fraction $T_f$ selects $\operatorname{round}(16T_f)$ acquisitions without replacement, corresponding to 4, 8, 12, or 16 affected dates for $T_f=0.25$, 0.50, 0.75, or 1.00. For each affected image, SCG mask channels are averaged into a two-dimensional strength map; for the mixed condition, cloud and shadow strengths are first combined by their channel-wise maximum. The $\operatorname{round}(S_fHW)$ strongest pixels are then selected and used as a hard binary replacement support: SCG-disturbed reflectance is used inside the support and native PASTIS reflectance is retained outside it. Thus $S_f\in\{0.25,0.50,0.75\}$ controls the exact requested area fraction, equal to 4,096, 8,192, or 12,288 pixels on the $128\times128$ PASTIS grid. Requested and realized binary support are consequently identical. SCG's \texttt{max\_lvl} controls perturbation intensity, not spatial area: the ranges are 0.85--1.00 for thick cloud, 0.35--0.65 for thin cloud or haze, and 0.25--0.55 for shadow; the mixed cloud component uses 0.75--1.00. The same perturbation registry, generated from base seed 1729 and sample metadata, is reused across models for each condition. Here $T_f$ is a controlled affected-date proportion, not a meteorological occurrence rate, and $S_f$ is imposed replacement support, not an estimate of observed cloud cover.

\subsection{Ablation design and common protocol}
\label{SM-app:ablation-experimental-design}

Table~\ref{SM-tab:app-ablation-design-matrix} maps each scientific control question to its models, matched quantities, and permitted interpretation.

DA abbreviates Delineate Anything, the single-date RGB field-instance baseline described in Subsection~\ref{SM-app:da-baselines}. FTNMT denotes the multitask Tanimoto loss with complement used for PTAViT training. In the attention experiments, F128 and F256 denote inputs resized to $128\times128$ and inputs retained at the native $256\times256$ resolution, respectively.

\begin{table}[!htbp]
\centering
\caption{Controlled ablation design matrix. The table groups the supplementary experiments by the main control questions and states how each question is evaluated under matched conditions. Sentinel-1 and Sentinel-1/Sentinel-2 fusion ablations are reported in the main manuscript's sensor-modality analysis (Section~\ref{Main-tfcl_sensor_mod_s2vss1}) rather than repeated here.}
\label{SM-tab:app-ablation-design-matrix}
\fontsize{7.5}{9}\selectfont
\begingroup
\renewcommand{\arraystretch}{1.22}
\setlength{\tabcolsep}{4pt}
\begin{tabularx}{\textwidth}{>{\raggedright\arraybackslash}p{0.17\textwidth}>{\raggedright\arraybackslash}p{0.18\textwidth}>{\raggedright\arraybackslash}p{0.27\textwidth}>{\raggedright\arraybackslash}X}
\toprule
Control question & Models or controls & Matched protocol & Interpretation supported \\
\midrule
Controlled cloud and shadow robustness & PTAViT3D native; PTAViT3D cloud-augmented & PASTIS S2, 16-date tensor, 28 registered evaluation conditions (native reference, mixed-family no-op, and 26 non-zero perturbations), 2,433 patches per condition & Quantifies native-input retention and incremental degradation under thick cloud, thin cloud, shadow, and mixed cloud--shadow perturbations. \\
Regional sensitivity to simulated contamination frequency & Same PTAViT3D native and cloud-augmented predictions as the cloud-stress experiment & Same patch registry and identical \(T_f\), \(S_f\), and perturbation-type schedule; results grouped by native PASTIS Sentinel-2 source tile & Quantifies region-dependent accuracy sensitivity at equivalent simulated contamination frequencies and severities and tests whether cloud augmentation reduces degradation beyond a pooled score. \\
2D temporal alternatives & PTAViT2D stacked; PTAViT2D four-date; PTAViT2D per-date mean & Same PASTIS registry; object metrics on native PASTIS and severe stress summaries & Separates 3D temporal modelling from channel-stacked 2D input, a four-date 2D subset, and post-hoc averaging of per-date predictions. \\
Per-date 2D delineation baseline & Released DA checkpoint; DA native-PASTIS fine-tune; DA SCG-augmented fine-tune & Per-date RGB predictions aggregated by mean occupancy, vote, union, and intersection; all 16 dates used at inference & Tests whether a field-specific 2D instance model can recover time-series performance through post-hoc temporal aggregation. \\
Attention mechanism & PTAViT3D; same architecture with window attention & Primary: support-matched F128 fixed-runtime training and fresh F256 validation. Additional: practical native-F256 training with explicitly unmatched attention support. & Separates the controlled support-matched comparison from a descriptive native-resolution deployment configuration. \\
Loss function & FTNMT, BCE, BCE+Dice & Same native-PASTIS and severe mixed-cloud subset & Checks whether the cloud-stress conclusions are primarily a loss-function artefact. \\
Sensor/fusion scope & S2-only, S1-only, and S1/S2 fusion & Main-manuscript Section~\ref{Main-tfcl_sensor_mod_s2vss1}, Fig.~\ref{Main-compare_all_models}, and Fig.~\ref{Main-AI4SmallFarmsS2S1Performance} & Keeps sensor and fusion evidence on datasets with both modalities; it is not duplicated in the PASTIS S2 control. \\
\bottomrule
\end{tabularx}
\endgroup
\end{table}

\FloatBarrier

The PASTIS PTAViT controls use the common optimization protocol described below. Experiment-specific changes are listed in Table~\ref{SM-tab:app-training-provenance}. Training uses AdamW through a zero-redundancy optimizer, an initial learning rate of $10^{-5}$, weight decay $10^{-2}$, gradient clipping at 1.0, and automatic mixed precision. The learning rate warms linearly from $10^{-8}$ for 10 epochs and then follows cosine annealing with warm restarts, using a 50-epoch restart period, a 100-epoch half-life, and a minimum of $10^{-8}$.

The standard PASTIS configuration uses $128\times128$ inputs, 10 Sentinel-2 bands, $T=16$, FTNMT loss, and a 500-epoch cap. Three folds are used for training, one for validation, and one for testing in each of the five rotations. Training dates are randomly spaced and validation dates are linearly spaced. One spatial transform---horizontal or vertical flip, elastic deformation, grid distortion, or shift--scale--rotation---is sampled and replayed consistently across all dates, labels, and masks. Table~\ref{SM-tab:app-training-provenance} states the model-specific input, adaptation, resource, and checkpoint choices needed to interpret the results that follow.

\begin{table}[!htbp]
\centering
\caption{Training and checkpoint configuration for the controlled ablations. Epoch lists are ordered from Fold~1 to Fold~5. For PTAViT controls, settings not stated in a row follow the common PTAViT protocol described in the text; DA settings are reported in its row and Section~\ref{SM-app:da-baselines}.}
\label{SM-tab:app-training-provenance}
\fontsize{6.4}{7.8}\selectfont
\begingroup
\renewcommand{\arraystretch}{1.12}
\setlength{\tabcolsep}{3.2pt}
\begin{tabularx}{\textwidth}{>{\raggedright\arraybackslash}p{0.17\textwidth}>{\raggedright\arraybackslash}p{0.43\textwidth}>{\raggedright\arraybackslash}X}
\toprule
Control & Training and adaptation & Checkpoint and evaluation \\
\midrule
PTAViT3D native & Native PASTIS, $T=16$, 10 bands, $128\times128$ inputs, FTNMT, and five-fold training. & Selected epochs: 438, 432, 498, 499, and 403. \\
PTAViT3D SCG-augmented & Following an initial soft-mask cloud-augmentation phase, the final continuation enforces the exact hard support defined above for at most 10 additional epochs. Optimizer and scheduler states are retained. Four GPUs with 3 samples/GPU give total batch 12. Cloud family is sampled uniformly from mixed, thick, thin, and shadow; $T_f\in\{0,0.25,0.50,0.75\}$, $S_f\in\{0.25,0.50,0.75\}$, seed 1729. & Native-PASTIS validation-loss selection at epochs 494, 488, 500, 504, and 480, initialized from epochs 490, 479, 498, 497, and 474. \\
PTAViT2D stacked and per-date & Stacked input has 160 channels (10 bands $\times$ 16 dates). Per-date input has 10 channels; its training date cycles as $(\mathrm{patch\ index}+\mathrm{epoch})\bmod16$, and all 16 predictions are averaged at inference. Continuations used different accelerator packs, so no single effective batch describes every fold trajectory. & Validation mIoU with validation loss as tie-breaker. Selected epochs---stacked native: 310, 442, 462, 425, 449; stacked SCG: 486, 487, 454, 498, 440; per-date native: 429, 444, 319, 486, 494; per-date SCG: 392, 478, 429, 470, 462. \\
PTAViT2D four-date & Fine-tuned from the stacked checkpoints with optimizer reset and learning rate $10^{-5}$. Four of 16 dates are retained per patch; training subsets vary by item and epoch, validation subsets are fixed per item, and the subset seed is 8008. Accelerator packs used 4 GPUs with 24 or 32 samples/GPU and, for native Fold~1, 8 GPUs with 16 samples/GPU. & Selected epochs---native: 409, 541, 561, 524, 548; SCG: 585, 586, 553, 597, 539. The four-date input is not one globally fixed quartet. \\
DA released checkpoint & Released \texttt{MykolaL/DelineateAnything} checkpoint; no PASTIS training. Each of the 16 dates is converted to uint8 RGB after a per-image 1st--99th percentile stretch. & Fixed released checkpoint; per-date instances are aggregated only after inference. \\
DA PASTIS fine-tunes & YOLO11x-seg (scale x; 62.05~M parameters), initialized from the released DA checkpoint; separate native-PASTIS and on-the-fly SCG five-fold runs at $128\times128$, each for 50 epochs with one randomly selected date per patch and draw. AdamW, $\mathrm{lr}_0=10^{-4}$, cosine decay to $\mathrm{lrf}=0.01$, 100 warm-up iterations; global batch 128 across eight GPUs, AMP, seed 1729. & Validation fitness is box mAP50--95 plus mask mAP50--95. Selected epoch (fitness), Fold~1--5---native: 22 (0.25327), 27 (0.21392), 23 (0.28057), 19 (0.22672), 37 (0.21159); SCG: 20 (0.41584), 21 (0.39007), 28 (0.29809), 9 (0.31060), 22 (0.21497). Each checkpoint evaluates only its held-out fold; no model ensemble. \\
Loss control & Native-PASTIS Fold~1, $T=16$, 500-epoch cap; the loss is FTNMT, BCE+Dice (1:1), or BCE. The selected FTNMT and BCE checkpoints used 8 GPUs with 3 samples/GPU; BCE+Dice used 4 GPUs with 3 samples/GPU. Continuation phases used different effective batches. & Validation mIoU, then validation loss: FTNMT epoch 459 (0.771658), BCE+Dice epoch 371 (0.771821), and BCE epoch 347 (0.770965). \\
\bottomrule
\end{tabularx}
\endgroup
\end{table}

\FloatBarrier

The perturbation schedule in Table~\ref{SM-tab:app-cloud-condition-matrix} spans the temporal and spatial fractions used for both pooled and regional analyses. The native PASTIS row is the unmodified reference; the mixed-family $T_f=0$ row is a registered no-op used to verify that the perturbation path leaves the input unchanged when no date is selected.

\begin{table}[!htbp]
\centering
\caption{Perturbation matrix used for the controlled cloud-stress experiment. The notation \(T_f\) denotes the fraction of the 16 dates perturbed and is used as a controlled proxy for contamination frequency within the input sequence; \(S_f\) denotes the requested spatial support fraction within each perturbed date. The same perturbation schedule is used for the pooled and regional summaries.}
\label{SM-tab:app-cloud-condition-matrix}
\fontsize{7}{9}\selectfont
\begingroup
\begin{tabularx}{\textwidth}{p{0.18\textwidth}p{0.24\textwidth}p{0.22\textwidth}X}
\toprule
Family & Temporal fractions & Spatial fractions & Conditions represented \\
\midrule
Native PASTIS & \(T_f=0\) & \(S_f=0\) & One unmodified reference condition, which can contain naturally occurring cloud contamination. \\
Mixed cloud-shadow & \(T_f\in\{0,0.25,0.50,0.75\}\) at \(S_f=0.50\); \(T_f=0.50\) at \(S_f\in\{0.25,0.75\}\); \(T_f=1.00\) at \(S_f\in\{0.25,0.50,0.75\}\) & 0.25--0.75 & Nine registered mixed-family settings, including the \(T_f=0\) no-op and the severe all-date/75\% area endpoint. \\
Thick cloud & \(T_f\in\{0.25,0.50,0.75\}\) at \(S_f=0.50\); \(T_f=1.00\) at \(S_f\in\{0.25,0.50,0.75\}\) & 0.25--0.75 & Six thick-cloud settings. \\
Thin cloud & \(T_f\in\{0.25,0.50,0.75\}\) at \(S_f=0.50\); \(T_f=1.00\) at \(S_f\in\{0.25,0.50,0.75\}\) & 0.25--0.75 & Six thin-cloud settings. \\
Cloud shadow & \(T_f\in\{0.25,0.50,0.75\}\) at \(S_f=0.50\); \(T_f=1.00\) at \(S_f\in\{0.25,0.50,0.75\}\) & 0.25--0.75 & Six shadow settings. \\
\bottomrule
\end{tabularx}
\endgroup
\end{table}

\FloatBarrier

\subsection{Object-level polygon matching metrics}
\label{SM-app:object-polygon-metrics}

Object-level metrics are evaluated after converting raster or instance-mask outputs into field polygons. For PTAViT3D, the predicted extent and boundary channels are polygonized with the same paper-aligned post-processing used for the production outputs, implemented in \texttt{preds2shp\_gold\_standard.py}. The implementation thresholds the boundary channel at 0.2, applies Guo--Hall thinning to the boundary mask, removes the thinned boundaries from the predicted extent, thresholds the refined extent at 0.4, removes polygons smaller than 100~m$^2$, removes extremely large artefacts, and simplifies the retained geometries with a 10~m tolerance. The reference polygons used in the ablation are generated with the same extent--boundary polygonizer from the reference extent and boundary labels.

For the DA baseline, predictions are treated differently because DA produces per-date object instances rather than extent--boundary raster channels. Sentinel-2 bands B4, B3, and B2 form the RGB input. Per-date instances are linked across dates separated by at most two positions. A candidate link requires polygon IoU of at least 0.01 and centroid separation of at most 80~m; its score is $1.0\,\mathrm{IoU}+0.35(1-d_{\mathrm{centroid}}/80)$ and must be at least 0.08. One-to-one links are resolved independently for each date pair. The masks in each resulting track are aggregated by mean occupancy, majority vote, union, or intersection; mean occupancy uses threshold 0.5, and outputs smaller than eight pixels are removed. Track-level masks are then polygonized and evaluated with the same polygon matching routine used for PTAViT3D. Overlapping track outputs are resolved deterministically by sorting tracks by temporal support, mean score, and area; later tracks lose pixels already assigned to earlier tracks.

Let $\mathcal{P}=\{P_i\}_{i=1}^{N_P}$ be the predicted polygons and $\mathcal{R}=\{R_j\}_{j=1}^{N_R}$ the reference polygons for a single patch. Candidate prediction--reference pairs are first restricted using a buffered bounding-box test: the predicted polygon bounding box is expanded by 15~m and compared with the reference polygon bounding box. For each candidate pair, the implementation computes polygon IoU, centroid distance, and Hausdorff distance. The assignment cost used is:
\[
C_{ij}
=
(1-\mathrm{IoU}(P_i,R_j))
+
0.3\,d_{\mathrm{centroid}}(P_i,R_j)
+
0.3\,d_{\mathrm{Hausdorff}}(P_i,R_j).
\]

Non-candidate pairs are assigned a large sentinel cost of $10^3$. A one-to-one linear sum assignment, historically formulated as the Hungarian method \citep{Kuhn1955Hungarian}, is then obtained with \texttt{scipy.optimize.linear\_sum\_assignment}. In the ablation runs reported here, the cost threshold is $\infty$; therefore, an assigned pair is retained if it is a finite candidate assignment, i.e. if its cost is below $10^3$.

For each patch, retained one-to-one polygon matches are counted as true positives. Predicted polygons that are not matched to any reference polygon are counted as false positives, and reference polygons that are not matched to any prediction are counted as false negatives:
\[
TP = |\mathcal{M}|,\qquad
FP = N_P - TP,\qquad
FN = N_R - TP,
\]
where $\mathcal{M}$ is the set of retained matched polygon pairs. Object precision, recall, and F1 are then computed as
\[
\mathrm{Precision}
=
\frac{TP}{TP+FP+\epsilon},
\qquad
\mathrm{Recall}
=
\frac{TP}{TP+FN+\epsilon},
\]
\[
F_1
=
\frac{2TP}{2TP+FP+FN+\epsilon},
\]
with $\epsilon=10^{-9}$ used only for numerical stability. These are object-count metrics: each field polygon contributes one object, regardless of its area.

The one-to-one assignment is important for interpreting the metric. If one reference field is split into several predicted polygons, at most one predicted fragment can be matched to that reference field and the remaining fragments become false positives. Conversely, if several reference fields are merged into a single predicted polygon, that prediction can match at most one reference field and the remaining reference fields become false negatives. Object-level F1 is therefore sensitive to field-instance splits, merges, omissions, and spurious detections.

This differs from raster-level metrics, where TP, FP, and FN are counted pixel-wise after valid-mask filtering. Raster F1 measures foreground-pixel overlap and can therefore be dominated by large fields. Object-level F1 instead measures whether individual field instances are detected as separate objects, making it more sensitive to instance-level delineation errors.

For condition-level summaries, the reported precision, recall, and F1 are arithmetic means of the patch-level values. The table values are therefore patch-averaged object metrics unless explicitly stated otherwise.

Polygon Line-to-Segment (PoLiS) distance \citep{6849454} and Hausdorff distance are computed only for the matched prediction--reference polygon pairs. Unmatched false-positive and false-negative objects do not directly contribute to these shape distances, but they do reduce object precision, recall, and F1. For this reason, low PoLiS should be interpreted together with object F1 and recall, especially for DA aggregation variants where conservative masks can align well for matched objects while still missing many field instances.

\subsection{Controlled cloud and shadow stress}
\label{SM-app:controlled-cloud-ablation}

Throughout this section, contamination-induced foreground-IoU degradation is computed separately for each trained model as \(\Delta_{\mathrm{cloud}}\mathrm{IoU}=\mathrm{IoU}_{\mathrm{native}}-\mathrm{IoU}_{\mathrm{perturbed}}\). Positive values indicate a decrease relative to that model's native-PASTIS score, and smaller values indicate greater robustness.

Table~\ref{SM-tab:app-ptavit3d-raster} and Fig.~\ref{SM-fig:app-ptavit3d-raster} summarize raster performance for native PASTIS, one moderate mixed condition, and the severe all-date/75\% area endpoints. The cloud-augmented PTAViT3D retains essentially the same native PASTIS foreground IoU as the native-trained model (0.737 versus 0.739). At the severe mixed cloud--shadow endpoint, foreground IoU falls from 0.739 to 0.707 for the native-trained model (\(\Delta_{\mathrm{cloud}}\mathrm{IoU}=0.032\)), but only from 0.737 to 0.722 for the cloud-augmented model (\(\Delta_{\mathrm{cloud}}\mathrm{IoU}=0.015\)). Under severe thin cloud, the corresponding decreases are 0.074 and 0.031. This experiment isolates perturbation effects more directly than the operational ODC experiment because the date registry and patch set are fixed.

\begin{table}[!htbp]
\centering
\caption{PTAViT3D raster headline results under controlled cloud and shadow stress. For each model, \(\Delta_{\mathrm{cloud}}\mathrm{IoU}=\mathrm{IoU}_{\mathrm{native}}-\mathrm{IoU}_{\mathrm{perturbed}}\); positive values indicate degradation and smaller values indicate greater robustness. The best foreground IoU, foreground F1, and Matthews correlation coefficient (MCC) within each condition are shown in bold.}
\label{SM-tab:app-ptavit3d-raster}
\fontsize{7}{9}\selectfont
\begingroup
\begin{tabular}{llrrrr}
\toprule
Model & Condition & Foreground IoU & \(\Delta_{\mathrm{cloud}}\mathrm{IoU}\) & Foreground F1 & MCC \\
\midrule
PTAViT3D native & native PASTIS & \textbf{0.739} & 0.000 & \textbf{0.833} & \textbf{0.640} \\
PTAViT3D cloud-augmented & native PASTIS & 0.737 & 0.000 & 0.831 & 0.638 \\
\midrule
PTAViT3D native & mixed, 75\% dates / 50\% area & 0.722 & 0.018 & 0.820 & 0.608 \\
PTAViT3D cloud-augmented & mixed, 75\% dates / 50\% area & \textbf{0.728} & \textbf{0.008} & \textbf{0.825} & \textbf{0.623} \\
\midrule
PTAViT3D native & mixed, all dates / 75\% area & 0.707 & 0.032 & 0.808 & 0.578 \\
PTAViT3D cloud-augmented & mixed, all dates / 75\% area & \textbf{0.722} & \textbf{0.015} & \textbf{0.819} & \textbf{0.610} \\
\midrule
PTAViT3D native & thick cloud, all dates / 75\% area & 0.708 & 0.032 & 0.809 & 0.575 \\
PTAViT3D cloud-augmented & thick cloud, all dates / 75\% area & \textbf{0.724} & \textbf{0.013} & \textbf{0.821} & \textbf{0.611} \\
\midrule
PTAViT3D native & thin cloud, all dates / 75\% area & 0.665 & 0.074 & 0.775 & 0.478 \\
PTAViT3D cloud-augmented & thin cloud, all dates / 75\% area & \textbf{0.706} & \textbf{0.031} & \textbf{0.807} & \textbf{0.575} \\
\midrule
PTAViT3D native & shadow, all dates / 75\% area & 0.729 & 0.010 & 0.825 & 0.630 \\
PTAViT3D cloud-augmented & shadow, all dates / 75\% area & \textbf{0.735} & \textbf{0.002} & \textbf{0.829} & \textbf{0.635} \\
\bottomrule
\end{tabular}
\endgroup
\end{table}

\FloatBarrier

\begin{figure}[!htbp]
\centering
\includegraphics[width=0.82\textwidth,height=0.72\textheight,keepaspectratio]{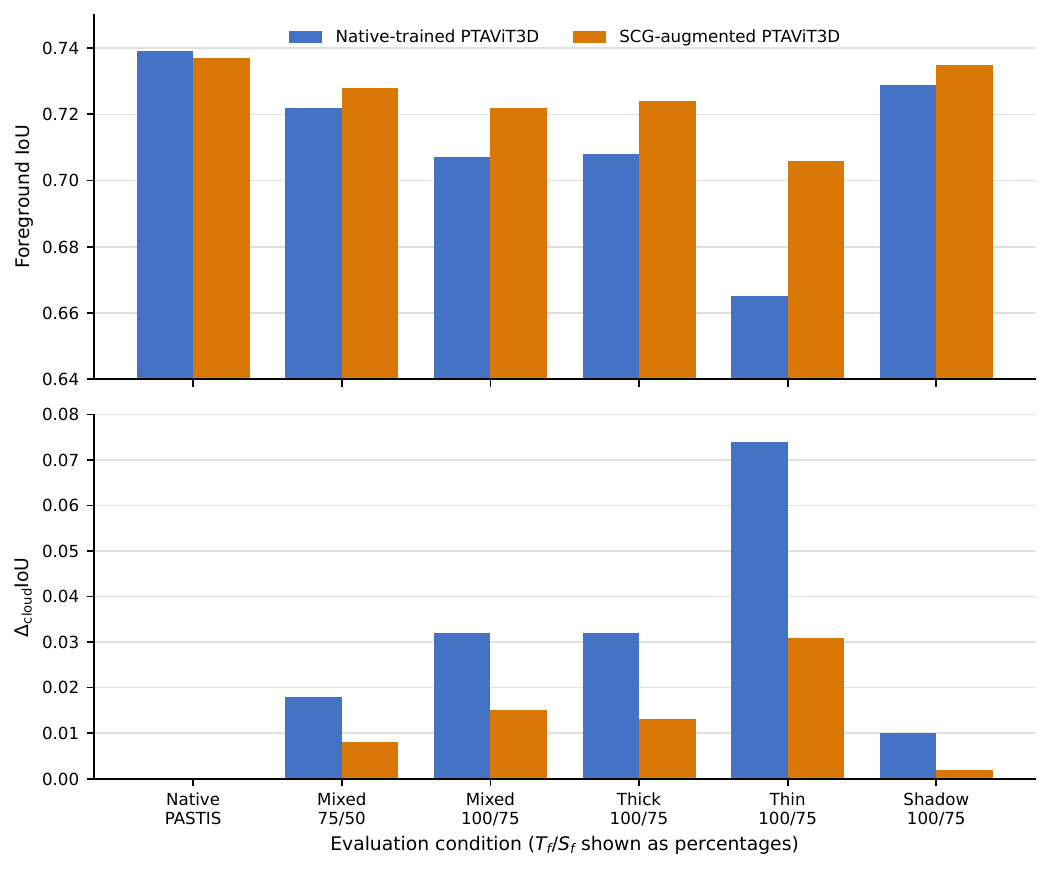}
\caption{Controlled cloud and shadow stress for PTAViT3D at raster level. The cloud-augmented model preserves native-PASTIS performance while reducing severe \(\Delta_{\mathrm{cloud}}\mathrm{IoU}\), with the largest absolute reduction under thin-cloud contamination.}
\label{SM-fig:app-ptavit3d-raster}
\end{figure}

Figure~\ref{SM-fig:app-full-degradation-matrix} and Table~\ref{SM-tab:app-full-degradation-values} expand the headline comparison to all 28 registered conditions. Degradation generally increases as more dates or a larger within-date area is perturbed. SCG-augmented training reduces the loss in nearly every non-zero condition, with the largest separation under thin cloud; the mixed-family no-op reproduces the native-PASTIS values.

\begin{figure}[!htbp]
\centering
\resizebox{\textwidth}{!}{
\begin{tabular}{cc}
\begin{tikzpicture}
\begin{axis}[
  xbar, xmin=0, xmax=0.08, width=0.48\textwidth, height=0.34\textwidth,
  xlabel={Foreground $\Delta_{\mathrm{cloud}}\mathrm{IoU}$}, ylabel={Mixed cloud--shadow},
  symbolic y coords={m02550,m05025,m05050,m05075,m07550,m10025,m10050,m10075},
  ytick=data,
  yticklabels={25/50,50/25,50/50,50/75,75/50,100/25,100/50,100/75},
  y dir=reverse, xmajorgrids, bar width=3pt,
  legend columns=2,
  legend style={font=\footnotesize, at={(0.5,1.04)}, anchor=south, draw=none},
  tick label style={font=\scriptsize}, label style={font=\small}
]
\addplot+[fill=blue!60, draw=blue!60] coordinates {(0.003,m02550) (0.007,m05025) (0.010,m05050) (0.011,m05075) (0.018,m07550) (0.018,m10025) (0.028,m10050) (0.032,m10075)};
\addplot+[fill=orange!80!black, draw=orange!80!black] coordinates {(0.002,m02550) (0.003,m05025) (0.005,m05050) (0.005,m05075) (0.008,m07550) (0.009,m10025) (0.014,m10050) (0.015,m10075)};
\legend{Native, SCG-trained}
\end{axis}
\end{tikzpicture}
&
\begin{tikzpicture}
\begin{axis}[
  xbar, xmin=0, xmax=0.08, width=0.48\textwidth, height=0.34\textwidth,
  xlabel={Foreground $\Delta_{\mathrm{cloud}}\mathrm{IoU}$}, ylabel={Thick cloud},
  symbolic y coords={t02550,t05050,t07550,t10025,t10050,t10075},
  ytick=data,
  yticklabels={25/50,50/50,75/50,100/25,100/50,100/75},
  y dir=reverse, xmajorgrids, bar width=3pt,
  tick label style={font=\scriptsize}, label style={font=\small}
]
\addplot+[fill=blue!60, draw=blue!60] coordinates {(0.003,t02550) (0.009,t05050) (0.020,t07550) (0.021,t10025) (0.029,t10050) (0.032,t10075)};
\addplot+[fill=orange!80!black, draw=orange!80!black] coordinates {(0.002,t02550) (0.004,t05050) (0.009,t07550) (0.010,t10025) (0.013,t10050) (0.013,t10075)};
\end{axis}
\end{tikzpicture}
\\
\begin{tikzpicture}
\begin{axis}[
  xbar, xmin=0, xmax=0.08, width=0.48\textwidth, height=0.31\textwidth,
  xlabel={Foreground $\Delta_{\mathrm{cloud}}\mathrm{IoU}$}, ylabel={Thin cloud},
  symbolic y coords={n02550,n05050,n07550,n10025,n10050,n10075},
  ytick=data,
  yticklabels={25/50,50/50,75/50,100/25,100/50,100/75},
  y dir=reverse, xmajorgrids, bar width=3pt,
  tick label style={font=\scriptsize}, label style={font=\small}
]
\addplot+[fill=blue!60, draw=blue!60] coordinates {(0.005,n02550) (0.016,n05050) (0.032,n07550) (0.024,n10025) (0.055,n10050) (0.074,n10075)};
\addplot+[fill=orange!80!black, draw=orange!80!black] coordinates {(0.002,n02550) (0.007,n05050) (0.013,n07550) (0.010,n10025) (0.021,n10050) (0.031,n10075)};
\end{axis}
\end{tikzpicture}
&
\begin{tikzpicture}
\begin{axis}[
  xbar, xmin=0, xmax=0.08, width=0.48\textwidth, height=0.31\textwidth,
  xlabel={Foreground $\Delta_{\mathrm{cloud}}\mathrm{IoU}$}, ylabel={Cloud shadow},
  symbolic y coords={s02550,s05050,s07550,s10025,s10050,s10075},
  ytick=data,
  yticklabels={25/50,50/50,75/50,100/25,100/50,100/75},
  y dir=reverse, xmajorgrids, bar width=3pt,
  tick label style={font=\scriptsize}, label style={font=\small}
]
\addplot+[fill=blue!60, draw=blue!60] coordinates {(0.002,s02550) (0.004,s05050) (0.006,s07550) (0.007,s10025) (0.011,s10050) (0.010,s10075)};
\addplot+[fill=orange!80!black, draw=orange!80!black] coordinates {(0.000,s02550) (-0.000,s05050) (0.001,s07550) (0.000,s10025) (0.002,s10050) (0.002,s10075)};
\end{axis}
\end{tikzpicture}
\end{tabular}
}
\caption{Full controlled perturbation matrix shown as paired \(\Delta_{\mathrm{cloud}}\mathrm{IoU}\) bars rather than a difference heatmap. Each y tick gives \(T_f/S_f\), the percentage of dates perturbed and requested perturbed area. Shorter bars are better; the SCG-trained PTAViT3D loses less IoU across nearly all non-zero perturbations, especially under severe thin cloud.}
\label{SM-fig:app-full-degradation-matrix}
\end{figure}

\begin{table}[!htbp]
\centering
\caption{Full foreground-IoU degradation table corresponding to Fig.~\ref{SM-fig:app-full-degradation-matrix}. For each model, \(\Delta_{\mathrm{cloud}}\mathrm{IoU}=\mathrm{IoU}_{\mathrm{native}}-\mathrm{IoU}_{\mathrm{perturbed}}\), so smaller values are better; a negative value indicates a negligible gain relative to that model's native-PASTIS condition.}
\label{SM-tab:app-full-degradation-values}
\fontsize{6.2}{7.4}\selectfont
\begingroup
\resizebox{\textwidth}{!}{
\begin{tabular}{llrrrrrr}
\toprule
Condition & Family & \(T_f\) & \(S_f\) & Native IoU & SCG IoU & Native \(\Delta_{\mathrm{cloud}}\mathrm{IoU}\) & SCG \(\Delta_{\mathrm{cloud}}\mathrm{IoU}\) \\
\midrule
native PASTIS & none & 0.00 & 0.00 & 0.739 & 0.737 & \textbf{0.000} & \textbf{0.000} \\
mixed\_t000\_s050 & mixed & 0.00 & 0.50 & 0.739 & 0.737 & \textbf{0.000} & \textbf{0.000} \\
mixed\_t025\_s050 & mixed & 0.25 & 0.50 & 0.736 & 0.735 & 0.003 & \textbf{0.002} \\
mixed\_t050\_s025 & mixed & 0.50 & 0.25 & 0.732 & 0.734 & 0.007 & \textbf{0.003} \\
mixed\_t050\_s050 & mixed & 0.50 & 0.50 & 0.730 & 0.732 & 0.010 & \textbf{0.005} \\
mixed\_t050\_s075 & mixed & 0.50 & 0.75 & 0.729 & 0.732 & 0.011 & \textbf{0.005} \\
mixed\_t075\_s050 & mixed & 0.75 & 0.50 & 0.722 & 0.728 & 0.018 & \textbf{0.008} \\
mixed\_t100\_s025 & mixed & 1.00 & 0.25 & 0.721 & 0.728 & 0.018 & \textbf{0.009} \\
mixed\_t100\_s050 & mixed & 1.00 & 0.50 & 0.711 & 0.723 & 0.028 & \textbf{0.014} \\
mixed\_t100\_s075 & mixed & 1.00 & 0.75 & 0.707 & 0.722 & 0.032 & \textbf{0.015} \\
shadow\_t025\_s050 & shadow & 0.25 & 0.50 & 0.738 & 0.737 & 0.002 & \textbf{0.000} \\
shadow\_t050\_s050 & shadow & 0.50 & 0.50 & 0.736 & 0.737 & 0.004 & \textbf{0.000} \\
shadow\_t075\_s050 & shadow & 0.75 & 0.50 & 0.733 & 0.735 & 0.006 & \textbf{0.001} \\
shadow\_t100\_s025 & shadow & 1.00 & 0.25 & 0.732 & 0.736 & 0.007 & \textbf{0.000} \\
shadow\_t100\_s050 & shadow & 1.00 & 0.50 & 0.729 & 0.735 & 0.011 & \textbf{0.002} \\
shadow\_t100\_s075 & shadow & 1.00 & 0.75 & 0.729 & 0.735 & 0.010 & \textbf{0.002} \\
thick\_t025\_s050 & thick & 0.25 & 0.50 & 0.737 & 0.735 & 0.003 & \textbf{0.002} \\
thick\_t050\_s050 & thick & 0.50 & 0.50 & 0.730 & 0.733 & 0.009 & \textbf{0.004} \\
thick\_t075\_s050 & thick & 0.75 & 0.50 & 0.719 & 0.728 & 0.020 & \textbf{0.009} \\
thick\_t100\_s025 & thick & 1.00 & 0.25 & 0.719 & 0.726 & 0.021 & \textbf{0.010} \\
thick\_t100\_s050 & thick & 1.00 & 0.50 & 0.710 & 0.724 & 0.029 & \textbf{0.013} \\
thick\_t100\_s075 & thick & 1.00 & 0.75 & 0.708 & 0.724 & 0.032 & \textbf{0.013} \\
thin\_t025\_s050 & thin & 0.25 & 0.50 & 0.735 & 0.735 & 0.005 & \textbf{0.002} \\
thin\_t050\_s050 & thin & 0.50 & 0.50 & 0.724 & 0.730 & 0.016 & \textbf{0.007} \\
thin\_t075\_s050 & thin & 0.75 & 0.50 & 0.707 & 0.724 & 0.032 & \textbf{0.013} \\
thin\_t100\_s025 & thin & 1.00 & 0.25 & 0.715 & 0.727 & 0.024 & \textbf{0.010} \\
thin\_t100\_s050 & thin & 1.00 & 0.50 & 0.684 & 0.715 & 0.055 & \textbf{0.021} \\
thin\_t100\_s075 & thin & 1.00 & 0.75 & 0.665 & 0.706 & 0.074 & \textbf{0.031} \\
\bottomrule
\end{tabular}
}
\endgroup
\end{table}

\begingroup
Object metrics show that the raster gains do not come at the expense of vector geometry. Table~\ref{SM-tab:app-ptavit3d-object} and Fig.~\ref{SM-fig:app-ptavit3d-object} report object F1 and matched-pair PoLiS from the native PTAViT3D polygonization path. Because PoLiS includes only one-to-one matches, it should be read with object F1, precision, and recall. Under severe thin cloud, cloud augmentation raises object F1 from 0.680 to 0.694 and lowers PoLiS from 47.65~m to 45.69~m.
\par\endgroup

\begin{table}[!htbp]
\centering
\caption{PTAViT3D object and polygon headline results. PoLiS and Hausdorff distance are computed on matched polygon pairs; unmatched objects enter precision, recall, and F1. The best value within each condition is shown in bold, with lower values preferred for PoLiS and Hausdorff distance.}
\label{SM-tab:app-ptavit3d-object}
\fontsize{7}{9}\selectfont
\begingroup
\begin{tabular}{llrrrrr}
\toprule
Model & Condition & Object F1 & Precision & Recall & PoLiS (m) & Hausdorff (m) \\
\midrule
PTAViT3D native & native PASTIS & \textbf{0.710} & 0.665 & \textbf{0.809} & 43.65 & 136.6 \\
PTAViT3D cloud-augmented & native PASTIS & \textbf{0.710} & \textbf{0.667} & 0.807 & \textbf{43.54} & \textbf{136.4} \\
\midrule
PTAViT3D native & mixed severe & 0.683 & 0.618 & \textbf{0.826} & 43.87 & 137.4 \\
PTAViT3D cloud-augmented & mixed severe & \textbf{0.694} & \textbf{0.635} & 0.819 & \textbf{43.40} & \textbf{135.8} \\
\midrule
PTAViT3D native & thick severe & 0.687 & 0.624 & \textbf{0.827} & 44.46 & 138.8 \\
PTAViT3D cloud-augmented & thick severe & \textbf{0.699} & \textbf{0.642} & 0.819 & \textbf{43.70} & \textbf{136.5} \\
\midrule
PTAViT3D native & thin severe & 0.680 & 0.624 & \textbf{0.817} & 47.65 & 148.1 \\
PTAViT3D cloud-augmented & thin severe & \textbf{0.694} & \textbf{0.647} & 0.807 & \textbf{45.69} & \textbf{142.3} \\
\midrule
PTAViT3D native & shadow severe & 0.700 & 0.649 & 0.810 & 43.16 & 135.1 \\
PTAViT3D cloud-augmented & shadow severe & \textbf{0.703} & \textbf{0.650} & \textbf{0.817} & \textbf{42.92} & \textbf{134.3} \\
\bottomrule
\end{tabular}
\endgroup
\end{table}

\FloatBarrier

\begin{figure}[!htbp]
\centering
\resizebox{\textwidth}{!}{
\begin{tabular}{cc}
\begin{tikzpicture}
\begin{axis}[
  xbar, xmin=0.66, xmax=0.72, width=0.48\textwidth, height=0.31\textwidth,
  xlabel={Object F1}, ylabel={Condition},
  symbolic y coords={native,mixed,thick,thin,shadow},
  ytick=data,
  yticklabels={native PASTIS,mixed,thick,thin,shadow},
  y dir=reverse, xmajorgrids, bar width=4pt,
  legend columns=2,
  legend style={font=\footnotesize, at={(0.5,1.04)}, anchor=south, draw=none},
  tick label style={font=\scriptsize}, label style={font=\small}
]
\addplot+[fill=blue!60, draw=blue!60] coordinates {(0.710,native) (0.683,mixed) (0.687,thick) (0.680,thin) (0.700,shadow)};
\addplot+[fill=orange!80!black, draw=orange!80!black] coordinates {(0.710,native) (0.694,mixed) (0.699,thick) (0.694,thin) (0.703,shadow)};
\legend{Native, SCG-trained}
\end{axis}
\end{tikzpicture}
&
\begin{tikzpicture}
\begin{axis}[
  xbar, xmin=42, xmax=49, width=0.48\textwidth, height=0.31\textwidth,
  xlabel={PoLiS (m)}, ylabel={Condition},
  symbolic y coords={native,mixed,thick,thin,shadow},
  ytick=data,
  yticklabels={native PASTIS,mixed,thick,thin,shadow},
  y dir=reverse, xmajorgrids, bar width=4pt,
  tick label style={font=\scriptsize}, label style={font=\small}
]
\addplot+[fill=blue!60, draw=blue!60] coordinates {(43.65,native) (43.87,mixed) (44.46,thick) (47.65,thin) (43.16,shadow)};
\addplot+[fill=orange!80!black, draw=orange!80!black] coordinates {(43.54,native) (43.40,mixed) (43.70,thick) (45.69,thin) (42.92,shadow)};
\end{axis}
\end{tikzpicture}
\end{tabular}
}
\caption{Object-level controlled cloud-stress results. Higher object F1 is better; lower matched-pair PoLiS indicates better geometric agreement for matched polygons.}
\label{SM-fig:app-ptavit3d-object}
\end{figure}

\FloatBarrier

\subsection{Regional sensitivity to simulated cloud and shadow frequency}
\label{SM-app:regional-cloud-stress}

The synthetic cloud experiments above aggregate performance over the full PASTIS test set. This provides a useful global summary, but it can hide regional differences in model behaviour. We therefore grouped the same matched predictions by native Sentinel-2 source tile. Increasing \(T_f\) simulates a greater within-sequence frequency of contamination by affecting a larger fraction of the 16 acquisitions. Because every source tile is evaluated with the same \(T_f\), \(S_f\), and perturbation-type schedule, differences in \(\Delta_{\mathrm{cloud}}\mathrm{IoU}\) quantify region-dependent sensitivity at an equivalent simulated contamination burden. The analysis is a controlled regional sensitivity experiment rather than an estimate of observed regional cloud climatology.

Table~\ref{SM-tab:app-regional-cloud-stress} reports regional foreground-IoU results for the native and cloud-augmented PTAViT3D models. Native-PASTIS scores vary substantially, most notably on T31TFJ. This matters because absolute performance under added cloud stress partly reflects each region's native-input difficulty. The cloud-augmented model gives very similar native-PASTIS performance overall, with small positive differences in T30UXV, T31TFM, and T32ULU and a small negative difference in T31TFJ.

Under severe synthetic contamination, the degradation pattern is consistent across regions. Thin cloud is the most damaging perturbation in all four regions, followed by thick or mixed cloud depending on the source tile, while cloud shadow produces smaller decreases. Cloud-augmented training reduces \(\Delta_{\mathrm{cloud}}\mathrm{IoU}\) in nearly all regional stress cases. Averaged over the four severe endpoints used here (mixed 100/75, thick 100/75, thin 100/75, and shadow 100/75), mean \(\Delta_{\mathrm{cloud}}\mathrm{IoU}\) is 0.036 for the native-trained model and 0.015 for the cloud-augmented model, a reduction of 0.021. The largest reductions occur for thin-cloud contamination, especially in T31TFM and T32ULU, where cloud augmentation reduces \(\Delta_{\mathrm{cloud}}\mathrm{IoU}\) by approximately 0.052 and 0.055, respectively.

At matched contamination frequencies and severities, both native-input difficulty and the magnitude of degradation vary by source tile, but the benefit of cloud-augmented training is not confined to a pooled score. It reduces degradation across multiple source regions, with the largest benefits under thin cloud, where spectral contamination is less visually obvious than thick cloud but still changes the input distribution.

\begin{table}[!htbp]
\centering
\caption{Regional sensitivity of PTAViT3D under matched synthetic cloud stress. Every source tile is evaluated under the same schedule; for example, 100/75 denotes perturbation of all 16 dates with 75\% spatial support per affected date. For each model and source tile, \(\Delta_{\mathrm{cloud}}\mathrm{IoU}=\mathrm{IoU}_{\mathrm{native}}-\mathrm{IoU}_{\mathrm{perturbed}}\); positive values indicate degradation. Reduction is the native-trained value minus the cloud-augmented value. These matched comparisons quantify regional accuracy sensitivity, not observed cloud climatology.}
\label{SM-tab:app-regional-cloud-stress}
\fontsize{6.5}{8}\selectfont
\begingroup
\resizebox{\textwidth}{!}{
\begin{tabular}{llrrrrrr}
\toprule
Region & Condition & Patches & Native IoU & Cloud-aug. IoU & Native \(\Delta_{\mathrm{cloud}}\mathrm{IoU}\) & Cloud-aug. \(\Delta_{\mathrm{cloud}}\mathrm{IoU}\) & Reduction \\
\midrule
T30UXV & Native PASTIS & 531 & 0.807 & 0.809 & 0.000 & 0.000 & 0.000 \\
T30UXV & Mixed 75/50 & 531 & 0.797 & 0.801 & 0.010 & 0.008 & 0.002 \\
T30UXV & Mixed 100/75 & 531 & 0.789 & 0.793 & 0.018 & 0.016 & 0.002 \\
T30UXV & Shadow 100/75 & 531 & 0.793 & 0.800 & 0.015 & 0.009 & 0.005 \\
T30UXV & Thick 100/75 & 531 & 0.791 & 0.796 & 0.016 & 0.013 & 0.003 \\
T30UXV & Thin 100/75 & 531 & 0.772 & 0.788 & 0.035 & 0.022 & 0.013 \\
\midrule
T31TFJ & Native PASTIS & 623 & 0.561 & 0.548 & 0.000 & 0.000 & 0.000 \\
T31TFJ & Mixed 75/50 & 623 & 0.536 & 0.540 & 0.025 & 0.008 & 0.017 \\
T31TFJ & Mixed 100/75 & 623 & 0.516 & 0.531 & 0.046 & 0.017 & 0.029 \\
T31TFJ & Shadow 100/75 & 623 & 0.544 & 0.555 & 0.018 & -0.007 & 0.025 \\
T31TFJ & Thick 100/75 & 623 & 0.519 & 0.535 & 0.042 & 0.013 & 0.029 \\
T31TFJ & Thin 100/75 & 623 & 0.476 & 0.512 & 0.085 & 0.036 & 0.049 \\
\midrule
T31TFM & Native PASTIS & 723 & 0.800 & 0.800 & 0.000 & 0.000 & 0.000 \\
T31TFM & Mixed 75/50 & 723 & 0.782 & 0.791 & 0.017 & 0.009 & 0.008 \\
T31TFM & Mixed 100/75 & 723 & 0.770 & 0.787 & 0.030 & 0.014 & 0.016 \\
T31TFM & Shadow 100/75 & 723 & 0.794 & 0.797 & 0.006 & 0.004 & 0.002 \\
T31TFM & Thick 100/75 & 723 & 0.768 & 0.788 & 0.031 & 0.013 & 0.019 \\
T31TFM & Thin 100/75 & 723 & 0.714 & 0.766 & 0.086 & 0.034 & 0.052 \\
\midrule
T32ULU & Native PASTIS & 556 & 0.796 & 0.796 & 0.000 & 0.000 & 0.000 \\
T32ULU & Mixed 75/50 & 556 & 0.779 & 0.788 & 0.016 & 0.008 & 0.009 \\
T32ULU & Mixed 100/75 & 556 & 0.763 & 0.781 & 0.033 & 0.015 & 0.018 \\
T32ULU & Shadow 100/75 & 556 & 0.791 & 0.793 & 0.005 & 0.004 & 0.001 \\
T32ULU & Thick 100/75 & 556 & 0.761 & 0.783 & 0.035 & 0.013 & 0.022 \\
T32ULU & Thin 100/75 & 556 & 0.711 & 0.767 & 0.084 & 0.029 & 0.055 \\
\bottomrule
\end{tabular}
}
\endgroup
\end{table}

\FloatBarrier

\subsection{2D temporal encodings and per-date aggregation}
\label{SM-app:temporal-2d-baselines}

\begingroup
The 2D controls cover a 16-date channel-stacked PTAViT2D, per-date PTAViT2D with temporal averaging, and a four-date stress case. In Table~\ref{SM-tab:app-2d3d-severe} and Fig.~\ref{SM-fig:app-2d3d-severe}, the stacked model is a strong baseline, but cloud-augmented PTAViT3D has the highest severe-average object F1 and the lowest matched-object geometry errors. The weaker per-date mean confirms that simple post-hoc averaging is not equivalent to learned spatio-temporal features.
\par\endgroup

\begin{table}[!htbp]
\centering
\caption{Severe-cloud average object metrics for 2D temporal encodings and PTAViT3D. Severe average is the mean over the severe mixed, thick, thin, and shadow endpoints. The best value in each metric column is shown in bold, with lower values preferred for PoLiS and Hausdorff distance.}
\label{SM-tab:app-2d3d-severe}
\fontsize{7}{9}\selectfont
\begingroup
\begin{tabular}{lrrrrr}
\toprule
Model family & Object F1 & Precision & Recall & PoLiS (m) & Hausdorff (m) \\
\midrule
PTAViT3D native & 0.687 & 0.629 & \textbf{0.820} & 44.79 & 139.9 \\
PTAViT3D cloud-augmented & \textbf{0.698} & 0.644 & 0.816 & \textbf{43.93} & \textbf{137.2} \\
PTAViT2D stacked & 0.690 & 0.686 & 0.758 & 48.52 & 150.9 \\
PTAViT2D stacked, cloud-augmented & 0.693 & 0.682 & 0.758 & 46.34 & 144.8 \\
PTAViT2D four-date & 0.665 & 0.655 & 0.755 & 50.05 & 154.7 \\
PTAViT2D four-date, cloud-augmented & 0.663 & 0.635 & 0.765 & 47.17 & 146.1 \\
PTAViT2D per-date mean & 0.494 & 0.717 & 0.486 & 82.91 & 255.0 \\
PTAViT2D per-date mean, cloud-augmented & 0.525 & \textbf{0.719} & 0.515 & 73.27 & 227.0 \\
\bottomrule
\end{tabular}
\endgroup
\end{table}

\begin{figure}[!htbp]
\centering
\resizebox{\textwidth}{!}{
\begin{tabular}{cc}
\begin{tikzpicture}
\begin{axis}[
  xbar, xmin=0.45, xmax=0.72, width=0.49\textwidth, height=0.38\textwidth,
  xlabel={Severe-average object F1}, ylabel={Model family},
  symbolic y coords={d3n,d3s,s2n,s2s,f2n,f2s,p2n,p2s},
  ytick=data,
  yticklabels={3D native,3D SCG,2D stacked,2D stacked SCG,2D 4-date,2D 4-date SCG,2D per-date,2D per-date SCG},
  y dir=reverse, xmajorgrids, bar width=4pt,
  tick label style={font=\scriptsize}, label style={font=\small}
]
\addplot+[fill=blue!60, draw=blue!60] coordinates {(0.687,d3n) (0.698,d3s) (0.690,s2n) (0.693,s2s) (0.665,f2n) (0.663,f2s) (0.494,p2n) (0.525,p2s)};
\end{axis}
\end{tikzpicture}
&
\begin{tikzpicture}
\begin{axis}[
  xbar, xmin=40, xmax=85, width=0.49\textwidth, height=0.38\textwidth,
  xlabel={Severe-average PoLiS (m)}, ylabel={Model family},
  symbolic y coords={d3n,d3s,s2n,s2s,f2n,f2s,p2n,p2s},
  ytick=data,
  yticklabels={3D native,3D SCG,2D stacked,2D stacked SCG,2D 4-date,2D 4-date SCG,2D per-date,2D per-date SCG},
  y dir=reverse, xmajorgrids, bar width=4pt,
  tick label style={font=\scriptsize}, label style={font=\small}
]
\addplot+[fill=orange!80!black, draw=orange!80!black] coordinates {(44.79,d3n) (43.93,d3s) (48.52,s2n) (46.34,s2s) (50.05,f2n) (47.17,f2s) (82.91,p2n) (73.27,p2s)};
\end{axis}
\end{tikzpicture}
\end{tabular}
}
\caption{Severe-stress comparison of 2D temporal encodings and 3D time-series modelling. The stacked 2D model is competitive in object F1, but PTAViT3D cloud augmentation gives the best object F1 and matched-polygon geometry.}
\label{SM-fig:app-2d3d-severe}
\end{figure}

\FloatBarrier

\subsection{Delineate Anything baselines}
\label{SM-app:da-baselines}

\begingroup
We selected Delineate Anything (DA; \citealp{Lavreniuk2025DelineateAnything}) as the modern 2D field-delineation baseline because it is a recent field-specific instance-segmentation model rather than a generic semantic segmentation network. DA adopts YOLOv11 for instance segmentation and was developed with FBIS-22M, which contains 672,909 image patches and 22,926,427 field-instance masks spanning reported ground sampling distances from 0.25 to 10~m. On the authors' FBIS-22M test split, the full DA model was reported at 0.720 mAP@0.5 and 0.477 mAP@0.5:0.95, compared with 0.339/0.197 for SAM and 0.382/0.235 for SAM2; reported per-image latency was 25~ms for DA and 10,370~ms for SAM2. These dataset-specific results motivated its selection as a strong, publicly described single-date field-instance comparator; they are not assumed to transfer unchanged to PASTIS.

DA is evaluated in two complementary roles. The released FBIS-22M checkpoint provides a zero-shot PASTIS baseline: it is applied independently to each of the 16 Sentinel-2 dates and its instances are linked and aggregated across time. We then give DA task-specific adaptation by fine-tuning it on the repurposed PASTIS field annotations. Each training draw contains one patch label and one randomly selected date from the 16-date sequence, avoiding a 16-fold repetition of the same parcel label. Separate five-fold, 50-epoch variants use native PASTIS and SCG-augmented inputs. At evaluation, all 16 dates are processed and aggregated by mean occupancy, majority vote, union, or intersection. This design distinguishes zero-shot transfer from a fairer, task-adapted DA comparison without changing DA's single-date architecture.

Both fine-tuned variants use the released YOLO11x-seg scale x checkpoint (62,051,411 parameters; SHA256 \texttt{e3dc\allowbreak{}da35\allowbreak{}7800\allowbreak{}83ae\allowbreak{}aefe\allowbreak{}9425\allowbreak{}b73b\allowbreak{}15a3\allowbreak{}0561\allowbreak{}cdc1\allowbreak{}2c43\allowbreak{}2770\allowbreak{}41d0\allowbreak{}4f9e\allowbreak{}88ed\allowbreak{}e029}) with Ultralytics 8.4.15 at $128\times128$ pixels. No layer is frozen by the user; Ultralytics automatically freezes \texttt{model.23.dfl.conv.weight}. Optimization uses AdamW with initial learning rate $10^{-4}$, momentum 0.937, configured weight decay 0.01, and cosine decay to a final factor of 0.01. A configured three-epoch warm-up resolves to the implementation minimum of 100 iterations. The global batch of 128 is distributed across eight GPUs (16 images per GPU) without gradient accumulation. Automatic mixed precision is enabled; the seed is 1729 and deterministic execution is disabled. All runs complete 50 epochs, so patience 50 does not trigger early stopping.

The active segmentation-training transforms use HSV settings 0.015/0.7/0.4, rotation up to $180^{\circ}$, translation and scale up to 0.25, and horizontal and vertical flips with probability 0.5. Blur, median blur, grayscale conversion, and CLAHE are each logged with probability 0.01. Mosaic, MixUp, CutMix, copy-paste, shear, perspective, multi-scale training, and dropout are disabled; overlapping masks are retained with mask ratio 4. The remaining stored fields are \texttt{rect=false}, \texttt{close\_mosaic=10}, \texttt{bgr=0}, and \texttt{copy\_paste\_mode=flip}; because copy-paste is disabled, its stored mode is inactive. The stored \texttt{augment=false} flag controls augmented prediction at inference and does not disable the training transforms above. Although \texttt{auto\_augment=randaugment} and \texttt{erasing=0.4} are stored in the generic configuration, the segmentation transform path does not consume these fields.

For the SCG variant, perturbations are generated on the fly after random-date selection and no refresh manifest is used. The completed run metadata do not retain the resolved disturbance family, temporal frequency, spatial coverage, or blend mode; these values are therefore not inferred from launcher fallback settings. This limitation concerns DA training-time augmentation only; the registered evaluation perturbations follow the fixed protocol defined above.

Checkpoint selection maximizes the sum of validation box and mask mAP50--95, using one canonical date per validation patch; the fold-specific epochs and fitness values are listed in Table~\ref{SM-tab:app-training-provenance}. During epoch-level validation, the stored configuration is \texttt{conf=null}, NMS IoU 0.7, \texttt{max\_det=300}, and \texttt{agnostic\_nms=false}; because confidence is null, its effective numeric validation value is not recovered from the run record. At test time, each fold checkpoint processes all 16 dates of its own held-out shard, and predictions from different fold models are not ensembled. Object-aware inference uses confidence 0.005, mask threshold 0.5, retina masks, and minimum-area filters of 8 pixels and 100~m$^2$. The downstream NMS IoU, maximum-detection count, and agnostic/class-aware mode were not retained. Per-condition means pool all 2,433 held-out patches with equal patch weight, while TP, FP, and FN are summed.

Under this protocol, SCG-augmented fine-tuning did not improve DA object F1 on either native PASTIS or the severe synthetic-cloud evaluation. This result is specific to the present training and aggregation settings and does not imply a general incompatibility between instance segmentation and cloud augmentation.

This is a 2D temporal-aggregation control rather than a direct replication of the DA paper or a claim that DA and PTAViT3D solve the same input problem. DA consumes RGB images one date at a time and has no mechanism for joint temporal reasoning, Sentinel-1 fusion, or explicit cloud-contaminated time-series modelling. The comparison asks whether a strong single-date instance model plus post-hoc aggregation is sufficient, or whether joint spatio-temporal modelling behaves differently under the controlled cloud-stress protocol.
\par\endgroup

\begingroup
Figure~\ref{SM-fig:app-da-pretrained} isolates raster aggregation for the released DA checkpoint. Mean occupancy is strongest, reaching 0.7166 foreground IoU on native PASTIS and 0.6565 averaged over the four severe endpoints; intersection is overly conservative. Raster IoU can nevertheless obscure fragmentation and missed objects, so Table~\ref{SM-tab:app-da-finetuned} compares released and fine-tuned DA checkpoints under the same object-aware union aggregation and Hungarian-matched metrics. Native-PASTIS fine-tuning raises DA object F1 from 0.338 to 0.494 on native PASTIS and from 0.378 to 0.489 on the severe average, while reducing PoLiS from 79.31 to 51.69~m and from 66.78 to 54.28~m, respectively. Fine-tuning therefore gives DA a material task-specific gain. The resulting DA scores remain below the corresponding PTAViT3D values, but the comparison remains a single-date RGB aggregation control rather than an architecture-equivalent test.
\par\endgroup

\begin{figure}[!htbp]
\centering
\includegraphics[width=0.82\textwidth,height=0.72\textheight,keepaspectratio]{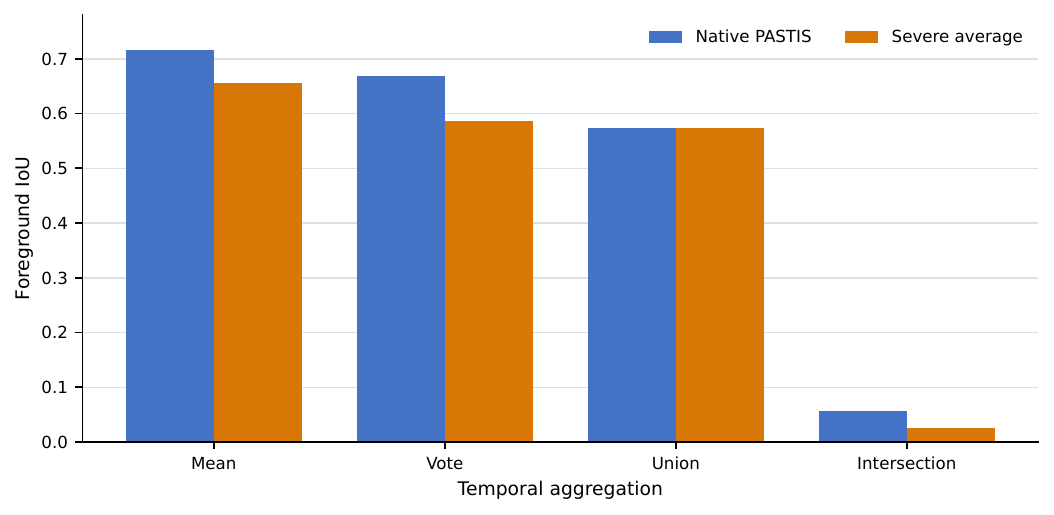}
\caption{Pretrained DA applied independently to each date with RGB input preprocessing and temporal aggregation. Mean occupancy provides the best raster IoU, while intersection is overly conservative.}
\label{SM-fig:app-da-pretrained}
\end{figure}

\begin{table}[!htbp]
\centering
\caption{Released and PASTIS-fine-tuned DA checkpoints compared with PTAViT3D using common object metrics. Every DA row uses object-aware union aggregation over all 16 dates, the DA aggregation with the highest object F1. Severe average is the mean over severe mixed, thick, thin, and shadow endpoints. The best value in each metric column is shown in bold; lower PoLiS is better.}
\label{SM-tab:app-da-finetuned}
\fontsize{7}{9}\selectfont
\begingroup
\begin{tabular}{lrrrr}
\toprule
Model & Native F1 & Severe F1 & Native PoLiS (m) & Severe PoLiS (m) \\
\midrule
DA released FBIS-22M checkpoint & 0.338 & 0.378 & 79.31 & 66.78 \\
DA fine-tuned on native PASTIS & 0.494 & 0.489 & 51.69 & 54.28 \\
DA fine-tuned with SCG augmentation & 0.266 & 0.232 & 53.22 & 53.87 \\
PTAViT3D native & \textbf{0.710} & 0.687 & 43.65 & 44.79 \\
PTAViT3D cloud-augmented & \textbf{0.710} & \textbf{0.698} & \textbf{43.54} & \textbf{43.93} \\
\bottomrule
\end{tabular}
\endgroup
\end{table}

\begin{table}[!htbp]
\centering
\caption{Object-level sensitivity of PASTIS-fine-tuned DA to temporal aggregation. Severe average is the mean over severe mixed, thick, thin, and shadow endpoints. Union gives the highest object F1 for both training regimes, whereas intersection gives the lowest PoLiS among matched objects.}
\label{SM-tab:app-da-aggregation-sensitivity}
\fontsize{7}{9}\selectfont
\begingroup
\begin{tabular}{llrrrr}
\toprule
Training regime & Aggregation & Native F1 & Severe F1 & Native PoLiS (m) & Severe PoLiS (m) \\
\midrule
DA fine-tuned on native PASTIS & mean & 0.343 & 0.352 & 35.43 & 38.17 \\
DA fine-tuned on native PASTIS & vote & 0.332 & 0.337 & 34.62 & 37.14 \\
DA fine-tuned on native PASTIS & union & \textbf{0.494} & \textbf{0.489} & 51.69 & 54.28 \\
DA fine-tuned on native PASTIS & intersection & 0.292 & 0.299 & \textbf{34.04} & \textbf{36.99} \\
\midrule
DA fine-tuned with SCG augmentation & mean & 0.224 & 0.203 & 48.59 & 49.59 \\
DA fine-tuned with SCG augmentation & vote & 0.219 & 0.196 & 48.19 & 48.78 \\
DA fine-tuned with SCG augmentation & union & \textbf{0.266} & \textbf{0.232} & 53.22 & 53.87 \\
DA fine-tuned with SCG augmentation & intersection & 0.204 & 0.187 & \textbf{47.11} & \textbf{47.90} \\
\bottomrule
\end{tabular}
\endgroup
\end{table}

\begingroup
Table~\ref{SM-tab:app-da-aggregation-sensitivity} explains the metric trade-off behind the selected DA aggregation: mean occupancy is stronger in raster IoU, whereas object union retains more temporally detected fields and gives the highest object F1 at the cost of a higher PoLiS among matched polygons. Figure~\ref{SM-fig:app-da-finetuned} visualizes the fine-tuned DA and PTAViT3D rows under native PASTIS and severe stress.
\par\endgroup

\begin{figure}[!htbp]
\centering
\resizebox{\textwidth}{!}{
\begin{tabular}{cc}
\begin{tikzpicture}
\begin{axis}[
  xbar, xmin=0.20, xmax=0.75, width=0.48\textwidth, height=0.28\textwidth,
  xlabel={Object F1}, ylabel={Model},
  symbolic y coords={dac,das,pn,ps},
  ytick=data,
  yticklabels={DA native-trained,DA SCG-trained,PTAViT3D native,PTAViT3D SCG},
  y dir=reverse, xmajorgrids, bar width=4pt,
  legend columns=2,
  legend style={font=\footnotesize, at={(0.5,1.04)}, anchor=south, draw=none},
  tick label style={font=\scriptsize}, label style={font=\small}
]
\addplot+[fill=blue!60, draw=blue!60] coordinates {(0.494,dac) (0.266,das) (0.710,pn) (0.710,ps)};
\addplot+[fill=orange!80!black, draw=orange!80!black] coordinates {(0.489,dac) (0.232,das) (0.687,pn) (0.698,ps)};
\legend{Native PASTIS, Severe average}
\end{axis}
\end{tikzpicture}
&
\begin{tikzpicture}
\begin{axis}[
  xbar, xmin=40, xmax=56, width=0.48\textwidth, height=0.28\textwidth,
  xlabel={PoLiS (m)}, ylabel={Model},
  symbolic y coords={dac,das,pn,ps},
  ytick=data,
  yticklabels={DA native-trained,DA SCG-trained,PTAViT3D native,PTAViT3D SCG},
  y dir=reverse, xmajorgrids, bar width=4pt,
  tick label style={font=\scriptsize}, label style={font=\small}
]
\addplot+[fill=blue!60, draw=blue!60] coordinates {(51.69,dac) (53.22,das) (43.65,pn) (43.54,ps)};
\addplot+[fill=orange!80!black, draw=orange!80!black] coordinates {(54.28,dac) (53.87,das) (44.79,pn) (43.93,ps)};
\end{axis}
\end{tikzpicture}
\end{tabular}
}
\caption{Object-level comparison between PASTIS-fine-tuned per-date DA and PTAViT3D. DA is evaluated over all 16 dates with union aggregation; PTAViT3D consumes the temporal tensor directly.}
\label{SM-fig:app-da-finetuned}
\end{figure}

Figures~\ref{SM-fig:app-da-native-visual} and \ref{SM-fig:app-da-cloud-visual} illustrate the common temporal and polygon-evaluation contract for one Fold~5 patch; these examples were not used for checkpoint or threshold selection. The native sequence in Fig.~\ref{SM-fig:app-da-native-visual} already contains strongly cloud-affected dates, while other dates expose the parcel pattern. PTAViT3D recovers many of the dominant parcel boundaries, whereas the DA aggregation panels show additional subdivisions and omissions that differ among mean, voting, and union. After severe thin-cloud perturbation of all 16 dates (Fig.~\ref{SM-fig:app-da-cloud-visual}), PTAViT3D retains a spatially coherent boundary response and both approaches still recover many major parcels; the DA panels continue to show aggregation-dependent fragmentation. The examples are qualitative illustrations of the behaviours summarized by object F1 and PoLiS, not evidence beyond the full test-set metrics.

\begin{figure}[!htbp]
\centering
\includegraphics[width=0.98\textwidth,height=0.72\textheight,keepaspectratio]{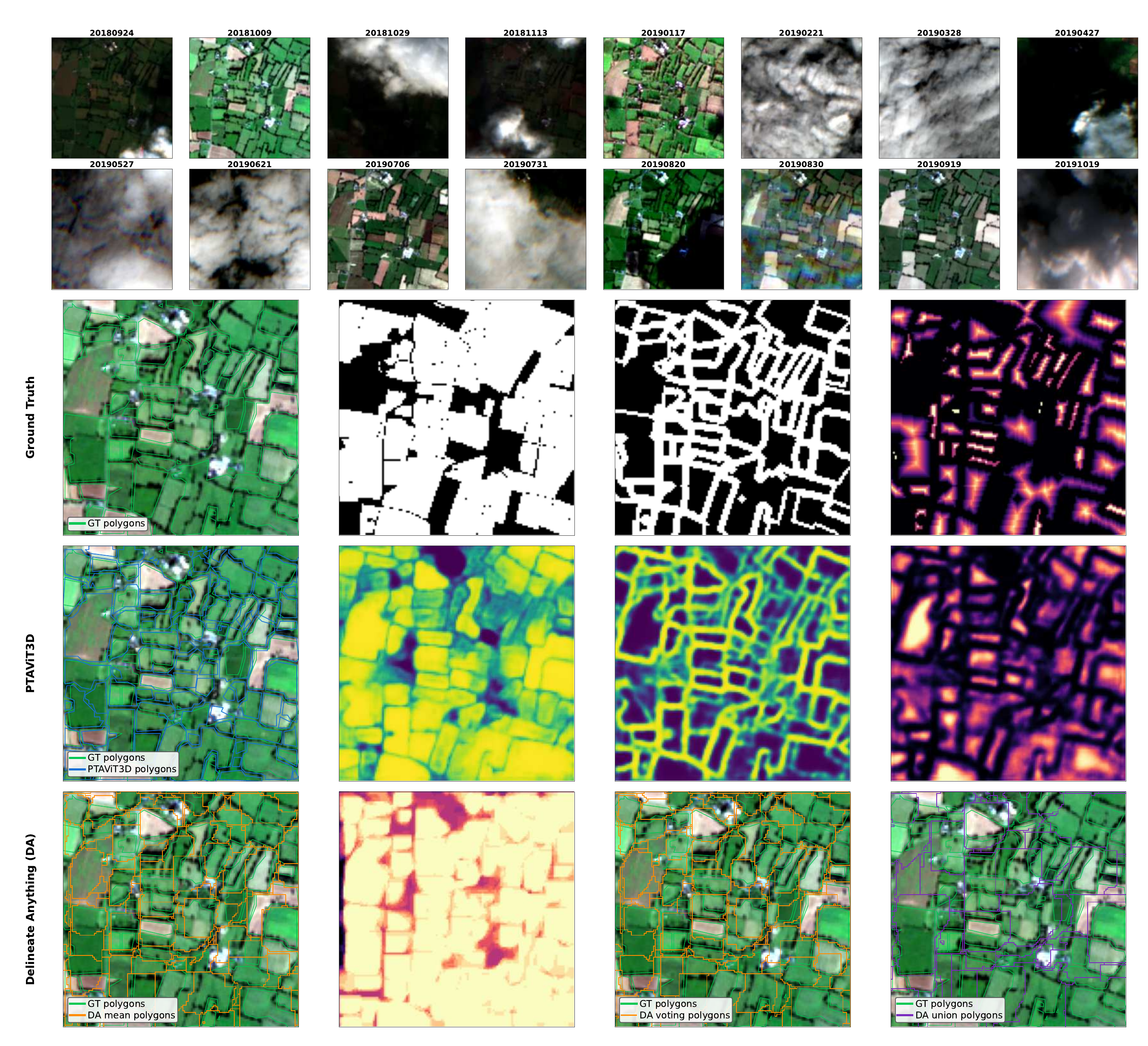}
\caption{Native-PASTIS visual comparison for case01\_native, Fold~5 patch~10234. The first two rows show the 16 Sentinel-2 dates from 20180924 to 20191019; overlay panels use the native image at time index~1 (20181009). Lower rows show reference polygons and labels, PTAViT3D polygon/probability/boundary/distance outputs, and DA mean, voting, and union polygon outputs. Reference polygons use extent threshold 0.4, boundary threshold 0.2, minimum area 100~m$^2$, and simplification tolerance 10~m.}
\label{SM-fig:app-da-native-visual}
\end{figure}

\begin{figure}[!htbp]
\centering
\includegraphics[width=0.98\textwidth,height=0.72\textheight,keepaspectratio]{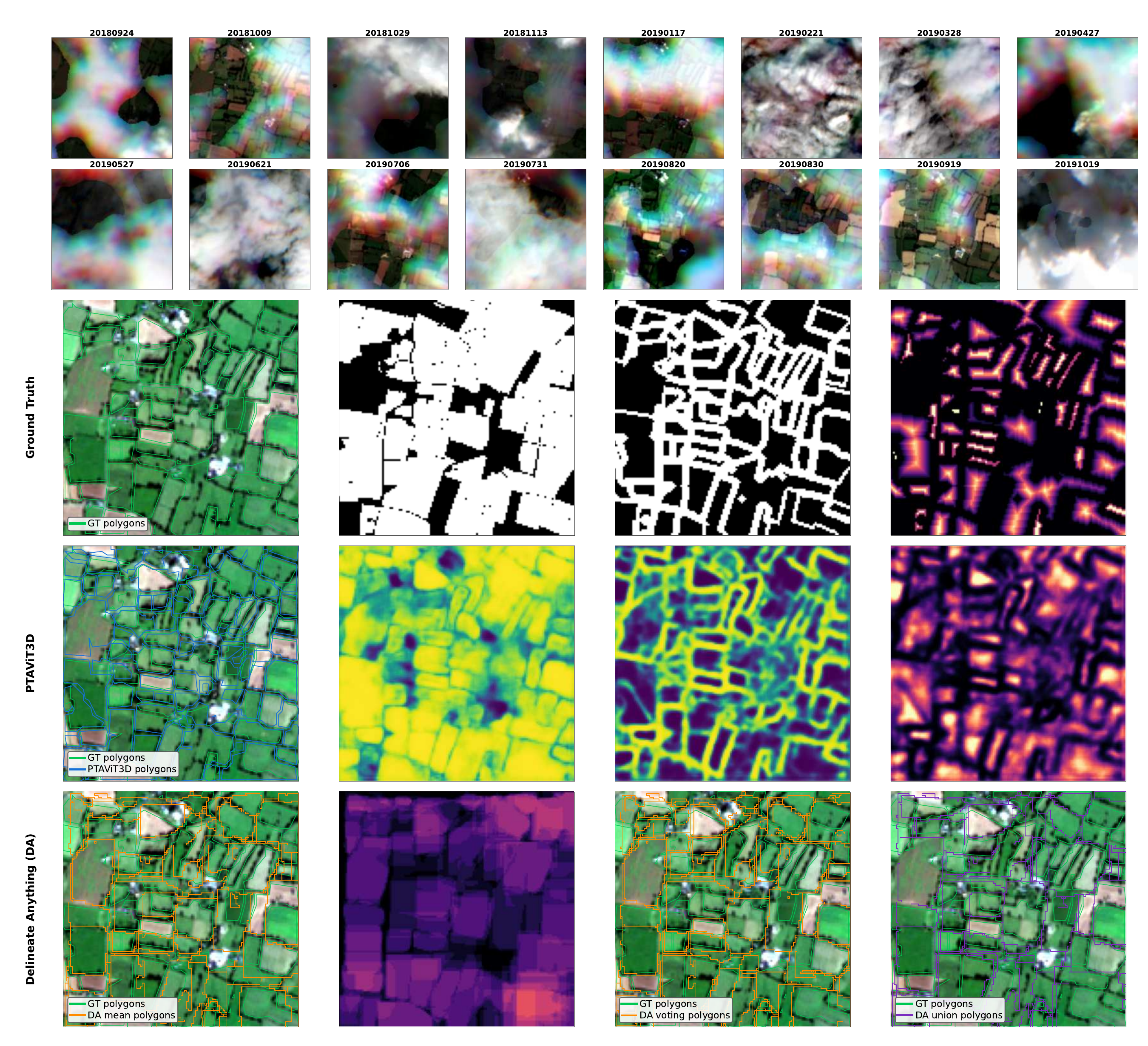}
\caption{Severe thin-cloud visual comparison for case01\_randomcloud, Fold~5 patch~10234, condition thin\_t100\_s075 with SCG seed 10940518. All 16 dates are perturbed at 75\% spatial support; overlay panels use the matched native PASTIS image at time index~1 (20181009) so polygon geometry can be compared directly. Lower rows show reference polygons and labels, PTAViT3D cloud-augmented outputs, and DA mean, voting, and union polygon outputs. Reference polygons use extent threshold 0.4, boundary threshold 0.2, minimum area 100~m$^2$, and simplification tolerance 10~m.}
\label{SM-fig:app-da-cloud-visual}
\end{figure}

\FloatBarrier
\FloatBarrier

\subsection{Comparison of PTA3D with matched-support window attention}
\label{SM-supp:pta-vs-window-attention}

\subsubsection{Purpose of the comparison}

The purpose of this experiment is to compare the resource behaviour of Patch Tanimoto Attention 3D (PTA3D) with a standard windowed self-attention baseline under the closest practical matched conditions. The two attention operators cannot be made identical: PTA3D uses a Tanimoto-style patch similarity with channel partitioning, while window attention uses scaled dot-product softmax attention over local windows. Therefore, we do not claim that the two modules compute the same function. Instead, we match the outer tensor contract and the effective temporal--spatial support, and then compare latency, peak GPU memory, and out-of-memory behaviour.

The stage-level comparison is complemented by a completed, fixed-runtime FoTW experiment in which PTA3D is replaced throughout the otherwise unchanged architecture by full-temporal, per-stage support-matched window attention. This end-to-end control tests whether the stage-level resource advantage persists without sacrificing segmentation accuracy.

Let an intermediate feature tensor be
\[
X \in \mathbb{R}^{B \times C \times T \times H \times W},
\]
where \(B\) is batch size, \(C\) is the number of channels, \(T\) is the number of temporal acquisitions, and \(H,W\) are the spatial dimensions. Both attention modules receive the same tensor shape, use the same batch size, channel count, number of heads, dtype, GPU, warm-up schedule, number of measured iterations, and CUDA-synchronised timing protocol.

\subsubsection{Effective-support matching}

\begingroup
The matching rule can be defined directly from the feature tensor, without reference to either software implementation. At a given network layer, let the tensor contain \(C\) channels, \(T\) acquisition dates, and an \(H\times W\) spatial grid. PTA3D is configured by three \emph{partition counts}, \((s_c,s_h,s_w)\). These counts divide the channel and spatial axes, giving a patch with dimensions
\[
(C',H',W')=
\left(\frac{C}{s_c},\frac{H}{s_h},\frac{W}{s_w}\right).
\]
The non-causal PTA3D configuration used in this study retains all \(T\) dates. Its temporal--spatial comparison box is therefore
\[
(T,H',W')=
\left(T,\frac{H}{s_h},\frac{W}{s_w}\right).
\]

Window attention is specified differently: \((W_t,W_h,W_w)\) gives the literal number of temporal and spatial tokens within each local window. We therefore define the support-matched window by
\[
(W_t,W_h,W_w)=(T,H',W')
=\left(T,\frac{H}{s_h},\frac{W}{s_w}\right).
\]
Consequently, PTA3D and window attention see the same outer tensor and the same \(T\times H'\times W'\) temporal--spatial box in every matched row. Their channel operations remain different: PTA3D compares channel-partitioned patches, whereas window attention applies softmax attention to full-channel token embeddings. Support matching therefore controls the information span, not the mathematical form of the two operators.

For the microbenchmark, \(C=64\), the PTA3D channel patch is fixed at \(C'=8\), and the spatial patch is fixed at \(H'=W'=16\). This gives \((s_c,s_h,s_w)=(C/8,H/16,W/16)\) for PTA3D and \((W_t,W_h,W_w)=(T,16,16)\) for window attention. Table~\ref{SM-tab:supp-attention-support-geometry} displays the resulting temporal--spatial box separately for both operators; the final two columns are intentionally identical row by row.
\par\endgroup

\begin{table}[!htbp]
\centering
\caption{Matched-support geometry for the PTA3D and window-attention microbenchmark. PTA parameters are partition counts, whereas window parameters are literal support sizes. The two rightmost columns explicitly show that the temporal--spatial box \(T\times H'\times W'\) is identical for the two operators in every row.}
\label{SM-tab:supp-attention-support-geometry}
\fontsize{7.5}{9}\selectfont
\begingroup
\setlength{\tabcolsep}{4pt}
\begin{tabularx}{\textwidth}{cccccX}
\toprule
\(H=W\) & \(T\) & PTA partitions & PTA channel & \multicolumn{2}{c}{Matched temporal--spatial box} \\
\cmidrule(lr){5-6}
 & & \((s_c,s_h,s_w)\) & patch \(C'\) & PTA \((T,H',W')\) & Window \((W_t,W_h,W_w)\) \\
\midrule
64  & 4  & \((8,4,4)\) & 8 & \((4,16,16)\)  & \((4,16,16)\) \\
64  & 8  & \((8,4,4)\) & 8 & \((8,16,16)\)  & \((8,16,16)\) \\
64  & 16 & \((8,4,4)\) & 8 & \((16,16,16)\) & \((16,16,16)\) \\
128 & 4  & \((8,8,8)\) & 8 & \((4,16,16)\)  & \((4,16,16)\) \\
128 & 8  & \((8,8,8)\) & 8 & \((8,16,16)\)  & \((8,16,16)\) \\
128 & 16 & \((8,8,8)\) & 8 & \((16,16,16)\) & \((16,16,16)\) \\
\bottomrule
\end{tabularx}
\endgroup
\end{table}

\FloatBarrier
\subsubsection{Relation to standard window attention}

Standard window attention computes scaled dot-product attention among tokens inside each local window:
\[
A_w
=
\mathrm{softmax}
\left(
\frac{Q_wK_w^\top}{\sqrt{d}}
\right),
\qquad
Y_w=A_wV_w,
\]
where \(d\) is the per-head embedding dimension. Local windows reduce the quadratic cost of full global attention, but the attention matrix is still formed over all tokens inside each window.

PTA3D instead treats each channel--spatial patch, at each time position, as the comparison unit. Similarity is computed between query and key patches using a Tanimoto-style score,
\[
\mathcal{T}(q,k)
=
\frac{\langle q,k\rangle}
{\langle q,q\rangle+\langle k,k\rangle-\langle q,k\rangle+\epsilon},
\]
where \(q\) and \(k\) are patch vectors and \(\epsilon\) is a small constant for numerical stability. The implementation then compacts the query-side similarity information and applies the resulting patch-level emphasis to the values by element-wise multiplication before reshaping back to the original tensor structure. Thus, PTA3D is closer to a global patch-emphasis mechanism than to token-to-token softmax attention within a local window.

This distinction is important for interpretation. The comparison below is not a proof that Tanimoto similarity and softmax attention are interchangeable. It is a resource comparison under matched temporal and spatial support, designed to test whether the PTA3D mechanism provides a lower-memory alternative for the type of spatio-temporal feature tensors used in this work.

\FloatBarrier
\subsubsection{Matched-support resource benchmark}

The resource benchmark used \(B=2\), \(C=64\), four heads, bfloat16 autocast, \(T\in\{4,8,16\}\), and \(H=W\in\{64,128\}\). Each setting used 20 warm-up iterations and 100 measured iterations. Timing was CUDA-synchronised, and peak memory was measured during the corresponding benchmark pass. Out-of-memory cases were retained and reported rather than removed.

Figure~\ref{SM-fig:supp-attention-matched-support} and Table~\ref{SM-tab:supp-attention-matched-support} show the matched-support results for \(H=W=128\). PTA3D uses substantially less peak reserved GPU memory than the matched-support window-attention baseline. At \(T=4\), PTA3D uses 0.975~GB compared with 10.252~GB for window attention. At \(T=8\), PTA3D uses 2.354~GB compared with 40.439~GB. At \(T=16\), PTA3D completes the forward--backward benchmark with 6.256~GB peak reserved memory, while the matched-support window-attention baseline runs out of memory on the tested H100 80GB GPU.

\begin{figure}[!htbp]
\centering
\includegraphics[width=\textwidth,height=0.72\textheight,keepaspectratio]{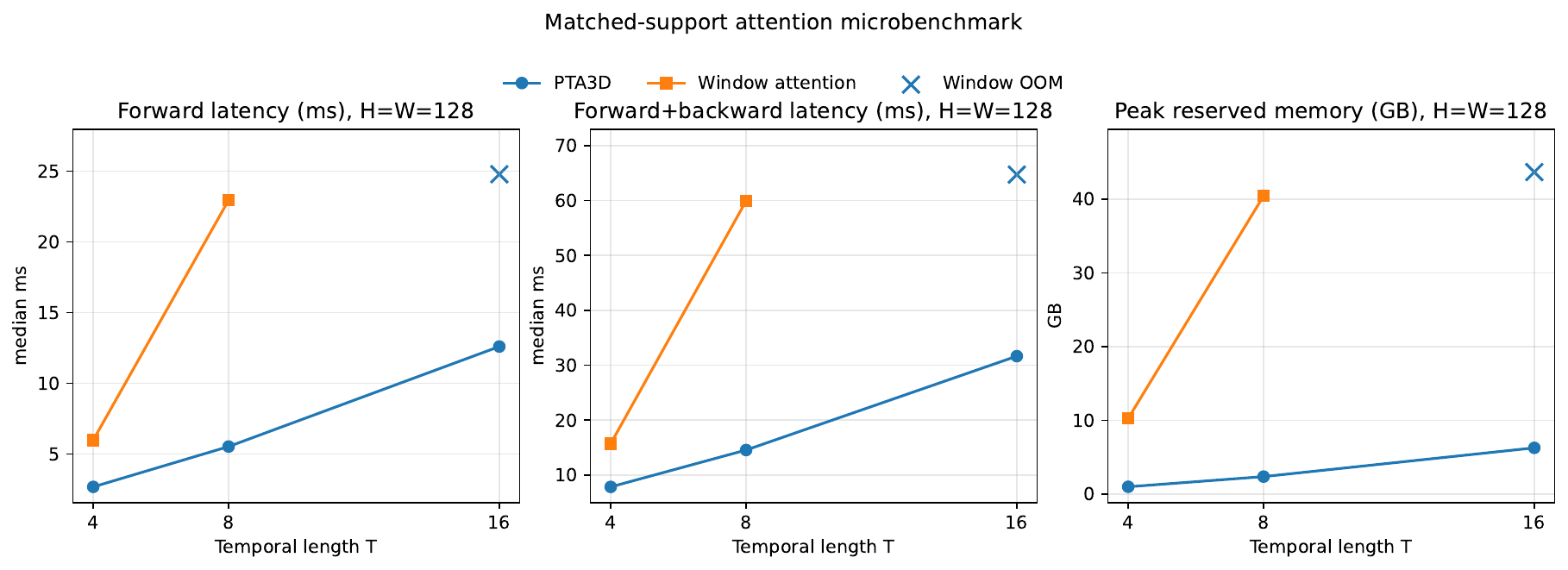}
\caption{Matched-support stage-level resource benchmark for PTA3D and window attention at \(H=W=128\). Both modules use the same input tensor contract, batch size, channel count, number of heads, dtype, GPU, and timing protocol. The window baseline is configured with \((W_t,W_h,W_w)=(T,16,16)\), matching the effective temporal and spatial support of PTA3D. Window-attention out-of-memory cases are retained and marked.}
\label{SM-fig:supp-attention-matched-support}
\end{figure}

\begin{table}[!htbp]
\centering
\caption{Matched-support stage-level attention benchmark at \(H=W=128\). Latency values are medians over 100 measured iterations after 20 warm-up iterations. Peak memory reports CUDA reserved memory during the forward--backward benchmark. OOM cases are retained.}
\label{SM-tab:supp-attention-matched-support}
\fontsize{7.5}{9}\selectfont
\setlength{\tabcolsep}{2.5pt}
\begin{tabular}{rllrrrrrr}
\toprule
\(T\) & \shortstack{PTA\\partitions} & \shortstack{Window\\support} & \shortstack{PTA fwd.\\(ms)} & \shortstack{Window fwd.\\(ms)} & \shortstack{PTA fwd.+bwd.\\(ms)} & \shortstack{Window fwd.+bwd.\\(ms)} & \shortstack{PTA peak\\(GB)} & \shortstack{Window peak\\(GB)} \\
\midrule
4  & \((8,8,8)\) & \((4,16,16)\)  & 2.67  & 5.98  & 7.85  & 15.74 & 0.975 & 10.252 \\
8  & \((8,8,8)\) & \((8,16,16)\)  & 5.52  & 22.94 & 14.56 & 59.94 & 2.354 & 40.439 \\
16 & \((8,8,8)\) & \((16,16,16)\) & 12.60 & OOM   & 31.66 & OOM   & 6.256 & OOM \\
\bottomrule
\end{tabular}
\end{table}

\FloatBarrier

Table~\ref{SM-tab:supp-attention-matched-support-h64} reports the corresponding $H=W=64$ rows from the same matched-support run. At $T=4$, window attention is faster in this smaller setting (4.36 versus 6.23~ms for forward--backward), but it reserves 2.615~GB compared with 0.236~GB for PTA3D. From $T=8$, PTA3D is both faster and lower-memory: at $T=8$, forward--backward latency is 6.24 versus 15.31~ms and peak reserved memory is 0.486 versus 10.170~GB; at $T=16$, the corresponding values are 7.87 versus 60.36~ms and 0.975 versus 40.252~GB. Thus, the spatial-size control confirms PTA3D's memory advantage for every tested temporal length, while its latency advantage begins at $T\geq 8$ on the $64\times64$ grid.

\begin{table}[!htbp]
\centering
\caption{Matched-support stage-level attention benchmark at $H=W=64$, drawn from the same run and protocol as Table~\ref{SM-tab:supp-attention-matched-support}. Latency values are medians over 100 measured iterations after 20 warm-up iterations. Peak memory reports CUDA reserved memory during the forward--backward benchmark.}
\label{SM-tab:supp-attention-matched-support-h64}
\fontsize{7.5}{9}\selectfont
\setlength{\tabcolsep}{2.5pt}
\begin{tabular}{rllrrrrrr}
\toprule
$T$ & \shortstack{PTA\\partitions} & \shortstack{Window\\support} & \shortstack{PTA fwd.\\(ms)} & \shortstack{Window fwd.\\(ms)} & \shortstack{PTA fwd.+bwd.\\(ms)} & \shortstack{Window fwd.+bwd.\\(ms)} & \shortstack{PTA peak\\(GB)} & \shortstack{Window peak\\(GB)} \\
\midrule
4  & \((8,4,4)\) & \((4,16,16)\)  & 1.78 & 1.64  & 6.23 & 4.36  & 0.236 & 2.615 \\
8  & \((8,4,4)\) & \((8,16,16)\)  & 1.77 & 5.79  & 6.24 & 15.31 & 0.486 & 10.170 \\
16 & \((8,4,4)\) & \((16,16,16)\) & 2.66 & 23.93 & 7.87 & 60.36 & 0.975 & 40.252 \\
\bottomrule
\end{tabular}
\end{table}

The benchmark supports the practical motivation for PTA3D as a memory-efficient attention mechanism for satellite image time series. Under matched temporal and spatial support, PTA3D is lower-memory at every completed setting; it is also faster from $T\geq 8$ at $H=W=64$ and across all completed $H=W=128$ comparisons. This result should be interpreted together with the architectural caveat above: the two operators are support-matched, but they are not identical mechanisms. In particular, PTA3D obtains its memory behaviour by operating on channel-partitioned patch comparisons and compacting patch similarities, while window attention performs softmax attention over full-channel token embeddings inside each local spatio-temporal window.

\FloatBarrier
\subsubsection{End-to-end FoTW comparisons at F128 and native F256}

We report two end-to-end experiments with different purposes. The primary control replaces PTA3D throughout the otherwise unchanged architecture with full-temporal, per-stage support-matched window attention. It retains the same two-date FoTW input, data split, training pipeline, and effective total batch size. Native \(256\times256\) (F256) training of this window replacement was not practical on the available H100 80~GB GPUs, so both variants were trained on FoTW inputs resized on the fly to \(128\times128\) (F128). Both runs had a 150-epoch cap and the same 48-hour wall-clock allocation. PTAViT3D packed 16 samples per GPU on two GPUs, whereas the window replacement's raw-fit limit was four samples per GPU and therefore required eight GPUs to retain the same total batch of 32. The fixed-runtime endpoints were epoch 114 for PTAViT3D and epoch 111 for the window replacement. The protocol matches input, optimization batch, and allocation duration, but not GPU-hours because the window variant uses four times as many accelerators.

For a checkpoint-specific native-resolution comparison, both F128 endpoints were evaluated from scratch on the original \(256\times256\) FoTW validation inputs with a newly accumulated confusion matrix. Table~\ref{SM-tab:supp-attention-fullmodel-native256}(a) distinguishes the online F128 endpoint summaries from these fresh F256 metrics. Non-learned rotary-position buffers were regenerated at the native grid for both variants, and the window model's learned relative-position bias table was bicubically interpolated to the native grid.

The second experiment asks how a practical window configuration behaves when trained directly at native F256. It uses the same FoTW architecture outside the attention replacement, optimizer settings, and effective total batch of 80. To fit native resolution, the window model uses $W_t=1$ and feasible stage-wise spatial windows, whereas PTA3D retains full temporal support and different effective spatial patches at several stages. The models also use separately selected best-mIoU epochs and different accelerator counts. Panel~(b) therefore reports a resource-constrained native-resolution configuration of practical interest; it is not interpreted as an operator-matched attention ablation.

\begin{table}[!htbp]
\centering
\caption{End-to-end FoTW attention comparisons. Panel (a) is the primary full-temporal, per-stage support-matched experiment: models were trained at F128 under the same 48-hour limit and effective total batch, and their endpoint checkpoints were reevaluated from scratch at native F256. The F128 columns are online endpoint summaries; the F256 columns use a reset confusion matrix. Panel (b) is a practical native-F256 training experiment that matches effective total batch but not temporal--spatial attention support. Differences in panel (a) are calculated from unrounded scores and can differ by one unit in the final displayed decimal from subtraction of the rounded entries.}
\label{SM-tab:supp-attention-fullmodel-native256}
\fontsize{7.2}{9}\selectfont
\begingroup
\setlength{\tabcolsep}{3.5pt}
\begin{adjustbox}{max width=\textwidth}
\begin{tabular}{lrrrrr}
\toprule
\multicolumn{6}{l}{\textit{(a) Primary support-matched experiment: F128 training and fresh native-F256 evaluation}} \\
\cmidrule(lr){1-6}
Model & Endpoint & \multicolumn{1}{c}{F128 online} & \multicolumn{3}{c}{Fresh native-F256 evaluation} \\
\cmidrule(lr){3-3}\cmidrule(lr){4-6}
 & epoch & mIoU & mIoU & Accuracy & Macro Dice \\
\midrule
PTAViT3D & 114 & 0.8791 & 0.8786 & 0.9450 & 0.9347 \\
Window replacement & 111 & 0.8763 & 0.8751 & 0.9430 & 0.9327 \\
PTAViT3D -- Window & -- & +0.0028 & +0.0034 & +0.0019 & +0.0020 \\
\bottomrule
\end{tabular}
\end{adjustbox}
\par\smallskip
\begin{adjustbox}{max width=\textwidth}
\begin{tabular}{lrrrrll}
\toprule
\multicolumn{7}{l}{\textit{(b) Practical native-F256 training with unmatched attention support}} \\
\cmidrule(lr){1-7}
Model & Best epoch & mIoU & Field recall & Field precision & Training pack & Attention support \\
\midrule
PTAViT3D & 123 & 0.8793 & 0.9239 & 0.8959 & 5/GPU \(\times\) 16 = 80 & Full \(T\); PTA patches \\
Window replacement & 56 & 0.8669 & 0.9187 & 0.8819 & 2/GPU \(\times\) 40 = 80 & \(W_t=1\); feasible stage windows \\
\bottomrule
\end{tabular}
\end{adjustbox}
\endgroup
\end{table}

Table~\ref{SM-tab:supp-attention-fullmodel-resources} reports the accelerator packing and peak-memory evidence for the primary support-matched experiment. It makes explicit that equal total batch and wall-clock duration do not imply equal GPU-hours.

\begin{table}[!htbp]
\centering
\caption{Training and native-resolution inference resource evidence for the primary support-matched FoTW comparison on H100 80~GB GPUs. Training peak memory is CUDA reserved memory per GPU. The window replacement ran out of memory at training batch 5/GPU and native-F256 inference batch 2; the PTAViT3D inference search stopped at the largest tested batch, 8.}
\label{SM-tab:supp-attention-fullmodel-resources}
\fontsize{8}{10}\selectfont
\begin{tabular}{lrrrrrr}
\toprule
Model & Train B/GPU & GPUs & Total B & Train peak/GPU & Native-F256 B & Native peak \\
\midrule
PTAViT3D & 16 & 2 & 32 & 43.304~GB (50.8\%) & 8 & 26.0\% \\
Window replacement & 4 & 8 & 32 & 78.111~GB (91.6\%) & 1 & 85.1\% \\
\bottomrule
\end{tabular}
\end{table}

In the primary support-matched experiment, PTAViT3D is higher on all three fresh native-F256 metrics, by \(+0.0034\) mIoU, \(+0.0019\) accuracy, and \(+0.0020\) macro Dice, using unrounded scores. These single-run differences are descriptive rather than estimates of seed-to-seed uncertainty. The recorded resource difference is larger: the same total training batch used two GPUs for PTAViT3D and eight for the window replacement, and the largest tested native-F256 inference batches were 8 and 1, respectively. The practical native-F256 rows show the same ordering across mIoU, field recall, and field precision, but their larger separation is not attributed to the attention operator alone because temporal--spatial support, selected epoch, and accelerator packing were not matched.

\FloatBarrier

\subsection{Loss-function control}
\label{SM-app:loss-control}

The Fold~1 loss-function control compares the multitask Tanimoto-with-complement loss used for PTAViT3D (FTNMT) with binary cross-entropy (BCE) and an equal-weight BCE+Dice objective. All three variants are evaluated on native PASTIS and the same severe mixed cloud--shadow condition. Their selected checkpoints and effective training batches are reported in Table~\ref{SM-tab:app-training-provenance}; because the continuation phases did not maintain one common effective batch across all trajectories, this experiment is a bounded sensitivity check rather than a fully optimization-path-matched loss ablation.

Table~\ref{SM-tab:app-loss-control} and Fig.~\ref{SM-fig:app-loss-control} show the resulting trade-off among precision, recall, and polygon geometry.

BCE+Dice gives the highest object F1 in both conditions, while FTNMT gives the highest recall and lowest matched-polygon PoLiS. These results do not identify the loss as a causal explanation of the full cloud-robustness matrix, for which the native and cloud-augmented PTAViT3D models remain the primary comparison.

\begin{table}[!htbp]
\centering
\caption{Loss-function object metrics for native PASTIS and severe mixed cloud--shadow conditions. The best value within each condition is shown in bold, with lower PoLiS preferred.}
\label{SM-tab:app-loss-control}
\fontsize{7}{9}\selectfont
\begingroup
\begin{tabular}{llrrrr}
\toprule
Loss & Condition & Object F1 & Precision & Recall & PoLiS (m) \\
\midrule
BCE & native PASTIS & 0.713 & \textbf{0.786} & 0.690 & 49.06 \\
BCE & mixed severe & 0.689 & \textbf{0.743} & 0.682 & 52.32 \\
BCE+Dice & native PASTIS & \textbf{0.726} & 0.759 & 0.733 & 46.25 \\
BCE+Dice & mixed severe & \textbf{0.707} & 0.721 & 0.737 & 48.98 \\
FTNMT & native PASTIS & 0.704 & 0.642 & \textbf{0.823} & \textbf{43.54} \\
FTNMT & mixed severe & 0.672 & 0.590 & \textbf{0.849} & \textbf{44.13} \\
\bottomrule
\end{tabular}
\endgroup
\end{table}

\begin{figure}[!htbp]
\centering
\includegraphics[width=0.75\textwidth,height=0.72\textheight,keepaspectratio]{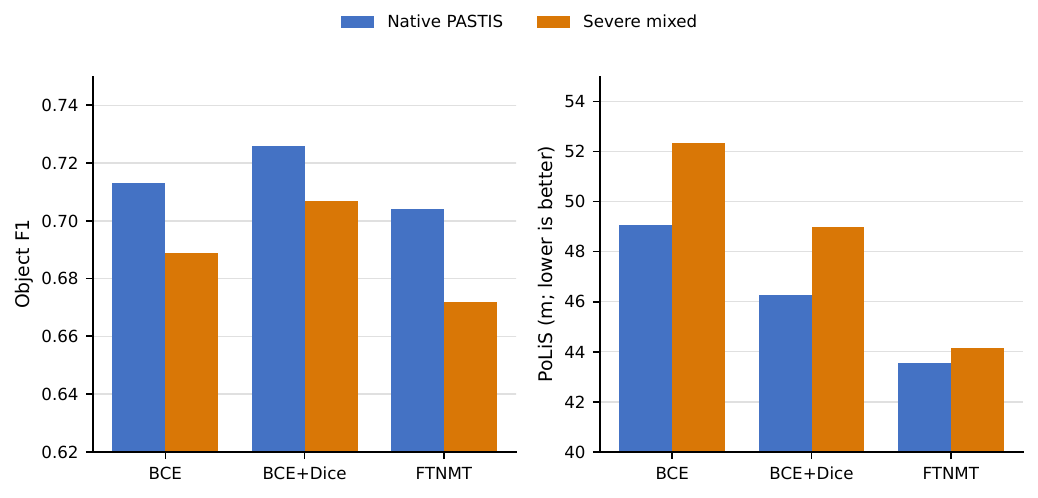}
\caption{Bounded Fold~1 loss-function sensitivity check under native PASTIS and severe mixed cloud--shadow stress. The alternative losses change the precision--recall and PoLiS trade-off; the comparison is not treated as a fully optimization-path-matched ablation.}
\label{SM-fig:app-loss-control}
\end{figure}

\FloatBarrier

\makeatletter
\let\fnum@figure\origfnum@figure
\let\fnum@table\origfnum@table
\makeatother

\endgroup

\section{Modelling characteristics}

The operational-model training configuration used a linear warm-up scheduler for the first epoch, followed by cosine annealing with warm restarts \citep{loshchilov2016sgdr}. The initial learning rate was \texttt{1.e-3}, the half-life was 25 epochs, the restart period was 50 epochs, and optimization used RAdam \citep{Liu2020On}. The controlled PASTIS ablations instead use the protocol in Section~\ref{SM-app:ablation-experimental-design} and Table~\ref{SM-tab:app-training-provenance}; the FoTW attention controls use the fixed-runtime protocols reported in Section~\ref{SM-supp:pta-vs-window-attention}.

\section{Computational considerations}
For the creation of the ePaddocks product, the PTAViT3D model was trained on eight V100 GPUs (32\,GB each) using training chips of size \(128\times128\) pixels. The model consumed 4 time instances of Sentinel-2 imagery with a batch size of 3 per GPU and was trained for 3 days. Inference was performed on the same GPUs with up to 16 Sentinel-2 time instances and a batch size of 52 per GPU.

For the benchmark experiments, the model was evaluated under more than one patch-size regime. ePaddocks and PASTIS were trained with \(128\times128\) patches, whereas Fields of the World (FoTW) and AI4SmallFarms were trained with \(256\times256\) patches. 

The standard FoTW, PASTIS, and AI4SmallFarms benchmark models were trained until convergence on a single DGX system equipped with \(8\times\)H100 GPUs. The fixed-runtime FoTW attention controls and their accelerator allocation are reported separately in Section~\ref{SM-supp:pta-vs-window-attention}.

\subsection{Toy inference throughput comparison}
We also carried out a small toy inference benchmark on an H100 GPU under mixed precision. The purpose of this experiment was to provide a simple throughput reference for PTAViT3D and to contrast it with a modern lighter 2D model, Delineate Anything \citep[hereafter DA;][]{Lavreniuk2025DelineateAnything}.

PTAViT3D was evaluated on a single input tensor of shape \(1\times4\times16\times256\times256\), i.e.\ one 16-step Sentinel-2 time series with four channels processed jointly. DA, which operates on 2D RGB inputs, was evaluated by sequentially applying the model to 16 RGB images of size \(256\times256\times3\) and summing the elapsed time. In both cases, the effective benchmark batch size was therefore 1.

Table~\ref{SM-tab:toy_inference_throughput} reports parameter count, multiply-add count, elapsed time, and throughput for this toy comparison. Throughput rates were calculated from the unrounded timings; they can therefore differ slightly from reciprocals calculated from the displayed three-decimal means.

\begin{table}[!htbp]
\centering
\caption{Toy inference throughput benchmark on an H100 GPU under mixed precision. PTAViT3D processes one joint \(16\)-step 4-channel time series, whereas Delineate Anything (DA) processes \(16\) separate RGB images sequentially. ``Mult-adds'' denotes the total multiply-add count reported by \texttt{torchinfo} for the corresponding input setting.}
\label{SM-tab:toy_inference_throughput}
\fontsize{8}{10}\selectfont
\begin{tabular}{lcccc}
\hline
Method & Params (M) & Mult-adds (G) & Time/series (s) & MPix/s \\
\hline
PTAViT3D & 144.3 & 562.4 & \(1.857 \pm 0.002\) & 0.565 \\
DA (16$\times$2D) & 62.1 & 25.5 & \(0.233 \pm 0.002\) & 4.491 \\
\hline
\end{tabular}
\end{table}

Under this setup, PTAViT3D achieved \(0.539\) series/s, whereas DA achieved \(4.283\) series/s. As expected, the lighter 2D DA model is substantially faster. This result should be interpreted as a practical throughput proxy rather than a perfectly matched architectural benchmark, since PTAViT3D processes the temporal stack jointly whereas DA processes the temporal snapshots independently.

\section{Algorithms}
\label{SM-section_algorithms}
Here we present with \textsc{pytorch} style pseudocode the implementation of some critical components of the modules we developed.

\subsection{S1--S2 feature-level fusion (PTAViT3D-CA)}
\label{SM-alg:s1s2_fusion}

\begin{python}[emphstyle=\ttb, 
caption={Feature-level S1--S2 fusion with cross-attention (PTAViT3D-CA).}, 
emph={ptavit3d_ca_fusion,EncStage,CrossAtt,FuseCAT,Decode,FuseHiLo,return},label={lst:s1s2_fusion}]
def ptavit3d_ca_fusion(X_s2, X_s1, K_stages):
    """
    Inputs:
        X_s2: S2 time series, shape [B, C2, T, H, W]
        X_s1: S1 time series, shape [B, C1, T, H, W]
    Output:
        F: fused features, shape [B, C, T, H', W']
    """

    # 1) Project each modality to 
    # a shared embedding dimension
    F_s2 = Conv1x1x1_s2(X_s2)     # [B, C, T, H, W]
    F_s1 = Conv1x1x1_s1(X_s1)     # [B, C, T, H, W]

    # 2) Early high-resolution 
    # fusion branch (used later)
    F_hi = FuseFirst(F_s2, F_s1)  # [B, C, T, H, W]

    # 3) Stem / initial downsampling 
    # (kept separate per modality)
    F_s2 = Stem_s2(F_s2)
    F_s1 = Stem_s1(F_s1)

    # 4) Multi-scale encoder with 
    # bidirectional cross-attention
    fusions = []
    for l in range(K_stages):

        # Encode each modality 
        # independently at scale l
        F_s2 = EncStage_s2[l](F_s2)
        F_s1 = EncStage_s1[l](F_s1)

        # Bidirectional cross-attention 
        # updates (non-linear fusion signal)
        F_s2 = F_s2+CrossAtt[l](query=F_s1,key=F_s2)
        F_s1 = F_s1+CrossAtt[l](query=F_s2,key=F_s1)

        # Scale-wise fused tensor 
        # (used as skip connection)
        G_l  = FuseCAT[l](concat(F_s2, F_s1))
        fusions.append(G_l)

    # 5) Decoder operating on 
    # fused hierarchy (skip connections)
    F = relu(fusions[-1])
    for l in range(K_stages - 1):
        F = UpCombine[l](F, fusions[-l-2])
        F = DecStage[l](F)

    # 6) Final high-/low-resolution refinement
    F = FuseHiLo(F, F_hi)

    return F
\end{python}

\subsection{S1 pre-processing transformation}
\label{SM-s1transformation_section}

\begin{python}[emphstyle=\ttb, caption={Transformation of Sentinel-1 image tiles for deep learning consumption.}, emph={transform_s1,symlog,return},label={s1transformation}]
def transform_s1(s1_tile):
    # Scale angle from degrees to radians
    s1_tile[0] = s1_tile[0] * np.pi / 180.

    # Define symmetric logarithm
    def symlog(x, epsilon=1.e-5):                 
        return np.sign(x) * np.log(np.abs(x) + epsilon)

    # Apply symlog to VH and VV bands
    for c in [-2, -1]:                            
        s1_tile[c] = symlog(s1_tile[c])
    return s1_tile
\end{python}

\subsection{Raster2Polygons}
\label{SM-alg:raster2polygons}
\begin{python}[emphstyle=\ttb, caption={Conversion of Raster Predictions to Shapefiles.}, 
emph={refined_threshold,preds2shapes,return},label={lst:shapefiles_refined}]
import pandas as pd
import geopandas as gpd
from rasterio import features
from shapely.geometry import shape
import numpy as np
import cv2

def refined_threshold(extent, boundary,
                      threshold_boundary, threshold_extent):

    # Threshold and thin the boundary
    tbound = (boundary > threshold_boundary).astype(np.uint8)
    tbound = cv2.ximgproc.thinning(tbound*255, 
                       cv2.ximgproc.THINNING_GUOHALL)
    tbound = cv2.dilate(tbound, (1,1), iterations=1)

    # Apply refined extent threshold
    thresh = extent * (255 - tbound)
    _,thresh = cv2.threshold(
        thresh, int(255*threshold_extent), 255, 0)
    
    return thresh.astype(np.int16)

def preds2shapes(extent, boundary,
                 threshold_boundary, threshold_extent, meta):
    # Obtain refined threshold mask
    thresh = refined_threshold(
        extent, boundary, threshold_boundary, threshold_extent)
    mask = thresh == 255

    # Generate shapes using rasterio
    shapes_rasterio = features.shapes(thresh, 
             mask=mask, transform=meta['transform'])
    
    # Convert to DataFrame
    df = pd.DataFrame(shapes_rasterio, 
         columns=['geometry', 'value'])
    df['geometry'] = df['geometry'].apply(shape)

    # Convert to GeoDataFrame
    gdf = gpd.GeoDataFrame(df, geometry='geometry')
    gdf.crs = meta['crs']
    
    return gdf
\end{python}

\end{document}